\documentclass{article}

\usepackage{arxiv}

\usepackage[utf8]{inputenc} 
\usepackage[T1]{fontenc}    
\usepackage{amsfonts}       
\usepackage{amsmath,amssymb}
\usepackage{tikz}
\usepackage{enumitem}
\usepackage{hyperref}
\definecolor{dark-green}{rgb}{0.08, 0.65, 0.24}
\hypersetup{
    colorlinks,
    urlcolor = blue,
    citecolor=dark-green
}
\usepackage{comment}
\usepackage{dsfont}
\usepackage{array}
\usepackage{amssymb}
\usepackage{amsmath}
\usepackage{colortbl}
\usepackage{xcolor}
\usepackage{array,multirow,graphicx}
\usepackage{float}
\usepackage{adjustbox}
\usepackage{tabularx,ragged2e,booktabs}
\usepackage{subcaption}
\usepackage{graphicx}
\usepackage{caption}
\usepackage{pgfplots}
\usepackage{adjustbox}
\usepackage{url}            
\usepackage{booktabs}       
\usepackage{nicefrac}       
\usepackage{microtype}      
\usepackage{natbib}
\usepackage{doi}
\usepackage{lipsum}
\definecolor{lgray}{gray}{0.85}
\definecolor{pastelred}{rgb}{1.0, 0.41, 0.38}
\definecolor{pastelgreen}{rgb}{0.47, 0.87, 0.47}
\definecolor{darkpastelblue}{rgb}{0.47, 0.62, 0.8}
\definecolor{darkpastelpurple}{rgb}{0.59, 0.44, 0.84}

\title{Sem@$K$: Is my knowledge graph embedding model semantic-aware?}

\date{}

\author{ \href{https://orcid.org/0000-0002-4682-422X}{\includegraphics[scale=0.06]{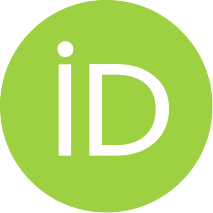}\hspace{1mm}Nicolas Hubert}\thanks{Corresponding author} \\
	Université de Lorraine, CNRS, LORIA, France\\
	Université de Lorraine, ERPI, France\\
	\texttt{nicolas.hubert@univ-lorraine.fr} \\
	\And
	\href{https://orcid.org/0000-0002-2017-8426}{\includegraphics[scale=0.06]{orcid.pdf}\hspace{1mm}Pierre Monnin} \\
	Université Côte d'Azur, Inria, CNRS, I3S, France\\
	\texttt{pierre.monnin@inria.fr} \\
        \And
	\href{https://orcid.org/0000-0002-9876-6906}{\includegraphics[scale=0.06]{orcid.pdf}\hspace{1mm}Armelle Brun} \\
	Université de Lorraine, CNRS, LORIA, France\\
	\texttt{armelle.brun@loria.fr} \\
        \And
	\href{https://orcid.org/0000-0002-4244-684X}{\includegraphics[scale=0.06]{orcid.pdf}\hspace{1mm}Davy Monticolo} \\
        Université de Lorraine, ERPI, France\\
	\texttt{davy.monticolo@univ-lorraine.fr} \\
}

\renewcommand{\headeright}{}
\renewcommand{\undertitle}{}
\renewcommand{\shorttitle}{Sem@$K$: Is my KGEM semantic-aware?}

\hypersetup{
pdftitle={Sem@$K$: Is my knowledge graph embedding model semantic-aware?},
pdfsubject={preprint},
pdfauthor={Nicolas Hubert},
pdfkeywords={First keyword, Second keyword, More},
}

\begin{document}
\maketitle
\begin{abstract}
Using knowledge graph embedding models (KGEMs) is a popular approach for predicting links in knowledge graphs (KGs). Traditionally, the performance of KGEMs for link prediction is assessed using rank-based metrics, which evaluate their ability to give high scores to ground-truth entities. However, the literature claims that the KGEM evaluation procedure would benefit from adding supplementary dimensions to assess.
That is why, in this paper, we extend our previously introduced metric Sem@$K$ that measures the capability of models to predict valid entities w.r.t. domain and range constraints.
In particular, we consider a broad range of KGs and take their respective characteristics into account to propose different versions of Sem@$K$.
We also perform an extensive study to qualify the abilities of KGEMs as measured by our metric.
Our experiments show that Sem@$K$ provides a new perspective on KGEM quality. Its joint analysis with rank-based metrics offers different conclusions on the predictive power of models. Regarding Sem@$K$, some KGEMs are inherently better than others, but this semantic superiority is not indicative of their performance w.r.t. rank-based metrics. In this work, we generalize conclusions about the relative performance of KGEMs w.r.t. rank-based and semantic-oriented metrics at the level of families of models. The joint analysis of the aforementioned metrics gives more insight into the peculiarities of each model. This work paves the way for a more comprehensive evaluation of KGEM adequacy for specific downstream tasks.
\end{abstract}


\section{Introduction}
\label{intro}
A knowledge graph (KG) is commonly seen as a directed multi-relational graph in which two nodes can be linked through potentially several semantic relationships. 
\begin{figure}
    \centering
    \includegraphics[scale=0.45]{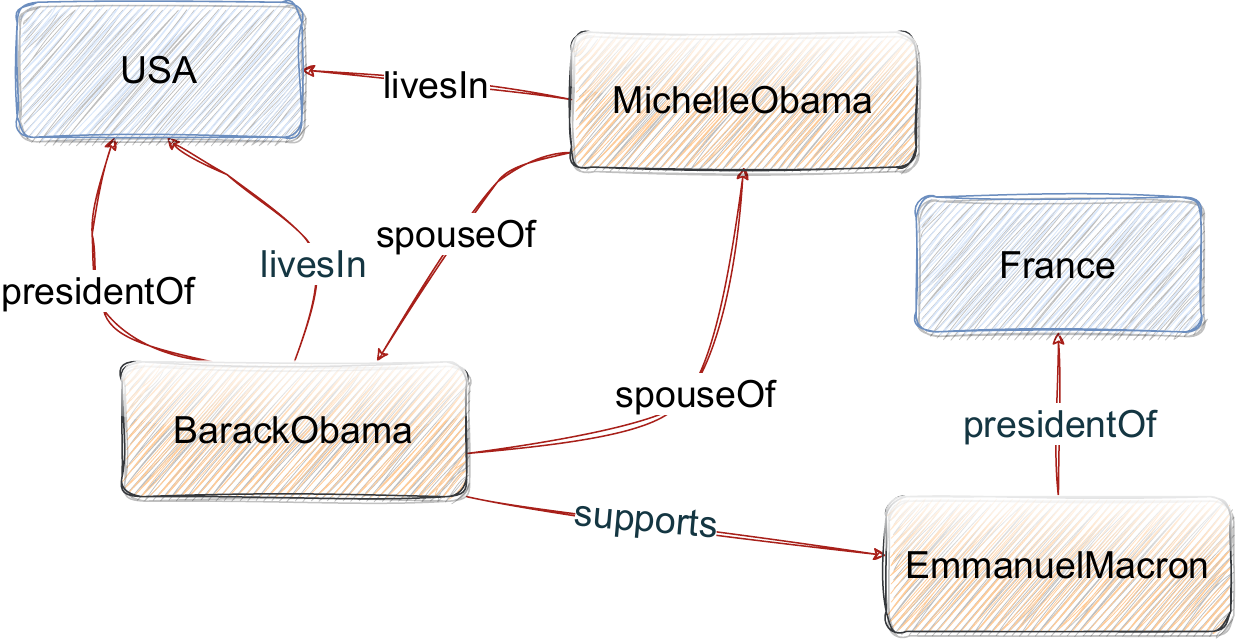}
    \caption{Excerpt of a KG containing some influential political figures and relations holding between them}
    \label{fig:sample-kg}
\end{figure}
More formally, a knowledge graph $\mathcal{KG} = (\mathcal{E},\mathcal{R},\mathcal{T})$ where $\mathcal{E}$, $\mathcal{R}$ and $\mathcal{T} \subseteq \mathcal{E} \times \mathcal{R} \times \mathcal{E}$ are a set of entities (nodes), relations (edge labels) and triples, respectively. 
A KG is represented as a collection of such triples - a.k.a. facts - denoted as $(h,r,t) \in \mathcal{T}$ where $h \in \mathcal{E}$ and $t \in \mathcal{E}$ are two entities of the graph and are respectively named the head and tail of the triple, while $r \in \mathcal{R}$ is a predicate that qualifies the nature of the relationship holding between these entities. For instance, in the sample KG depicted in Fig.~\ref{fig:sample-kg}: $\mathcal{E} =$ \texttt{\{BarackObama,MichelleObama,EmmanuelMacron,USA,France\}} and $\mathcal{R} =$ \texttt{\{spouseOf,presidentOf,supports,livesIn\}}.

KGs are inherently incomplete,  incorrect, or overlapping and thus major refinement tasks include entity matching, question answering, and link prediction~\cite{paulheim2017,wang2017}. The latter is the focus of this paper.
Link prediction (LP) aims at completing KGs by leveraging existing facts to infer missing ones.
In the LP task, one is provided with a set of incomplete triples, where the missing head (resp. tail) needs to be predicted. This amounts to holding a set of triples $\mathcal{T^\prime}$ where, for each triple, either the head $h$ or the tail $t$ is missing. This task can be subdivided into a head prediction phase - which consists in predicting the most plausible head $h$ for each $(?,r,t)$ - and a tail prediction phase - which consists in predicting the most plausible tail $t$ for each $(h,r,?)$. In the sample KG depicted in Fig.~\ref{fig:sample-kg}, an example of triple to be predicted during the tail prediction phase would be \texttt{(EmmanuelMacron,livesIn,?)}, where the expected tail to be inferred is \texttt{France}.
Training a Knowledge Graph Embedding Model (KGEM) firstly requires corrupting existing triples by replacing either their head $h$ or their tail $t$ with another entity to generate negative counterparts. This procedure is called negative sampling~\cite{transe,krompas,jain_iswc}. Secondly, the KGEM iteratively learns to assign higher scores to true triples than to their negative counterparts.

The performance of KGEM for LP is evaluated using rank-based metrics such as Hits@$K$, Mean Rank (MR), and Mean Reciprocal Rank (MRR) that assess whether ground-truth entities are indeed given higher scores~\cite{wang2017,rossi}. 
However, various works recently raised some caveats about such metrics~\cite{berrendorf,hoyt,tiwari}. 
For instance, they are not well-suited for drawing comparisons across datasets~\cite{berrendorf}.
More importantly, they only provide a partial picture of KGEM performance~\cite{berrendorf}.
Indeed, LP can lead to nonsensical triples, such as \texttt{(Barack\-Obama,isFatherOf,USA)}, being predicted as highly plausible facts, although they violate constraints on the domain and range of relations~\cite{jain_iswc,wang2019}. 
KGEMs with such issues may nevertheless reach a satisfying performance in terms of rank-based metrics.

Few works propose to go beyond the mere traditional quantitative performance of KGEMs and address their ability to capture the semantics of the original KG, \textit{e.g.}, domain and range constraints, hierarchy of classes~\cite{paulheim,jain_eswc,monnin}.
According to Berrendorf \textit{et al.}~\cite{berrendorf}, this would give a more complete picture of the performance of a KGEM.
This is why we advocate for additional qualitative and semantic-oriented metrics to supplement traditional rank-based metrics and propose Sem@$K$ to address this need. The relevance of using semantic-oriented metrics -- more specifically Sem@$K$ -- is clearly visible in Fig.~\ref{fig:example}: Sem@$K$ provides a supplementary dimension to the evaluation procedure and allows to confidently choose between two models.
They are equally good in terms of rank-based metrics but Model A predicts entities that are semantically valid w.r.t. the range constraint.

More specifically, in this work, our goal is to assess the ability of popular KGEMs to capture the semantic profile of relations in a LP task, \textit{i.e.}, whether KGEMs predict entities that respects domain and range of relations.
Henceforth, we refer to this aspect as the \emph{semantic awareness} of KGEMs.
To do so, we build on Sem@$K$, a semantic-oriented metric that we previously introduced~\cite{ekaw,dl4kg}
In~\cite{ekaw}, Sem@$K$ was specifically defined for the recommendation task which was seen as predicting tails for a unique target relation.
Sem@$K$ was then extended in~\cite{dl4kg} to the more generic LP task, where not only tails but also heads are corrupted and all relations are considered.
In the present work, we deepen the study of the semantic awareness of KGEMs by proposing different versions of Sem@$K$ that take into account the different characteristics of KGs (\textit{e.g.}, hierarchy of types). 
Moreover, the semantic awareness of a wider range of KGEMs is analyzed -- especially convolutional models. Likewise, a broader array of KGs is used, in order to benchmark the semantic capabilities of KGEMs on mainstream LP datasets. Thus, the following research questions are addressed:
\begin{itemize}
    \item RQ1: how semantic-aware agnostic KGEMs are?
    \item RQ2: how the evaluation of KGEM semantic awareness should adapt to the typology of KGs?
    \item RQ3: does the evaluation of KGEM semantic awareness offer different conclusions on the relative superiority of some KGEMs compared to rank-based metrics?
\end{itemize}
Accordingly, the main contributions of this work are: 
\begin{itemize}
    \item to evaluate KGEM semantic awareness on any kind of KG, we extend a previously defined semantic-oriented metric and tailor it to support a broader range of KGs.
    \item we perform an extensive study of the semantic awareness of state-of-the-art KGEMs on mainstream KGs. We show that most of the observed trends apply at the level of families of models.
    \item we perform a dynamic study of the evolution of KGEM semantic awareness vs. their performance in terms of rank-based metrics along training epochs.
    We show that a trade-off may exist for most KGEMs.
    \item our study supports the view that agnostic KGEMs are quickly able to infer the semantics of KG entities and relations.
\end{itemize}

The remainder of the paper is structured as follows. 
Related work is presented in Section~\ref{relatedwork}. 
Section~\ref{motivations} outlines the main motivations for assessing the semantic awareness of KGEMs. Section~\ref{sematk} subsequently presents Sem@$K$, the semantic-oriented metric that fulfills this purpose. Sem@$K$ comes in different flavors based on the typology of the datasets at hand and the intended use cases. In Section~\ref{experiments}, we detail the datasets and KGEMs used in this work, before presenting the experimental findings in Section~\ref{results}. A thorough discussion is provided in Section~\ref{discussion}.
Lastly, Section~\ref{conclusion} outlines future research directions.

\section{Related work}
\label{relatedwork}
\subsection{Link prediction using knowledge graph embedding models}
Several LP approaches have been proposed to complete KGs. Symbolic approaches relying on rule-based~\cite{amie,amieplus,amie3,anyburl,safran} or path-based reasoning~\cite{pra,deepath,minerva} are somewhat popular but are not considered in the present work. Instead, KGEMs are the focus of this paper.

In particular, this work is concerned with assessing the semantic awareness of agnostic KGEMs. The semantic awareness of such models is defined as their ability to score higher entities whose types belong to the domain and range of relations. Agnostic KGEMs are defined as KGEMs that rely solely on the structure of the KG to learn entity and relation representations. In this respect, these models differ according to several criteria, such as the nature of the embedding space or the type of scoring function~\cite{wang2017,wang2021,ji2021,rossi}. In this work, KGEMs are considered with respect to three main families of models that are traditionally distinguished in the literature.  

\textbf{Geometric models} are additive models that consider relations as geometric operations in the latent embedding space. A head entity $h$ is spatially transformed using an operation $\tau$ that depends on the relation embedding $r$. Then the distance between the resulting vector and the tail entity $t$ is used as a measure for assessing the plausibility of a fact $(h, r, t)$. A distance-based function $\delta$ is used to define the scoring function $f$ of such KGEMs: ${f}(h,r,t)=\operatorname{\delta}(\operatorname{\tau}(h,r),t)$.
A large array of geometric models is purely translational and is based on TransE~\cite{transe} and its extensions. TransE was the earliest introduced geometric model for link prediction. It enforces the sum of the head and relation embeddings to lie in a close neighborhood of the tail embedding. The distance-based function $\delta$ is usually the $L1$ or $L2$ norm. TransE does not properly handle 1-to-N, N-to-1, nor N-to-N relations~\cite{wang2017} and yet has been found to be very efficient in multi-relational settings~\cite{chowdhury}. A myriad of translational models have been proposed since then, such as TransH~\cite{transh}, TransR~\cite{transr}, TransD~\cite{transd}, TransA~\cite{transa} and TransG~\cite{transg}. 
In addition to these purely translational models, some recent geometric models either replace or combine the translation operation with rotation-wise transformations to make their models even more expressive and deal with difficult relational patterns such as symmetric or anti-symmetric relations. A case in point is RotatE~\cite{rotate}, which considers relations as rotations in a complex latent space. $h$, $r$ and $t$ are all embedded in $\mathbb{C}^{d}$ where $d$ denotes the embeddings' dimension. The use of the rotation operation allows RotatE to properly address many relational patterns. In particular, RotatE is able to take account for symmetry -- which is not modeled correctly by TransE. In the wake of RotatE, a handful of other roto-translational models have subsequently been proposed in the literature, \textit{e.g.} QuatE~\cite{quate}, DualE~\cite{duale} and HAKE~\cite{hake}.

\textbf{Semantic matching models} are named this way as they usually use a similarity-based function to define their own scoring function. Semantic matching models are also referred to as matrix factorization models or tensor decomposition models, in the sense that a KG can be seen as a 3D adjacency matrix, in other words a three-way tensor. This tensor can be decomposed into a set of low-dimensional vectors that actually represent the entity and relation embeddings.
Semantic matching models are subdivided into bilinear and non-bilinear models. Bilinear models share the characteristic of capturing interactions between two entity vectors using multiplicative terms. RESCAL~\cite{rescal} is the earliest introduced bilinear model for LP. It models entities as vectors and relations as matrices. The components ${w}_{i,j}$ of the relation matrix $W_{r} \in \mathbb{R}^{d \times d}$ account for the interaction intensity between the $i$-th embedding component of $e_{h} \in \mathbb{R}^{d}$ and the $j$-th embedding component of $e_{t} \in \mathbb{R}^{d}$ under the relation $r$. As pointed out by the original author of RESCAL~\cite{nickel2016}, the parameter complexity of this model can drastically increase when the embedding dimension is large. DistMult~\cite{distmult} alleviates this scalability issue by enforcing all relation-matrices $W_{r}$ to be diagonal. Doing so, DistMult trades computational advantages for less expresiveness. Indeed, DistMult is not able to model anti-symmetric relations. However, DistMult still achieves state-of-the-art performance in most cases~\cite{kadlec}. ComplEx~\cite{complex} relies on complex-valued representations for both entities and relations. Because the Hadamard product is used in the scoring function, ComplEx is not commutative in the complex space and can properly model anti-symmetric relations. Other bilinear models with distinctive features were later proposed, such as Analogy~\cite{analogy}, SimplE~\cite{simple} and DihEdral~\cite{dihedral} that are supposedly more expressive models.
Another category of semantic matching models leverage interactions between entities and relations using non-bilinear operations. For instance, HolE~\cite{hole} uses the circular correlation operation, while TuckER~\cite{tucker} relies on the Tucker decomposition.

\textbf{Neural network-based models} rely on neural networks to perform LP. 
In neural networks, the parameters (\textit{e.g.} weights and biases) are organized into different layers, with usually non-linear activation functions between each of these layers. 
The first introduced neural-network based model for LP is Neural Tensor Network (NTN)~\cite{ntn}. It can be seen as a combination of multi-layer perceptrons (MLPs) and bilinear models~\cite{ji2021}. NTN defines a distinct neural network for each relation. 
This choice of parameterization makes NTN similar to RESCAL in the sense that both models achieve great expressiveness at the expense of computational concerns. 
The most recent neural network-based models rely on more sophisticated layers to perform a broader set of operations. Convolutional models are by far the most representative family of such models. 
They use convolutional layers to learn deep and expressive features from the input data, which pass through such layers to undergo convolution with low-dimensional filters~$\omega$. 
The resulting feature maps subsequently go through dense layers to obtain a final plausibility score. Compared with fully connected neural networks, convolution-based models are able to capture
complex relationships with fewer parameters by learning non-linear features. While ConvE~\cite{conve} reshapes head entity and relation embeddings before concatenating them into a unique input matrix to pass through convolutional layers, ConvKB~\cite{convkb} does not perform any reshaping and also puts the tail embedding into the concatenated input matrix. Other models were later proposed, such as ConvR~\cite{convr} and InteractE~\cite{interacte} that both process triples independently. Another branch of convolutional models also considers the local neighborhood around each central entity. These are based on Graph Neural Networks (GNNs). The most representative GNN-based model for LP is R-GCN~\cite{rgcn}. The key idea is to accumulate messages from the local neighborhood of the central node over multi-hop relations. By doing so, R-GCN is better able to model a long range of relational dependencies. However, R-GCN does not outperform baselines for the LP task~\cite{ferrari}. Subsequent models have claimed superiority over R-GCN: SACN~\cite{sacn} introduces a weighted GCN to adjust the amount of aggregated information from the local neighborhood and KBGAT~\cite{kbgat} relies on attention mechanism to generate more accurate embeddings. More recently, CompGCN~\cite{compgcn} and DisenKGAT~\cite{disenkgat} showcased impressive performance with regard to the LP task.

\subsection{Combining embeddings and semantics}
\label{relatedwork2}
The possibility of using additional semantic information has been extensively studied in recent works~\cite{jain_iswc,krompas,autoeter,tarp,tkrl,transc,transet}. In general, the semantic information stems directly from an ontology, originally defined by Gruber as an "explicit specification of a conceptualization"~\cite{gruber}. Ontologies formally describe a specific application domain of interest (\textit{e.g.} education, pharmacology, etc.) in which several classes (or concepts) and relations are identified and formally specified. Ontologies support KG construction by providing a schema that specifies the nature of entities, the semantic profile of relations, and other constraints that give the KG a semantic coherence.

A significant part of the literature incorporates such semantic information to constrain the negative sampling procedure and generate meaningful negative triples~\cite{jain_iswc,krompas}. For instance, type-constrained negative sampling (TCNS)~\cite{krompas} replaces the head or the tail of a triple with a random entity belonging to the same type (\texttt{rdf:type}) as the ground-truth entity. Jain \textit{et al.}~\cite{jain_iswc} go a step further and use ontological reasoning to iteratively improve KGEM performance by retraining the model on inconsistent predictions.

Semantic information can also be embedded in the model itself.
In fact, some KGEMs leverage ontological information such as entity types and hierarchy. 

Embedding models project entities and relations of a KG into a vector space. Thus, the semantics of the original KG may not be fully preserved~\cite{jain_iswc,paulheim}. As stated by Paulheim~\cite{paulheim}, because embeddings are not meant to preserve the semantics of the KG, they are not interpretable and this can severely hinder explainability in domains such as recommender systems. Consequently, Paulheim~\cite{paulheim} advocates for \textit{semantic embeddings}. Similarly, Jain \textit{et al.}~\cite{jain_eswc} perform a thorough evaluation of popular KGEMs to better assess whether embeddings can express similarities between entities of the same type. A key finding is that because of overlapping relations among entities of different types, fine-grained semantics cannot be properly reflected by embeddings For instance, the task of finding semantically similar entities does not always provide satisfying results when working with entity embeddings~\cite{jain_eswc}. 

\subsection{Evaluating KGEM performance for link prediction}
\label{relatedwork3}
KGEM performance is evaluated in two stages. During the validation phase, KGEM performance is evaluated on the validation set $\mathcal{T}_{valid}$ after regular -- often uniform -- intervals of epochs. This way, the best epoch is identified. During the test phase, KGEM performance is ultimately evaluated on the test set $\mathcal{T}_{test}$ after retrieving the optimal model parameters achieved on the best epoch of validation. Whether during the validation or test phase, KGEM performance is evaluated the same way: sifting through every triple of the test (resp. validation) set, both head and tail predictions are performed. In the case of head prediction, this amounts to taking a triple $(h,r,t)$ from the test (resp. validation) set, hiding the ground-truth head entity -- resulting in $(?,r,t)$ -- and letting the KGEM assign a score to every possible entity as a candidate for the head position. These scores are finally ordered and reflect the plausibility of such facts. Tail prediction is performed analogously on $(h,r,?)$.

In both cases, the rank of the ground-truth entity from the test (resp. validation) set is used to compute aggregated rank-based metrics based on the top-$K$ scored entities. The rank of the ground-truth entity can be determined in two different ways that depend on how observed facts -- \textit{i.e.} facts that already exist in the KG -- are considered. In the raw setting, observed facts outranking the ground-truth are not filtered out, while this is the case in the filtered setting. For instance, assuming head prediction is performed on the given ground-truth triple $($\texttt{BarackObama}, \texttt{livesIn}, \texttt{USA}$)$, a KGEM may assign a lower score to this triple than to the following triple: $($\texttt{MichelleObama}, \texttt{livesIn}, \texttt{USA}$)$. The latter triple actually represents an observed fact. 
In the raw setting, this triple would not be filtered out from top-$K$ scored triples. This can cause the evaluation procedure to not properly assess the KGEM performance. This is why in practice, the filtered setting is commonly preferred. In the present work, the filtered setting is also used.

KGEM performance is almost exclusively assessed using the following rank-based metrics: Hits@$K$, Mean Rank (MR), and Mean Reciprocal Rank (MRR)~\cite{hoyt}. Disagreements exist as to how and when these metrics can be used and compared properly. In the following, we recall their definitions and discuss their limits.
\\

\textbf{Hits@$K$}~(Eq.~(\ref{eq:hitsatk})) accounts for the proportion of ground-truth triples appearing in the first $K$ top-scored triples:
\begin{equation}
\mathrm{Hits}@ {K}=\frac{1}{|\mathcal{B}|} \sum_{q \in \mathcal{B}} \mathds{1}\left[\mathrm{rank}(q) \leq K\right]
\label{eq:hitsatk}
\end{equation}
where $\mathcal{B}$ is the batch of ground-truth triples, $\mathrm{rank}(q)$ is the position of the ground-truth triple $q$ in the sorted list of triples, and $\mathds{1}\left[\mathrm{rank}(q) \leq K\right]$ yields $1$ if $q$ is ranked between $1$ and $K$, $0$ otherwise.
This metric is bounded in the $ \left[0,1\right]$ range and its values increase with $K$, where the higher the better.

\textbf{Mean Rank (MR)}~(Eq.~(\ref{eq:mr})) corresponds to the arithmetic mean over ranks of ground-truth triples: 
\begin{equation}
\operatorname{MR}=\frac{1}{|\mathcal{B}|} \sum_{q \in \mathcal{B}} \mathrm{rank}(q)
\label{eq:mr}
\end{equation}
This metric is bounded in the $\left[0,|\mathcal{E}|\right]$ interval, where $|\mathcal{E}|$ stands for the number of entities in the KG, where the lower the better.

\textbf{Mean Reciprocal Rank (MRR)}~(Eq.~(\ref{eq:mrr})) corresponds to the arithmetic mean over reciprocals of ranks of ground-truth triples:
\begin{equation}
\operatorname{MRR}=\frac{1}{|\mathcal{B}|} \sum_{q \in \mathcal{B}} \frac{1}{\mathrm{rank}(q)}
\label{eq:mrr}
\end{equation}
Contrary to MR, MRR is a metric bounded in the $\left[0,1\right]$ interval. Higher results indicate better performance. Because this metric does not use any threshold $K$ compared to Hits@$K$, it is less sensitive to outliers. In addition, it is often used for performing early stopping and for tracking the best epoch during training~\cite{berrendorf,hoyt}.

As mentioned in Section~\ref{intro}, these metrics present some caveats.
LP is often used to complete knowledge graphs, where the Open World Assumption (OWA) prevails. KGs are incomplete and, due to the OWA, an unobserved triple used as a negative one can still be positive. It follows that traditional evaluation methods based on rank-based metrics may systematically underestimate the true performance of a KGEM~\cite{wang2019}.

In addition, the aforementioned rank-based metrics have intrinsic and theoretical flaws, as pointed out in several works~\cite{berrendorf,hoyt,tiwari}. For example, Hits@$K$ does not take into account triples whose rank is larger than $K$. As such, a model scoring the ground-truth in position $K + 1$ would be considered equally good as another model scoring the ground-truth in position $K + d$ with $d \gg 1$. It follows that Hits@$K$ is not a suitable metric for drawing comparisons between models~\cite{hoyt}.
MR alleviates this concern as it does not consider any threshold $K$. Therefore, MR allows to compare KGEM performance on the same dataset. Nonetheless, MR is sensitive to the number of KG entities (see Eq.~(\ref{eq:mr}))~\cite{berrendorf}: a MR of $10$ indicates very good performance if the set of entities is in the thousands, but it would indicate poor performance if the set of entities is much more restricted. Therefore, MR does not allow comparisons across datasets.

Recent works recommend using adjusted version of the aforementioned metrics. The Adjusted Mean Rank (AMR) proposed in \cite{ali2022} compares the mean rank against the expected mean rank under a model with random scores. In \cite{berrendorf}, Berrendorf \textit{et al.} transform the AMR to define the Adjusted Mean Rank Index (AMRI) bounded in the $ \left[0,1\right]$ interval. This way, a value of $1$ indicates an optimal performance of the model. A value of $0$ indicates a model performance similar to a model assigning random scores. A negative value indicates that the model performs worse than random. However, all these attempts at producing a better evaluation framework still focus on the quantitative assessment of KGEMs, \textit{i.e.} the improvement of already existing rank-based metrics.

In Section~\ref{relatedwork2}, several approaches incorporating the semantics of entities and relations into the embeddings were mentioned. However, in such cases, the use of semantic information such as entity types and the hierarchy of classes is intended to improve KGEM performance in terms of the aforementioned rank-based metrics only. The underlying semantics of KGs is considered as an additional source of information during training but the ability of KGEMs to generate predictions in accordance with these semantic constraints is never directly addressed.
This encourages further assessment of the semantic capabilities of KGEMs, as firstly suggested in~\cite{dl4kg}.
In our work, we directly address this issue by assessing to what extent KGEMs are able to give high scores to triples whose head (resp. tail) belongs to the domain (resp. range) of the relation. When such information is not available -- \textit{i.e.} the KG does not rely on a schema containing \texttt{rdfs:domain} and \texttt{rdfs:range} properties -- extensional constraints can still be used to evaluate the semantic capabilities of KGEMs, as detailed in Section~\ref{sematk}.

\section{Motivations and problem formulation}
\label{motivations}
\subsection{Motivating example}
\begin{figure}
    \centering
    \includegraphics[scale=0.55]{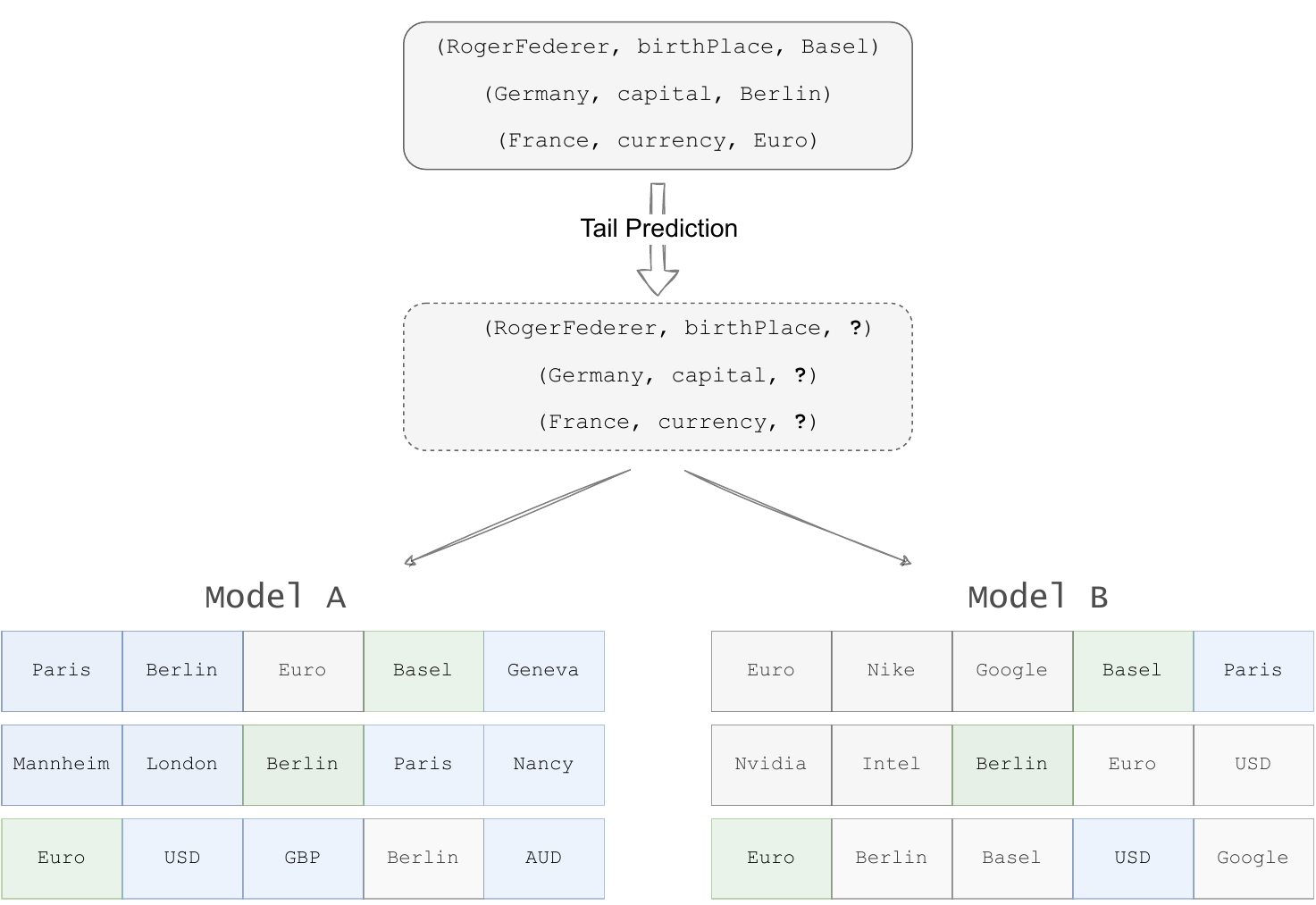}
    \caption{Motivating example. Tail prediction is performed for the three test triples contained in the upper insert. Model A and Model B output scores for each possible entity and only the top-five ranked tail candidates are depicted here. Model A and Model B have the same Hits@1, Hits@3 and Hits@5 values. But Model A has better semantic capabilities. Green, blue and white cells respectively denote the ground-truth entity, entities other than the ground-truth and semantically valid, and entities other than the ground-truth and semantically invalid}
    \label{fig:example}
\end{figure}

To motivate the use of Sem@$K$ to evaluate KGEM quality, this section builds upon a minimalist example which is representative of the issue encountered while benchmarking the performance of several KGEM on the same dataset. 
As depicted in Fig.~\ref{fig:example}, two KGEMs that have been trained on the whole training set are tested on a batch of test triples. These KGEMs are referred to as Model A and Model B. Without loss of generality, it is assumed that the test set only comprises the three triples shown in Fig.~\ref{fig:example}.
For the sake of clarity, only the tail prediction pass and the top-5 ranked candidate entities are depicted in Fig.~\ref{fig:example}. It should be noted that the performance of both models are strictly equal in terms of MR, MRR and Hits@$K$ with $K\leq5$: MR=8/3, Hits@1=1/3, Hits@3=2/3 and Hits@5=1. MRR can only be computed knowing the total number of entities in the KG but two models having the same MR on the same dataset have \textit{de facto} the same MRR as well. Distinguishing between these two models by relying solely on traditional rank-based metrics is not possible. One might even draw the misleading conclusion that these models are equally good. But they are not: Model A gives high scores to semantically valid entities with regard to the range of relations, while Model B semantic awareness is very low. In other words, Model B does not capture the semantic profile of  relations well. This case in point illustrates the need for using semantic-oriented metrics alongside common rank-based metrics, so as to better assess the overall quality of KGEMs.

\subsection{Problem formulation}
\label{problem-formulation}
The traditional evaluation of KGEMs solely based on rank-based metrics can be flawed for several reasons. First, KGEMs benchmarked on the same test sets can exhibit very similar results. Using only rank-based metrics, the final choice only depends on the best achieved MRR and/or Hits@$K$. This raises the questions whether the chosen model is actually the best one, or whether its slight superiority over other KGEMs can be due to other factors such as better hyperparameter tuning or better modeling of a relational pattern highly present in the test set. Moreover, using only rank-based metrics does not provide the full picture of KGEM quality for the downstream LP task, as some dimensions of KGEMs are left unassessed (see Section~\ref{relatedwork}). 
In this work, contrary to the mainstream approach consisting in comparing KGEM performance exclusively in terms of rank-based metrics, the trained KGEMs are also evaluated in terms of Sem@$K$ which measures the ability of KGEMs to predict semantically valid triples with respect to the domain and range of relations.

\section{Measuring KGEMs semantic awareness with Sem@$K$}
\label{sematk}

The standard LP evaluation protocol consists in reporting aggregated results, considering the rank-based metrics presented in Section~\ref{relatedwork3}. As mentioned above, these metrics only provide a partial picture of KGEM performance~\cite{berrendorf}. To give a more comprehensive assessment of KGEMs, we aim at assessing their semantic awareness using our proposed metric called Sem@$K$~\cite{ekaw,dl4kg}.
In~\cite{ekaw}, Sem@$K$ was specifically defined for the recommendation task which was seen as predicting tails for a unique target relation.
Sem@$K$ was then extended in~\cite{dl4kg} to the more generic LP task, where not only tails but also heads are corrupted and all relations are considered.
In this work, this original formalization for LP is presented in Section~\ref{sembase}, and enriched to take into account schemaless KGs (Section~\ref{sematk-ext}), or KGs with a class hierarchy (Section~\ref{sematk-wup}).
As a consequence, Sem@$K$ comes in 3 different versions (respectively denoted Sem@$K$[base], Sem@$K$[ext], and Sem@$K$[wup]) so as to adapt to KG typology. 
These distinct versions and their adequacy regarding the KG at hand are summarized in Table~\ref{tab:typology} and further detailed below.
In the following, when no suffix is provided, it is assumed that we are concerned with Sem@$K$ in general, regardless of the actual version.

\begin{table}
\centering
\caption{Typology of KGs and their respective adequacy for the presented Sem@$K$ versions}\label{tab:typology}
\begin{tabular}{lccc}
\toprule 
KG type & \multicolumn{1}{c}{Sem@$K$[ext]} & \multicolumn{1}{c}{Sem@$K$[base]} & \multicolumn{1}{c}{Sem@$K$[wup]} \\
\midrule
Schemaless & $\times$ &  &   \\
Schema-defined, w/o class hierarchy  & $\times$ & $\times$ &  \\
Schema-defined, w/ class hierarchy & $\times$ & $\times$ & $\times$  \\
\bottomrule
\end{tabular}
\end{table}

\subsection{Definition of Sem@$K$[base]}
\label{sembase}

This version of Sem@$K$ (Eq.~\eqref{eq:sematk}) accounts for the proportion of triples that are semantically valid in the first $K$ top-scored triples:

\begin{equation}
\mathrm{Sem} @ {K}=\frac{1}{|\mathcal{B}|} \sum_{q \in \mathcal{B}}\frac{1}{{K}}
\sum_{q' \in \mathcal{S}^{K}_q} \operatorname{compatibility}(q,q')
\label{eq:sematk}
\end{equation} 
where, given a ground-truth triple $q = (h,r,t)$, $\mathcal{S}^{K}_q$ is the list of the top-$K$ candidate triples scored by a given KGEM (\textit{i.e.} by predicting the tail for $(h,r,?)$ or the head for $(?,r,t)$). 
The function $\operatorname{compatibility}(q,q')$~(Eq.~(\ref{eq:compat})) assesses whether the candidate triple $q'$ is semantically compatible with its ground-truth counterpart $q$. In this work, by semantic compatibility we refer to the fact that the predicted head (resp. tail) belongs to the domain (resp. range) of the relation:
\begin{equation}
\operatorname{compatibility}(q,q') = \begin{cases}
      1, & \text{if}\ \operatorname{types}(q'_{h}) \cap \operatorname{domain}(q_{r}) \neq \emptyset \wedge 
      \operatorname{types}(q'_{t}) \cap \operatorname{range}(q_{r}) \neq \emptyset \\
      0, & \text{otherwise}
    \end{cases}
\label{eq:compat}
\end{equation}
where $\operatorname{types}(e)$ returns the types of entity $e$ and $\operatorname{domain}(r)$ (resp. $\operatorname{range}(r)$) is the domain (resp. range) of the relation $r$. $q_{r}$, $q'_{h}$, and $q'_{t}$ denote the ground-truth relation, the head and the tail of the ranked triple $q'$, respectively.
It is noteworthy that with this formula, we allow entities to instantiate multiple types, and domains and ranges to be defined with multiple types. 
Sem@$K$ is bounded in the $\left[0,1\right]$ interval. 
Compared to Hits@$K$ (Eq.~(\ref{eq:hitsatk})), Sem@$K$ is non-monotonic: increasing $K$ can lead to either lower or higher Sem@$K$ values.

\subsection{A note on Sem@K, untyped entities, and the Open World Assumption}
\label{subsubsection:sematk-owa-remark}

Traditional KGEM evaluation can be performed in all situations, regardless of whether entities are typed or whether the KG comes with a proper schema. When measuring the semantic capability of KGEMs -- \textit{e.g.} with Sem@$K$ -- some concerns arise. For instance, a fair question to ask is the following: how should untyped entities be considered? Indeed, in some KGs, some entities are left untyped. For example, in DBpedia the entity \texttt{dbr:1.\_FC\_Union\_Berlin\_players} does not belong to any other class than \texttt{owl:Thing}. In some other cases as in DB15K and DB100K~\cite{damato2021}, entities have incomplete typing.

Under the OWA, an untyped entity should not count in the calculation of Sem@$K$ since it is not possible to determine whether this entity has no types or no known types. 
Although this seems to be a fair option, it raises a major issue: it makes possible to score different sets of entities with rank-based metrics and Sem@$K$, which is not desirable. If there are $M$ untyped entities in an ordered list of ranked entities, Hits@$K$, MR and MRR are calculated regardless of this fact, \textit{i.e.} still taking into account the $M$ untyped entities. However, Sem@$K$ cannot be calculated for these $M$ untyped entities. As such, MR, MRR and Hits@$K$ would be calculated on the original entity set, whereas Sem@$K$ would be computed on a different set of entities. 
This issue would be even more acute in the case of Sem@$1$ when the first ranked entity is untyped: Hits@$1$ and Sem@$1$ would be calculated on two different entities, which is not acceptable. Consequently, one strategy consists in removing untyped entities from the evaluation protocol, both regarding rank-based and semantic-based metrics. By doing so, consistency is ensured in the ranked list of entities across rank-based and semantic metrics evaluation.

\subsection{Sem@$K$[ext] for schemaless KGs}
\label{sematk-ext}

Not all KGs come with a proper schema, \textit{e.g.} relations do not appear in any \texttt{rdfs:domain} or \texttt{rdfs:range} clauses. In that particular situation, it can still be desirable to assess the semantic awareness of KGEMs. One approach is to maintain a list of all entities that have been observed as heads (resp. tails) of each relation $r$ : $\mathrm{domain}(r) = \{e : \exists (e,r,t) \in \mathcal{T}\}$ (resp. $\mathrm{range}(r) = \{e : \exists (h,r,e) \in \mathcal{T}\}$). 
Hence, in this case, Sem@$K$[ext] is defined as in Eq.~\eqref{sematk} and~\eqref{eq:compat} but using these definitions of domain and range.
Therefore, an entity will be considered as a semantically valid head (resp. tail) with respect to the relation if this entity appears as a head (resp. tail) in any other triple observed in the KG with the same relation.

\subsection{Sem@$K$[wup]: a hierarchy-aware version of Sem@$K$}
\label{sematk-wup}

\begin{figure}
    \centering
    \includegraphics[scale=0.595]{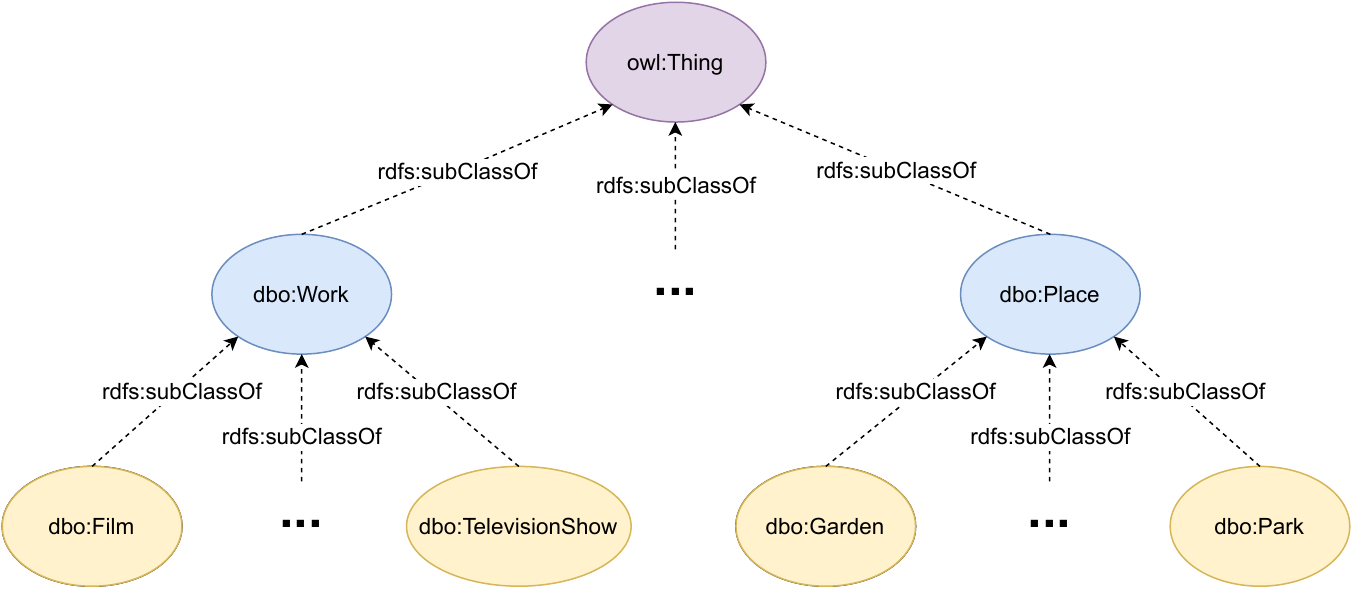}
    \caption{Excerpt from the DBpedia class hierarchy}
    \label{fig3}
\end{figure}

Sem@$K$ as previously defined equally penalizes all entities that are not of the expected type. 
However, KGs may be equipped with a class hierarchy that, in turn, can support a more fine-grained penalty for entities depending on the distance or similarity between their type and the expected domain (resp. range) in this hierarchy.
To illustrate, consider Figure~\ref{fig3} that depicts a subset of the DBpedia ontology \texttt{dbo} class hierarchy. 
Using the hierarchy-free version of Sem@$K$ for the test triple $($\texttt{dbr:The\_Social\_Network}, \texttt{dbo:director}, \texttt{dbr:David\_Fincher}$)$, predicting \texttt{dbr:Friends} or \texttt{dbr:Central\_Park} as head would be penalized the same way in the $\operatorname{compatibility}$ function. However, it is clear that an entity of class \texttt{dbo:TelevisionShow} is semantically closer to \texttt{dbo:Film} -- the domain of the relation \texttt{dbo:director} -- than an entity of class \texttt{dbo:Park}, and thus should be less penalized.

To leverage such a semantic relatedness between concepts in Sem@$K$, the $\operatorname{compatibility}$ function can be adapted accordingly: 
\begin{equation}
\operatorname{compatibility}(q,q^{\prime}) = \min\left(\max\limits_{\substack{c \in \mathrm{type}(q^{\prime}_h) \\ c^{\prime} \in \mathrm{domain}(q_r)}} \sigma(c, c^{\prime}),\ \max\limits_{\substack{c \in \mathrm{type}(q^{\prime}_t) \\ c^{\prime} \in \mathrm{range}(q_r)}} \sigma(c, c^{\prime})\right)
\label{compat-wup}
\end{equation}
\noindent where $\sigma(c, c')$ measure the semantic similarity between the two classes $c$ and $c'$ based on the class hierarchy. 
It should be noted that in this formula $\mathrm{type(e)}$ only returns the most specific classes instantiated by $e$.

Several similarity measures $\sigma$ have been proposed in the literature~\cite{rada,wup,leacock,resnik,li2003}.
In this work, the Wu-Palmer similarity~\cite{wup} (Eq.~\ref{eq:wup}) is used:
\begin{equation}\label{eq:wup}
\sigma\left(c, c^{\prime}\right)=\frac{2 \times \delta\left(c \wedge c^{\prime}, \rho\right)}{\delta\left(c, c \wedge c^{\prime}\right)+\delta\left(c^{\prime}, c \wedge c^{\prime}\right)+2 \times \delta\left(c \wedge c^{\prime}, \rho\right)}
\end{equation}
where $\rho$ is the root of the hierarchy (\textit{e.g.} \texttt{owl:Thing}), $\delta\left(c, c^{\prime}\right)$ is the number of edges linking $c$ to $c^{\prime}$, and $c \wedge c^{\prime}$ represents the least common subsumer of $c$ and $c^{\prime}$.
The Wu-Palmer similarity is well suited to a class hierarchy and provides a good indication of the semantic relatedness between the domain (resp. range) class and the classes of the chosen entity.
This gives rise to the Sem@$K$[wup] version.

Considering the example in Fig.~\ref{fig3}, using the Wu-Palmer score in the calculation of Sem@$K$, a head prediction of \texttt{dbr:Friends} and \texttt{dbr:Central\_Park} for the ground-truth triple $($\texttt{dbr:The\_Social\_Network}, \texttt{dbo:director}, \texttt{dbr:David\_Fincher}$)$ are now differently penalized.  The instance \texttt{dbr:Friends} is of type \texttt{dbo:TelevisionShow}, so we have: $\sigma\left(\texttt{dbo:TelevisionShow}, \texttt{dbo:Film}\right)=1/2$. The instance \texttt{dbr:Central\_Park} is of type \texttt{dbo:Park}, so we have: $\sigma\left(\texttt{dbo:Park}, \texttt{dbo:Film}\right) = 0$. 
This illustrates that incorporating the Wu-Palmer score into Sem@$K$ calculation leads to more precise semantic comparisons that take into account the available class hierarchy. It should be noted that comparing the classes of a candidate entity with the expected class can result in the same penalization as the vanilla version of Sem@$K$. However, in most cases, two classes do not lie in a completely disjoint part of the hierarchy of classes. As such, the Wu-Palmer score between the classes of a candidate entity and the expected class is rarely $0$.

\section{Experimental setting}
\label{experiments}
\subsection{Datasets}
In order to draw reliable and general conclusions, a broad range of KGs are used in our experiments. They have been chosen due to their mainstream adoption in recent research works around KGEMs for LP and the fact that they have different characteristics (\textit{e.g.} entities, relations, classes, presence of a class hierarchy). In this section, the schema-defined and schemaless KGs used in the experiments are detailed. Note that in our experiments, all the schema-defined KGs come with a class hierarchy inherited from either Freebase~\cite{freebase}, DBpedia~\cite{dbpedia}, YAGO~\cite{yago}, or \texttt{schema.org}. To meet requirements for evaluating KGEMs w.r.t. Sem@$K$ (\textit{i.e.} classes instantiated by entities, domain and range for relations), when necessary, subsets of schema-defined KGs are used, as explained in Section~\ref{schemadefinedkgs}. Among the 4 schema-defined KGs presented hereafter, FB15K237-ET, DB93K and YAGO3-37K are derived from already existing KGs, while YAGO4-19K was specifically created and is made available on Zenodo\footnote{\url{https://doi.org/10.5281/zenodo.7526244}} and GitHub\footnote{\url{https://github.com/pmonnin/YAGO4-LP}}. The other KGs are made available on GitHub\footnote{\url{https://github.com/nicolas-hbt/benchmark-sematk}}.

\subsubsection{Schema-defined KGs}
\label{schemadefinedkgs}
The statistics of the datasets FB15K237-ET, DB93K, YAGO3-37K and YAGO4-19K used in our experiments are provided in Table~\ref{tab:kg-schema}.
As discussed in Section~\ref{subsubsection:sematk-owa-remark}, to create an experimental evaluation setting as unbiased and flawless as possible, the schema-defined KGs used in the experiments are filtered to keep typed entities only. This way, Sem@$K$ is calculated under the CWA.

\textbf{FB15K237-ET} derives from FB15K~\cite{transe} -- a dataset extracted from the cross-domain Freebase KG~\cite{freebase}. In these experiments, we do not use FB15K, as it has been noted that this dataset suffers from test leakage, \textit{i.e.}, a large number of test triples can be obtained by simply inverting the position of the head and tail in the train triples~\cite{toutanova}.
The later introduced FB15K237 dataset~\cite{toutanova} is a subset of the original FB15K without these inverse relations. To the best of our knowledge, there is no schema-defined version of FB15K237. However, a schema-defined version of FB15K is provided in~\cite{tkrl}. Consequently, we based ourselves on this version of FB15K and mapped the extracted entity types, relation domains and ranges to the entities and relations found in FB15K237. The resulting schema-defined version of FB15K237 is named FB15K237-ET and includes only typed entities. Besides, validation and test sets contain triples whose relation have a well-defined domain (resp. range), as well as more than 10 possible candidates as head (resp. tail). This ensures Sem@$K$ is not unduly penalized and can be calculated on the same set of entities as Hits@$K$ and MRR, at least until $K=10$.

\textbf{DB93K} is a subset of DB100K, which was first introduced in~\cite{ding2018}. A slightly modified version of DB100K has been proposed in~\cite{damato2021}.
Contrary to the initial version of DB100K, the latter version is schema-defined: entities are properly typed and most relations have a domain and/or a range. This second version is considered in the following experiments. However, some inconsistencies were found in the dataset. Some DBpedia entities only instantiate Wikidata\footnote{\url{https://www.wikidata.org/}} or \texttt{schema.org}\footnote{\url{https://schema.org/}} classes, while instantiation of classes from the DBpedia ontology actually exist.
Moreover, some entities are only partially typed. It must also be noted that domains and ranges of relations have been extracted from DBpedia more than two years ago. DBpedia is a communautary and open-source project: DBpedia classification system relies on human curation, which sometimes implies a lack of coverage for some resources and updates for others. Consequently we associated all relations of DB100K to their most up-to-date domains and ranges\footnote{SPARQL queries were fired against DBpedia as of November 9, 2022}. Similarly for entities, we associated all entities to their most up-to-date classes.
Finally, we removed all untyped entities as well as validation and test relations having less than $10$ observed entities, so as not to unfairly penalize Sem@$K$ results in the validation and test phases.

\textbf{YAGO3-37K} derives from the schema-defined YAGO39K dataset~\cite{transc} extracted from the cross-domain YAGO3 KG~\cite{yago}. Compared to the original YAGO39K, in our experiments only typed entities are kept. In addition, relations having less than 10 observed heads or tails in the training set are discarded from the validation and test splits, for the same reason that keeping them would not reflect the actual Sem@$K$ values. It should be noted that in the YAGO3 ontology, most relations have very generic domains and ranges which are very close to the root of the ontology hierarchy. To produce a more challenging evaluation setting of the models' semantic awareness, a subset of hard relations was identified and only validation and test triples whose relation belongs to this subset are kept. The resulting dataset is named YAGO3-37K.

We built \textbf{YAGO4-19K} with several other subsets of the YAGO4 knowledge graph~\cite{tanonWS20}.
Similarly to other datasets, we focused on relations with a defined domain and range, and more than $10$ triples to constitute the validation and test sets.
We purposedly favored difficult relations to feature in the validation and test sets.
To enrich the training set, additional relations were added based on a manual selection. 
Selected relations in validation and test sets as well as additional relations in the training set can be found on the GitHub repository of the dataset. 
It should be noted that the class hierarchy associated with entities in YAGO 4 is \texttt{schema.org}. 

\begin{table}
\centering
\caption{Statistics of the schema-defined, hierarchical KGs used in the experiments}\label{tab:kg-schema}
\begin{tabular}{lrrrrrr}
\toprule 
Dataset & \multicolumn{1}{c}{$|\mathcal{E}|$} & \multicolumn{1}{c}{$|\mathcal{R}|$} & \multicolumn{1}{c}{$|\mathcal{C}|$} & \multicolumn{1}{c}{$|\mathcal{T}_{train}|$} & \multicolumn{1}{c}{$|\mathcal{T}_{valid}|$} & \multicolumn{1}{c}{$|\mathcal{T}_{test}|$}\\
\midrule
FB15K237-ET & 14,541 & 237 & 532 & 271,575 & 15,337 & 17,928 \\
DB93K  & 92,574 & 277 & 311 & 237,062 & 18,059 & 36,424 \\
YAGO3-37K & 37,335 & 33 & 132 & 351,599 & 4,220 & 4,016  \\
YAGO4-19K & 18,960 & 74 & 1,232 & 27,447 & 485 & 463  \\
\bottomrule
\end{tabular}
\end{table}

\subsubsection{Extensional KGs}
Another range of datasets used in these experiments do not come with an ontological schema. In particular, this means relations do not have a clearly-defined domain or range. Although Codex-S and Codex-M are based on the Wikidata schema which does possess property constraints linking subject types to value type constraints, we limit ourselves to KGs that represent this information with \texttt{rdfs:domain} and \texttt{rdfs:range} predicates. Consequently, in this work, we only report Sem@$K$[ext] results for Codex-S and Codex-M. The statistics of the datasets Codex-S, Codex-M and WN18RR used in our experiments are provided in Table~\ref{tab:kg-ext}.

\textbf{WN18RR}~\cite{conve} originates from WN18, which is a subset of the WordNet KG~\cite{wordnet}. As for FB15K, Toutanova \textit{et al.}~\cite{toutanova} reported a huge test leakage in the original WN18 dataset. More specifically, 94\% of the train triples have inverse relations that are linked to test triples. Dettmers \textit{et al.}~\cite{conve} remove all inverse relations to propose WN18RR. In WN18RR, entities are nouns, verbs, and adjectives. Relations such as \texttt{hypernym} and \texttt{derivationally related from} hold between observed entities. Such relations are not linked to any \texttt{rdfs:domain} or \texttt{rdfs:range} predicates. Besides, relations such as \texttt{derivationally related from} can contain nouns, verbs or adjectives as both head or tail. It follows that it is not possible to infer any expected entity type for relations -- in this case the word qualifier. This is why WN18RR is used in the extensional setting.

\textbf{Codex-S}~\cite{codex} and \textbf{Codex-M}~\cite{codex} are datasets extracted from Wikidata and Wikipedia. They cover a wider scope and purposely contain harder facts than most KGs~\cite{codex}. Consequently, these datasets prove to be more challenging for the link prediction task. Compared to WN18RR, Codex-S and Codex-M contain entity types, relation descriptions and Wikipedia page extracts. Nonetheless, Wikidata does not contain \texttt{rdfs:domain} or \texttt{rdfs:range} predicates. The property constraints present in Wikidata are harder to manipulate than the \texttt{rdfs:domain} and \texttt{rdfs:range} clauses found in DBpedia. As such, in our experiments Codex-S and Codex-M are used in the extensional setting. Codex-S and Codex-M initially come with already generated hard negatives. In our experiments, we do not directly use these provided negative triples. Instead, we use the same Uniform Random Negative Sampling schema as for other datasets.

\begin{table}
\centering
\caption{Statistics of the schemaless KGs used in the experiments}\label{tab:kg-ext}
\begin{tabular}{lrrrrrr}
\toprule 
Dataset & \multicolumn{1}{c}{$|\mathcal{E}|$} & \multicolumn{1}{c}{$|\mathcal{R}|$} & \multicolumn{1}{c}{$|\mathcal{T}_{train}|$} & \multicolumn{1}{c}{$|\mathcal{T}_{valid}|$} & \multicolumn{1}{c}{$|\mathcal{T}_{test}|$}\\
\midrule
Codex-S & 2,034 & 42 & 32,888 & 1,827 & 1,828 \\
Codex-M  & 17,050 & 51 & 185,584 & 10,310 & 10,311 \\
WN18RR & 40,943 & 11 & 86,834 & 3,034 & 3,134  \\
\bottomrule
\end{tabular}
\end{table}

\subsection{Baseline models}
In this work, the semantic awareness of the most popular semantically agnostic KGEMs is analyzed. More specifically, the translational models TransE~\cite{transe} and TransH~\cite{transh}, the semantic matching models DistMult~\cite{distmult}, ComplEx~\cite{complex} and SimplE~\cite{simple}, and the convolutional models ConvE~\cite{conve}, ConvKB~\cite{convkb}, R-GCN~\cite{rgcn} and CompGCN~\cite{compgcn} are considered. Note that in the analysis of the results in Section~\ref{results}, a distinction will be made between pure convolutional KGEMs (ConvE, ConvKB) and GNNs (R-GCN, CompGCN). Although the latter have convolutional layers, they are able to capture long-range interactions between entities due to their ability to consider k-hop neighborhoods.
The characteristics of the models used in our experiments are mentioned hereinafter and summarized in Table~\ref{tab-kgems}.

\begin{table}
\centering
\caption{Summary of the KGEMs used in the experiments}\label{tab-kgems}
\begin{tabular}{clcc}
\toprule
Model Family&Model&\multicolumn{1}{c}{Scoring Function}&
\multicolumn{1}{c}{Loss Function}\\
\midrule
\multirow{2}*{Geometric}&TransE     &$\left\|\mathbf{e}_h+\mathbf{e}_r-\mathbf{e}_t\right\|_p$ & Pairwise Hinge\\
&TransH      &$\left\|\mathbf{e}_{h_{\perp}}+\mathbf{d}_{r}-\mathbf{e}_{t_{\perp}}\right\|_p$ & Pairwise Hinge\\[7pt]
&DistMult     &$\left\langle\mathbf{e}_{h}, \mathbf{W}_{r}, \mathbf{e}_{t}\right\rangle$ & Pairwise Hinge\\
Semantic Matching& ComplEx      &$\operatorname{Re}\left(\mathbf{e}_{h} \odot \mathbf{e}_{r} \odot \overline{\mathbf{e}}_{t}\right)$ & Pointwise Logistic\\
&SimplE      &$\frac{1}{2}{\left(\left\langle\mathbf{e}_{h}^{h}, \mathbf{e}_{r}, \mathbf{e}_{t}^{t}\right\rangle + \left\langle\mathbf{e}_{h}^{t}, \mathbf{e}_{r}^{-1}, \mathbf{e}_{t}^{h}\right\rangle\right)}$ & Pointwise Logistic\\[7pt]
\multirow{4}*{Convolutional}&ConvE     &${g\left(\operatorname{vec}\left(g\left({\operatorname{concat}}\left(\widehat{\mathbf{e}}_h, \widehat{\mathbf{e}}_r\right) * \mathbf{\omega}\right)\right) \mathbf{W}\right) \cdot \mathbf{e}_t}$ & Binary Cross-Entropy\\
&ConvKB      &$\operatorname{concat}\left(g\left(\left[\mathbf{e}_h, \mathbf{e}_r, \mathbf{e}_t\right] * \mathbf{\omega}\right)\right) \cdot \mathbf{w}$ & Pointwise Logistic\\
&R-GCN      &DistMult decoder & Binary Cross-Entropy\\
&CompGCN      &ConvE decoder & Binary Cross-Entropy\\
\bottomrule
\end{tabular}
\end{table}

\textbf{TransE}~\cite{transe} is the earliest translational model. It learns representations of entities and relations such that for a triple $(h,r,t)$, $\mathbf{e}_{h}+\mathbf{e}_{r} \approx \mathbf{e}_{t}$, where $\mathbf{e}_{h}$, $\mathbf{e}_{r}$ and $\mathbf{e}_{t}$ are the head, relation and tail embeddings, respectively. The scoring function is 
\begin{equation}\label{eq:transe}
f(h,r,t) = d(\mathbf{e}_{h}+\mathbf{e}_{r}-\mathbf{e}_{t})
\end{equation}
where $d$ is a distance function, usually the $L1$ or $L2$ norm.

\textbf{TransH}~\cite{transh} is an extension of TransE. It allows entities to have distinct representations when involved in different relations. Specifically, $\mathbf{e}_{h}$ and $\mathbf{e}_{t}$ are projected into relation-specific hyperplanes with projection matrices $\mathbf{w}_{r}$. If $(h,r,t)$ holds, the projected entities $\mathbf{e}_{h_{\perp}} = \mathbf{e}_{h} - \mathbf{w}_{r}^{T}\mathbf{e}_{h}\mathbf{w}_{r}$ and $\mathbf{e}_{t_{\perp}} = \mathbf{e}_{t} - \mathbf{w}_{r}^{T}\mathbf{e}_{t}\mathbf{w}_{r}$ are expected to be linked by the relation-specific translation vector $\mathbf{d}_{r}$. Thus, the scoring function is
\begin{equation}\label{eq:transh}
f(h,r,t) = d(\mathbf{e}_{h_{\perp}}+\mathbf{d}_{r}-\mathbf{e}_{t_{\perp}})
\end{equation}
TransH often showcases better performance than TransE with only slightly more parameters~\cite{transh}.

\textbf{DistMult}~\cite{distmult} is a semantic matching model. It is characterized as such because it uses a similarity-based scoring function and matches the latent semantics of entities and relations by leveraging their vector space representations. More specifically, DistMult is a bilinear diagonal model that uses a trilinear dot product as its scoring function:
\begin{equation}\label{eq:distmult}
f(h,r,t) =\left\langle\mathbf{e}_{h}, \mathbf{W}_{r}, \mathbf{e}_{t}\right\rangle
\end{equation}
It is similar to RESCAL~\cite{rescal} -- the very first semantic matching model -- but restricts relation matrices $\mathbf{W}_{r} \in \mathbb{R}^{d \times d}$ to be diagonal. 

\textbf{ComplEx}~\cite{complex} is also a semantic matching model. It extends DistMult by using complex-valued vectors to represent entities and relations: $\mathbf{e}_{h}, \mathbf{e}_{r}, \mathbf{e}_{t} \in \mathbb{C}^{d}$. As a result, ComplEx is better able to model antisymmetric relations than DistMult~\cite{rotate}. Its scoring function uses the Hadamard product:
\begin{equation}\label{eq:complex}
f(h,r,t)=\operatorname{Re}\left(\mathbf{e}_{h} \odot \mathbf{e}_{r} \odot \overline{\mathbf{e}}_{t}\right)
\end{equation}
where $\overline{\mathbf{e}}_{t}$ denotes the conjugate of $\mathbf{e}_{t}$.

\textbf{SimplE}~\cite{simple} models each fact in both a direct and an inverse form. To do so, an entity $e$ is simultaneously represented by two vectors $\mathbf{e}^h, \mathbf{e}^t \in \mathbb{R}^{d}$. Depending on whether $e$ appears as head or tail in a given triple, either $\mathbf{e}^h$ or $\mathbf{e}^t$ is used. Consequently, the two entity representations $\mathbf{e}^h$ and $\mathbf{e}^t$ are learned independently. Likewise, each relation $r$ comes with a direct and an inverse vectors $\mathbf{e}_{r}$ and $\mathbf{e}_{r^{-1}}$. 
The scoring function reflects the interaction between all the aforementioned entity and relation embeddings: 
\begin{equation}\label{eq:simple}
f(h,r,t) = \frac{1}{2}(\left\langle\mathbf{e}^h_{h}, \mathbf{e}_{r}, \mathbf{e}^t_{t}\right\rangle + \left\langle\mathbf{e}^t_{h}, \mathbf{e}_{r^{-1}}, \mathbf{e}^h_{t}\right\rangle)
\end{equation}

\textbf{ConvE}~\cite{conve} first reshapes entity and relation embeddings and then concatenates them into a 2D matrix [h; r]. To model the interactions between entities and relations, ConvE subsequently uses 2D convolution over embeddings and layers of nonlinear features. The output is ultimately scored against the tail embedding $t$ using the dot product. More precisely, the following scoring function is used:
\begin{equation}\label{eq:conve}
f(h,r,t) = {g\left(\operatorname{vec}\left(g\left({\operatorname{concat}}\left(\widehat{\mathbf{e}}_h, \widehat{\mathbf{e}}_r\right) * \mathbf{\omega}\right)\right) \mathbf{W}\right) \cdot \mathbf{e}_t}
\end{equation}
where $g$ denotes a non-linear function, $\operatorname{vec}$ is the vectorization operation reshaping a tensor into a vector, $\operatorname{concat}$ is the concatenation operator, $*$ and $.$ denote a convolution and a dot product, respectively, $\widehat{\mathbf{e}}$ denotes a 2D reshaping of $e$ and $\omega$ is the set of convolutional filters.

\textbf{ConvKB}~\cite{convkb} also represents entities and relations as same-sized vectors. However, ConvKB does not reshape the embeddings of entities and relations. Plus, ConvKB also considers the tail embedding in the concatenation operation, thus obtaining the 2D matrix [h; r; t] after concatenation. Convolution by a set $\omega$ of $T$ filters of shape $1*3$ is applied on this input. The resulting $T*3$ feature map then passes through a dense layer with one neuron and a weight matrix $W$. Finally, the following scoring function assesses the plausibility of a given triple $(h,r,t)$:
\begin{equation}\label{eq:convkb}
f(h,r,t) = \operatorname{concat}\left(g\left(\left[\mathbf{e}_h, \mathbf{e}_r, \mathbf{e}_t\right] * \mathbf{\omega}\right)\right) \cdot \mathbf{w}
\end{equation}
It has been claimed that the concatenation of a set of feature maps generated by convolution should increase the learning ability of latent features compared to ConvE~\cite{ji2021}. However, recent works point out the evaluation procedure used in the original implementation of ConvKB~\cite{sacn,sun2020}, which may result in overly optimistic results on the benchmark datasets FB15K237 and WN18RR. Nonetheless, due to the popularity of this model, we choose to include it in our experiments.

\textbf{R-GCN}~\cite{rgcn} extends the idea of applying graph convolutional networks (GCNs) to multi-relational data. R-GCN operates on local graph neighborhoods and applies a convolution operation to the neighbors of each entity. By aggregating the messages coming from all the neighbors of an entity, the embedding of the latter is updated in accordance. Each entity thus has a hidden representation which directly depends on the hidden representations of its neighbors. This process of accumulating messages (\textit{i.e.} the hidden representations of neighboring entities) and aggregating them so as to update the hidden representation of the central node is performed for each layer of the R-GCN model. More formally, the hidden representation of the entity $i$ in the layer $(l+1)$ is defined as:
\begin{equation}\label{eq:rgcn}
h_i^{(l+1)}=\sigma\left(W_0^{(l)} h_i^{(l)}+\sum_{r \in R} \sum_{j \in N_i^r} \frac{1}{c_{i, r}} W_r^{(l)} h_j^{(l)}\right)
\end{equation}
\noindent where $\sigma(.)$ can be any element-wise activation function, $N_i^r$ denotes the set of neighboring entities of entity $i$ considering the relation $r$, $W_r^{(l)}$ is the relation-specific weight matrix for layer $l$ and $c_{i, r}$ is a normalization constant. To update entity hidden representation taking into account its previous state, a self-connection is also incorporated into the activation function $\sigma(.)$. Such layers can be stacked multiple times in order to better learn interactions and dependencies across several relational steps. However, as applying Eq.(\ref{eq:rgcn}) directly would dramatically increase the number of parameters for KGs having lots of different relation types, basis-decomposition and block-diagonal-decomposition are proposed in~\cite{rgcn} to reduce model parameter size and prevent overfitting. In the case of basis decomposition which is used in the original paper, the relation-specific weight matrix $W_r^{(l)}$ is decomposed into a linear combination of basis transformation $V_{b}^{(l)}$ with coefficients $a_{rb}^{(l)}$:
\begin{equation}\label{eq:basisdecompo-rgcn}
W_{r}^{(l)}=\sum_{b=1}^{B} a_{rb}^{(l)}V_{b}^{(l)}
\end{equation}

\textbf{CompGCN}~\cite{compgcn} improves over R-GCN by not only learning entity representations but also relation representations. Concretely, CompGCN performs a composition operation $\phi(.)$ over each edge in the neighborhood of a central node. The composed embeddings are subsequently convolved with direction-specific weight matrices $W_{O}$ and $W_{I}$ -- for original and inverse relations, respectively. In addition to the entity representation update, relation representations are also updated individually:
\begin{equation}\label{eq:basisdecompo-compgcn}
h_{r}^{(l+1)}=W_{rel}^{(l)}h_{r}^{(l)}
\end{equation}
\noindent where $W_{rel}$ is a learnable transformation matrix which projects all the relations into the same embedding space as entities.

\subsection{Implementation details and hyperparameters}
For the sake of comparisons, MRR, Hits@K and Sem@$K$ all need to rely on the same code implementation. More specifically, for R-GCN\footnote{\url{https://github.com/toooooodo/RGCN-LinkPrediction}} and CompGCN\footnote{\url{https://github.com/malllabiisc/CompGCN}}, existing implementations were reused and Sem@$K$ values were calculated on the trained models. Other KGEMs were implemented in PyTorch. To avoid time-consuming hyperparameter tuning, we took inspiration from the hyperparameters provided by LibKGE\footnote{\url{https://github.com/uma-pi1/kge}} for Codex-S, Codex-M, WN18RR and FB15K237. However, LibKGE does not benchmark all the datasets and models used in our experiments. For such models, we stick to the hyperparameters provided by the original authors, when available. For the models with no reported best hyperparameters, as well as for the remaining datasets used in the experiments -- \textit{i.e.} DB93K, YAGO3-37K and YAGO4-19K -- different combinations of hyperparameters were manually tried. We first trained our KGEMs for $1,000$ epochs, then noticed the best achieved results were found around epoch $400$ or below. Consequently, we stick to a maximum of $400$ epochs of training as in LibKGE (except R-GCN which is trained during $4,000$ epochs due to lower convergence to the best achieved results). For each positive triple in the training set, one corresponding negative triple is generated. To ensure fair comparisons between our models, we stick with Uniform Random Negative Sampling~\cite{transe}. The chosen hyperparameters leading to the best performance on the validation dataset are provided in Appendix (see Appendix~\ref{appendix:hyperparameters}, Table~\ref{tab:hyperparams}).
Recall that the objective of this work is to perform a fair and insightful assessment of the semantic awareness of KGEMs. As such, the intended purpose was not to reach optimal performance in terms of rank-based metrics. Instead, the objective is to identify a set of hyperparameters that provides satisfying and stable performance, and then study how Sem@$K$ behaves, both for a fixed epoch -- \textit{e.g.} for the best epoch in terms of MRR -- and dynamically w.r.t. the number of epochs.

\section{Results}
In the following, we perform an extensive analysis of the results obtained using the aforedescribed KGEMs and KGs. For the sake of clarity, the complete range of tables and plots are placed in Appendix~\ref{appendix:results} and \ref{appendix:plots}, respectively. When necessary to support our claim, some of them are duplicated in the body text.
\label{results}
\subsection{Semantic awareness of KGEMs}
This section draws on the Sem@$K$ values (see Tables~\ref{tab:schema-defined-results-compiled} and \ref{tab:schemaless-results-compiled}) achieved at the best epoch in terms of MRR to provide an analysis of the semantic awareness of state-of-the-art KGEMs. In other words, for such models we only consider a snapshot of their best epochs in terms of rank-based metrics.

A major finding is that models performing well with respect to rank-based metrics are not necessarily the most competitive when it comes to their semantic capabilities. On YAGO3-37K (see Table~\ref{tab:schema-defined-results-yago3-body}) for instance, ConvE showcases impressive MRR and Hits@K values. However, it is far from being the best KGEM in terms of Sem@$K$, as it is outperformed by CompGCN, R-GCN and all translational models.

\begin{table}
\centering
\caption{Results on YAGO3-37K. Bold fonts and gray cells denote the best achieved results and the worst achieved results among the models reported in the table, respectively. Full results are available in Appendix~\ref{appendix:results}, Table~\ref{tab:schema-defined-results-compiled}}
\label{tab:schema-defined-results-yago3-body}
        \begin{adjustbox}{width={\textwidth},totalheight={\textheight},keepaspectratio}%
        \begin{tabular}{cc|cccc|ccc|ccc|ccc}
                \toprule
            & & \multicolumn{4}{c|}{Rank-based} & \multicolumn{3}{c|}{Sem@$K$[base]} & \multicolumn{3}{c|}{Sem@$K$[wup]} & \multicolumn{3}{c}{Sem@$K$[ext]} \\
            Model Family & Model & MRR & H@1 & H@3 & H@10 & 
                S@1 & S@3 & S@10 &
                S@1 & S@3 & S@10 &
                S@1 & S@3 & S@10 \\
            \midrule
            \multirow{2}*{Geometric}& TransE &
$.184$&$.080$&$.198$&$.408$& $.989$&$.988$&$.988$& $.993$&$.994$&$.995$& $.897$&$.904$&$.911$\\
            & TransH &
$.187$&$.091$&$.199$&$.415$& $.995$&$.993$&$.993$& $.995$&$.997$&$.997$& $.901$&$.911$&$.925$ \\
            \midrule
            \multirow{3}*{Convolutional}& ConvE &
$\mathbf{.493}$&$\mathbf{.350}$&$\mathbf{.578}$&$\mathbf{.775}$& \cellcolor{lgray}$.933$&\cellcolor{lgray}$.923$&\cellcolor{lgray}$.910$& \cellcolor{lgray}$.977$&\cellcolor{lgray}$.977$&\cellcolor{lgray}$.975$& \cellcolor{lgray}$.893$&\cellcolor{lgray}$.879$&\cellcolor{lgray}$.871$\\
            & R-GCN &
$.115$&$.046$&$.110$&$.254$& $.993$&$.994$&$.993$& $.995$&$.996$&$.997$& $.933$&$.932$&$.928$ \\
            & CompGCN &
$.399$&$.269$&$.464$&$.663$& $.998$&$.997$&$.996$& $.999$&$.999$&$.998$& $.998$&$.994$&$.982$ \\
\bottomrule
\vspace{0.5mm}
        \end{tabular}
            \end{adjustbox}
\end{table}

\begin{figure}[htbp]
  \centering
  \begin{subfigure}{0.45\textwidth}
    \centering
    \includegraphics[width=\linewidth]{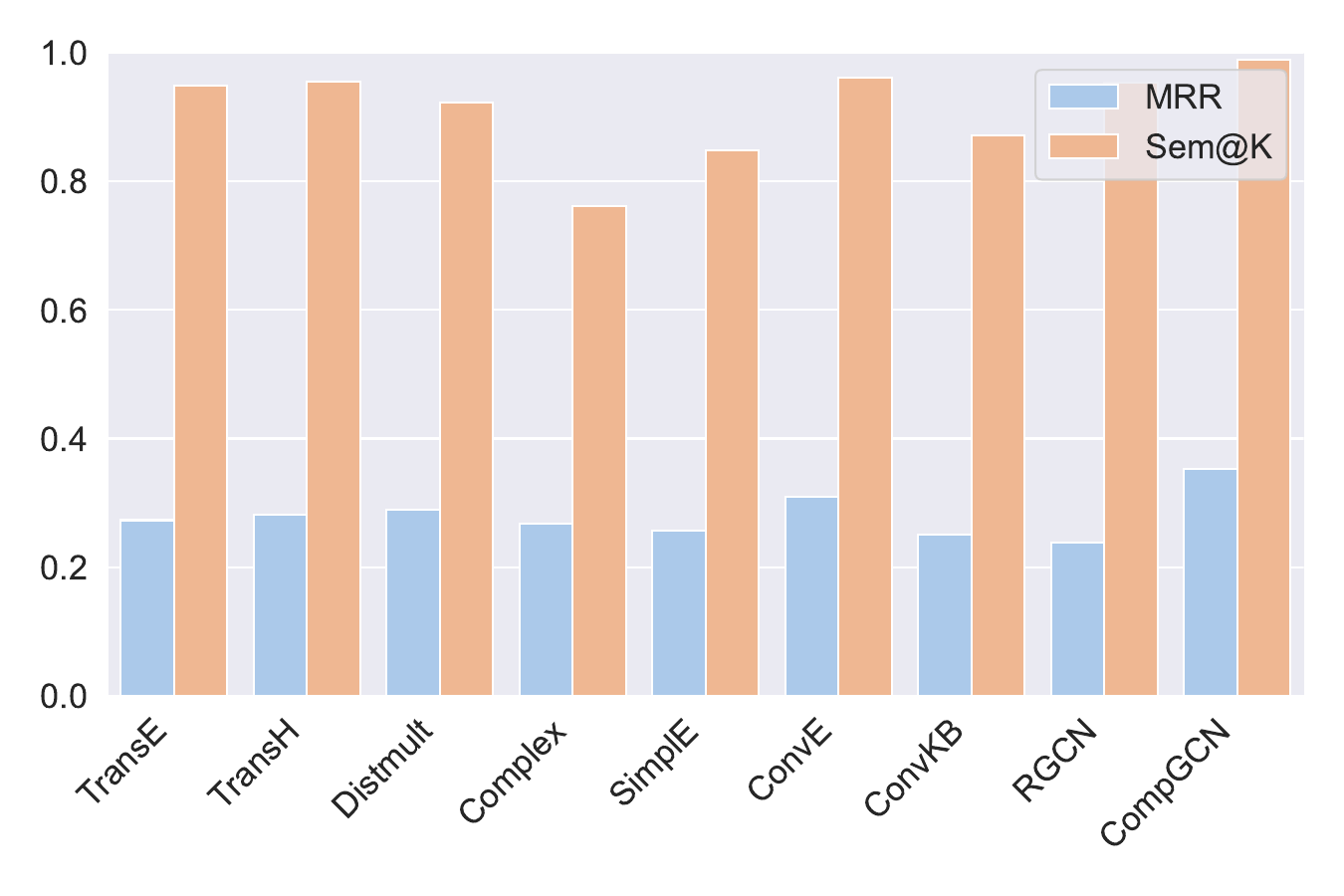}
    \caption{FB15K237-ET}
  \end{subfigure}
  \hfill
  \begin{subfigure}{0.45\textwidth}
    \centering
    \includegraphics[width=\linewidth]{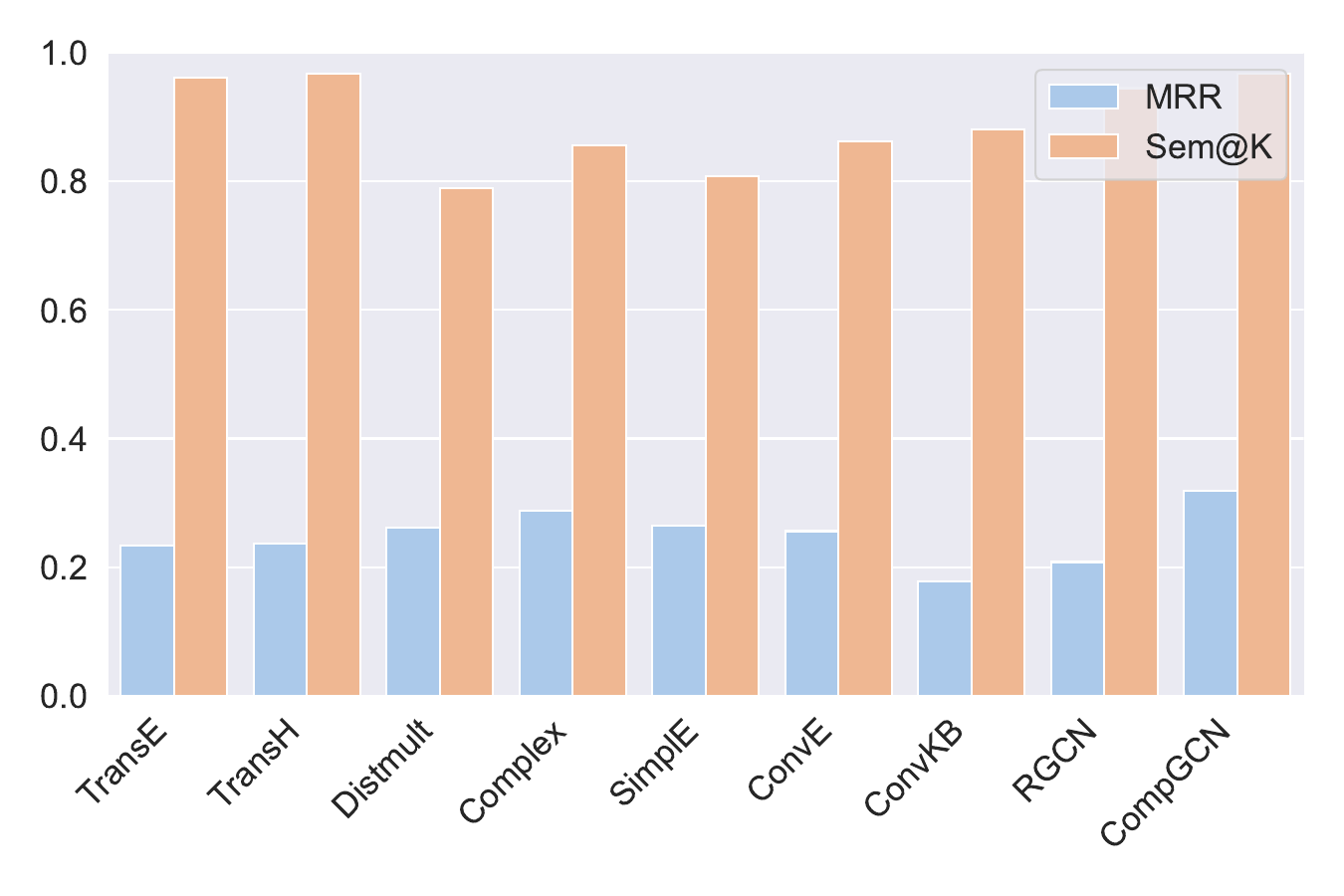}
    \caption{DB93K}
  \end{subfigure}

  \begin{subfigure}{0.45\textwidth}
    \centering
    \includegraphics[width=\linewidth]{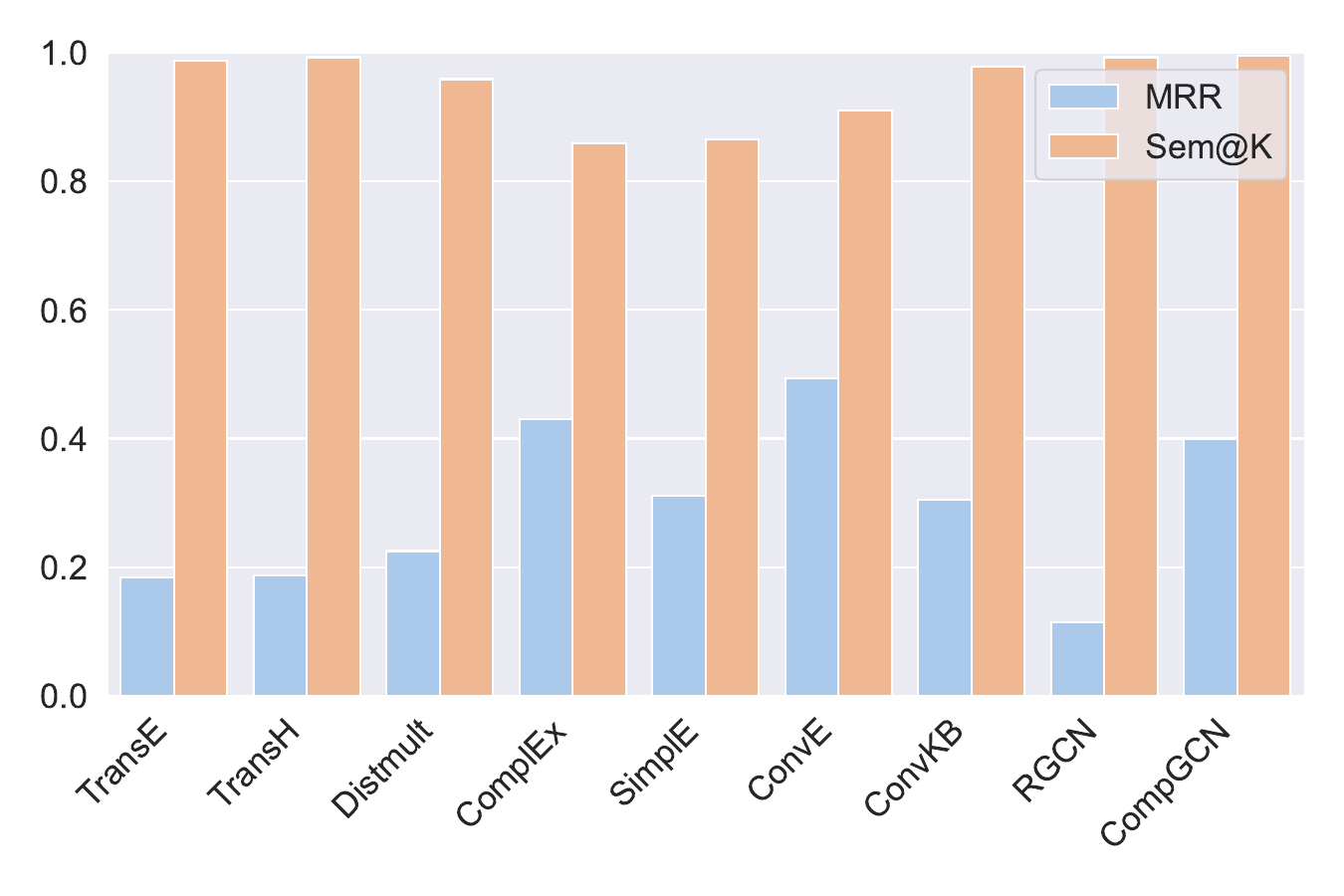}
    \caption{YAGO3-37K}
  \end{subfigure}
  \hfill
  \begin{subfigure}{0.45\textwidth}
    \centering
    \includegraphics[width=\linewidth]{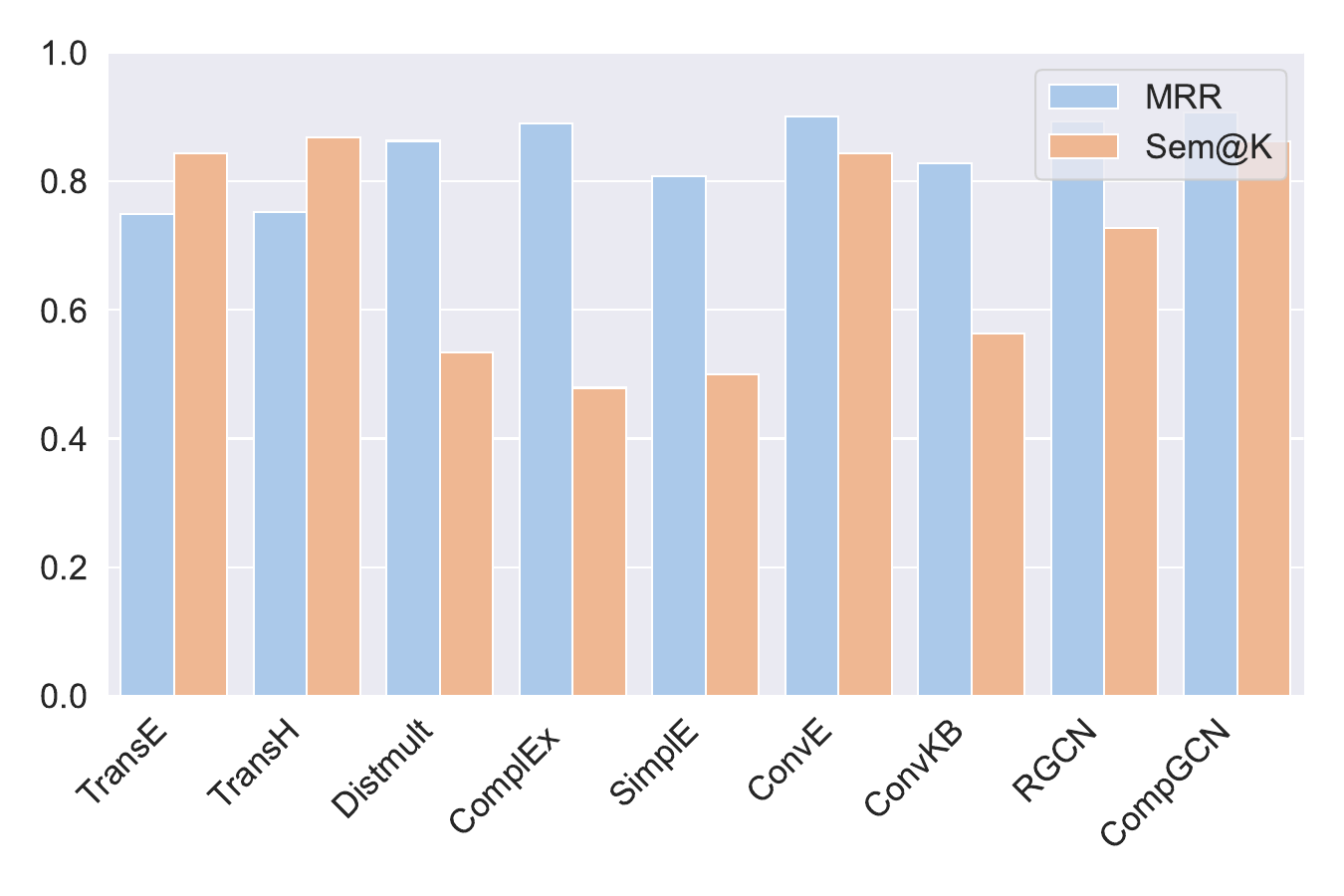}
    \caption{YAGO4-19K}
  \end{subfigure}
  \caption{MRR and Sem@10 results achieved at the best epoch in terms of MRR for each model and on each schema-defined dataset}\label{fig:barplots-schemadefined}
\end{figure}

\begin{figure}[htbp]
  \centering
  \begin{subfigure}{0.32\textwidth}
    \centering
    \includegraphics[width=\linewidth]{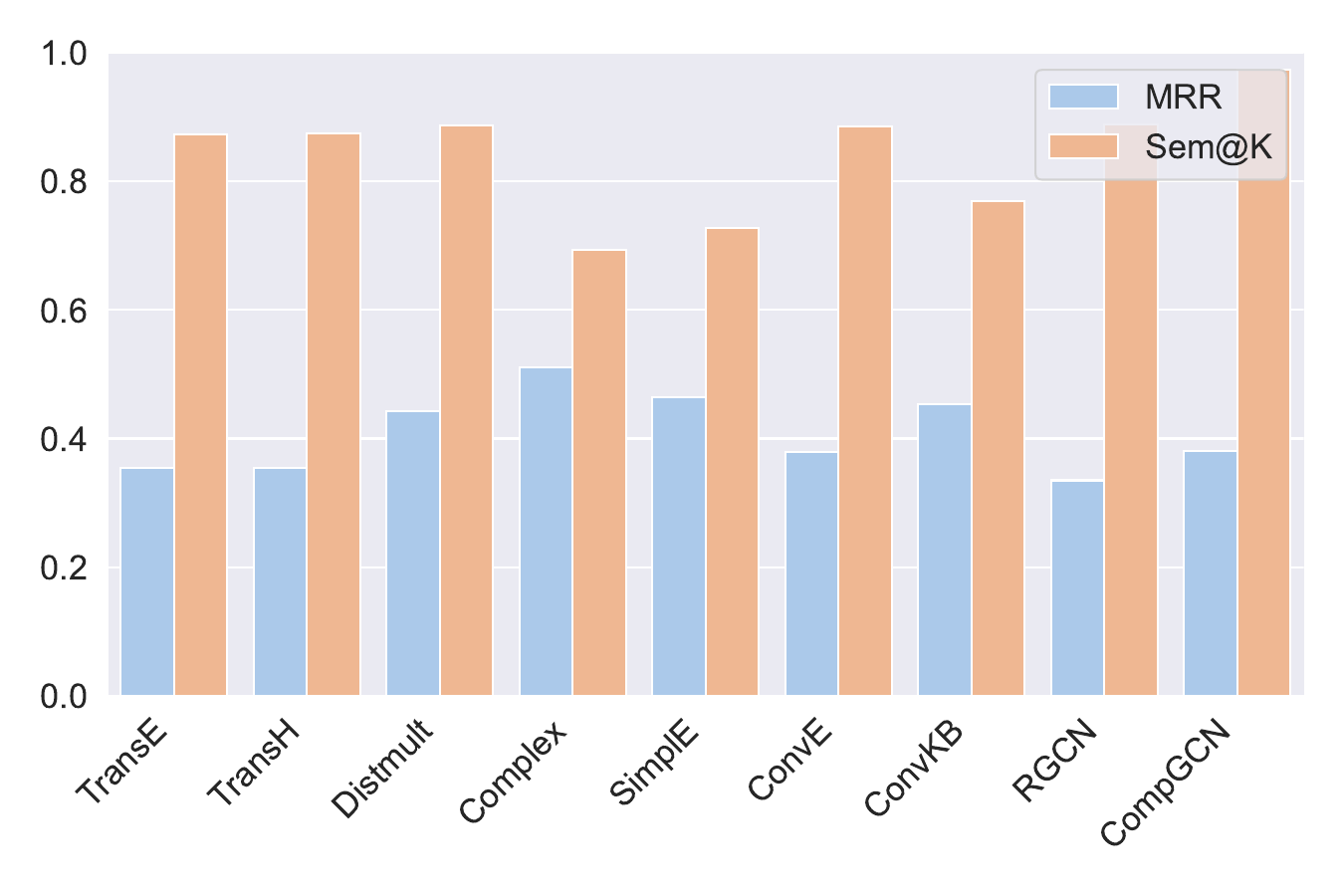}
    \caption{Codex-S}
  \end{subfigure}
  \begin{subfigure}{0.32\textwidth}
    \centering
    \includegraphics[width=\linewidth]{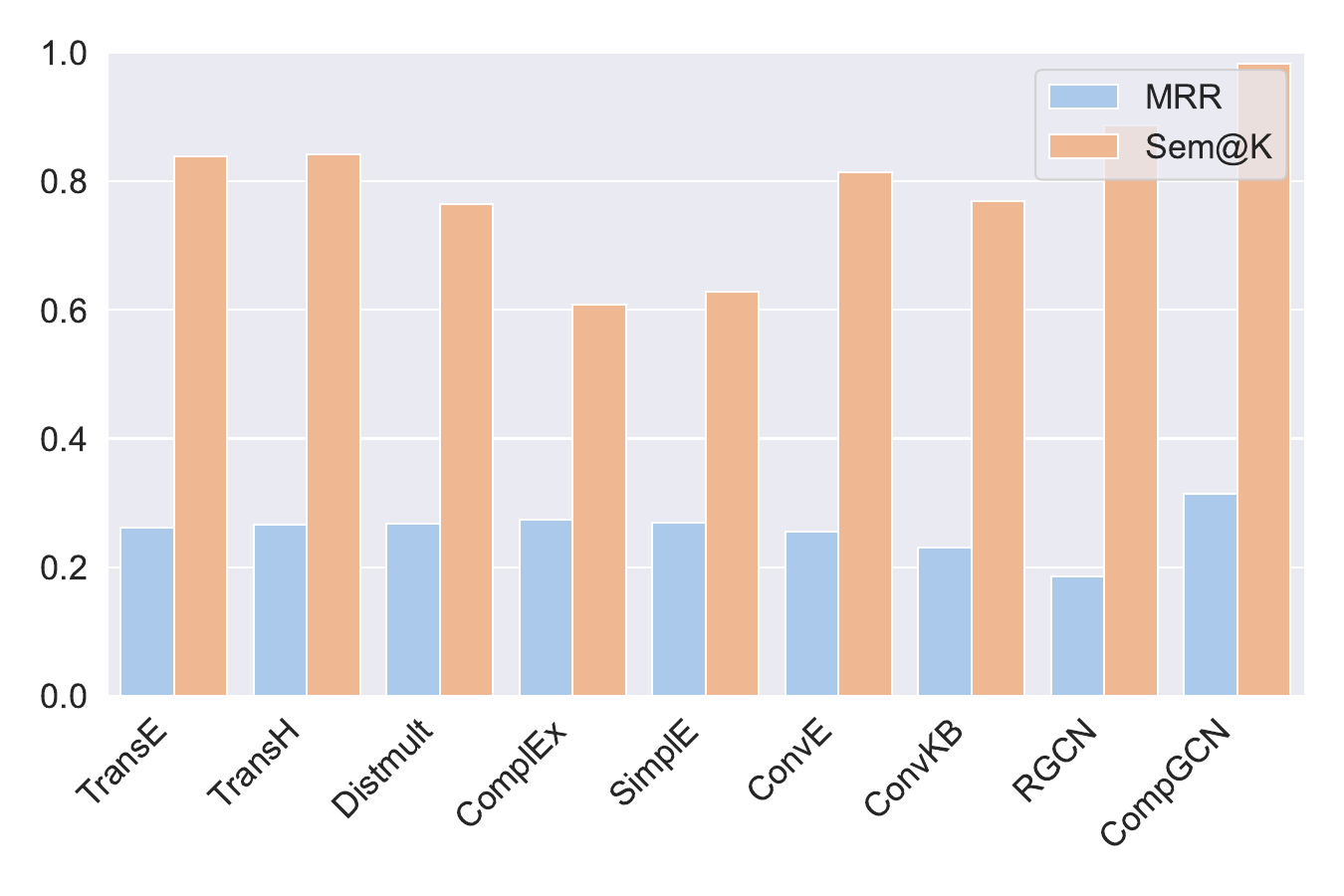}
    \caption{Codex-M}
  \end{subfigure}
  \begin{subfigure}{0.32\textwidth}
    \centering
    \includegraphics[width=\linewidth]{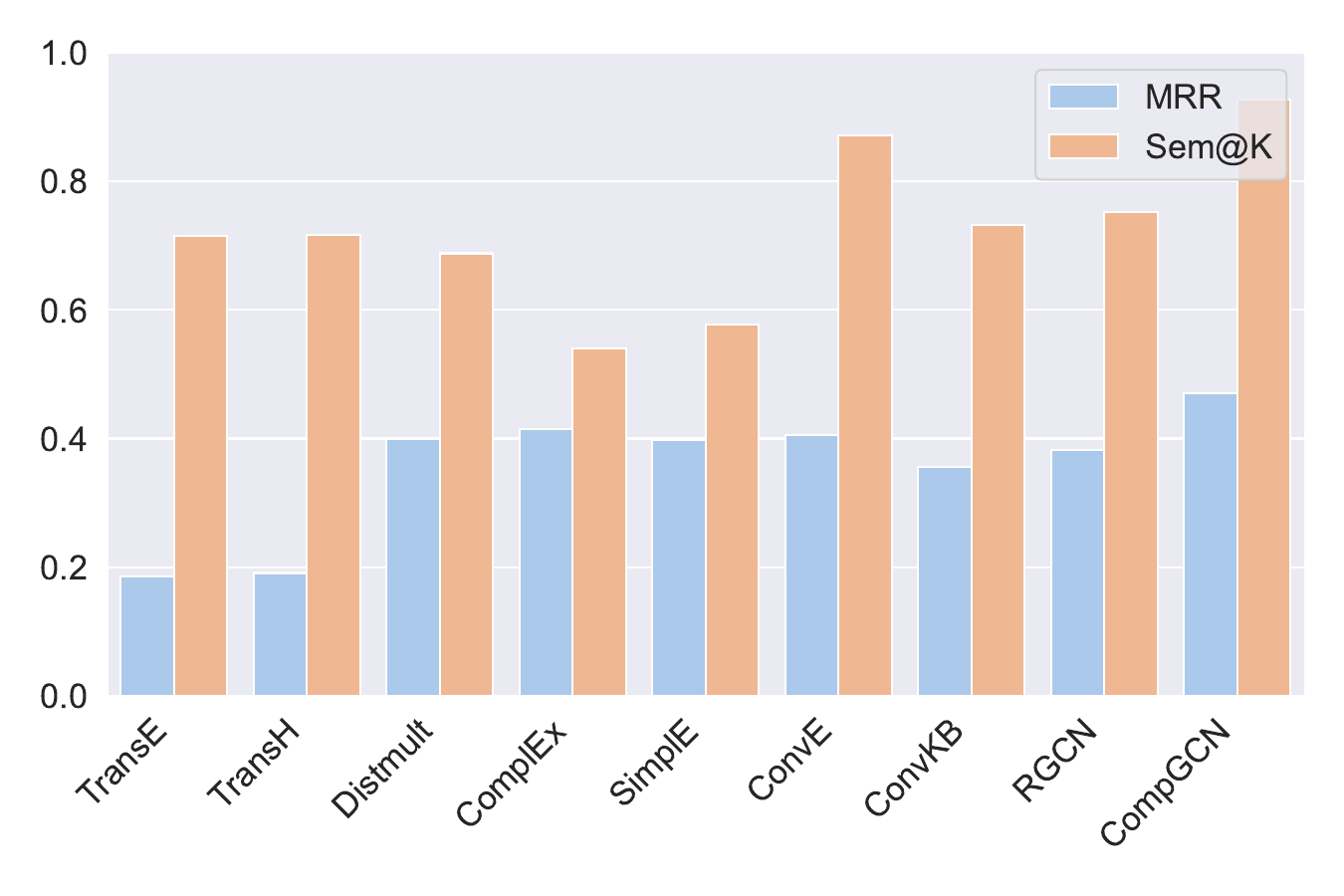}
    \caption{WN18RR}
  \end{subfigure}
  \caption{MRR and Sem@10 results achieved at the best epoch in terms of MRR for each model and on each schemaless dataset}\label{fig:barplots-schemaless}
\end{figure}

\begin{figure}
    \centering
    \begin{subfigure}[c]{0.87\textwidth}
        \centering
        \includegraphics[width=\textwidth]{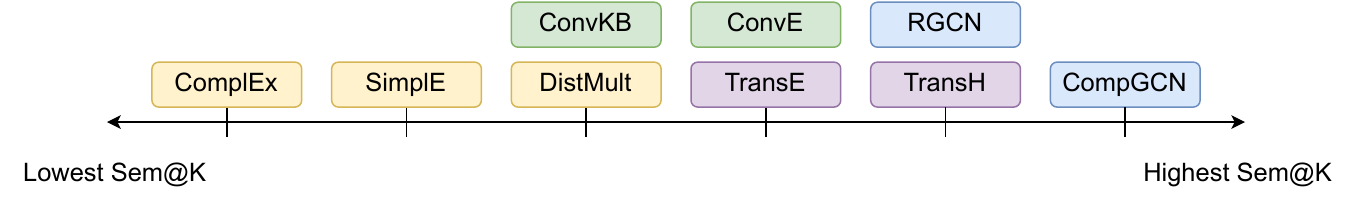}
        \caption{FB15K237-ET}
        \label{subfig:yardstick-fb15k237}
        \vspace*{7mm}
    \end{subfigure}
    \begin{subfigure}[c]{0.87\textwidth}
        \centering
        \includegraphics[width=\textwidth]{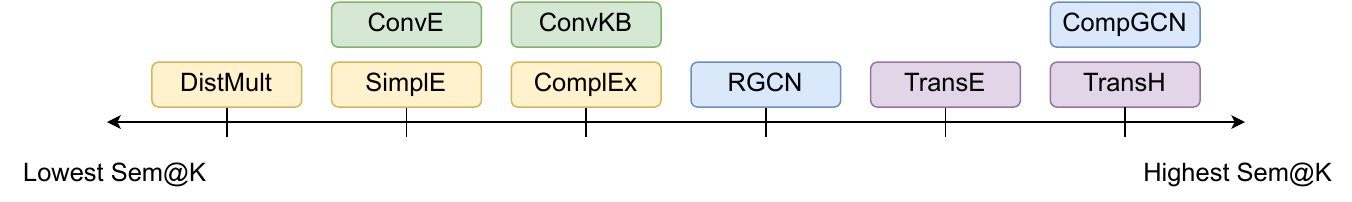}
        \caption{DB93K}
        \label{subfig:yardstick-db93k}
        \vspace*{7mm}
    \end{subfigure}
    \begin{subfigure}[c]{1.0\textwidth}
        \centering
        \includegraphics[width=\textwidth]{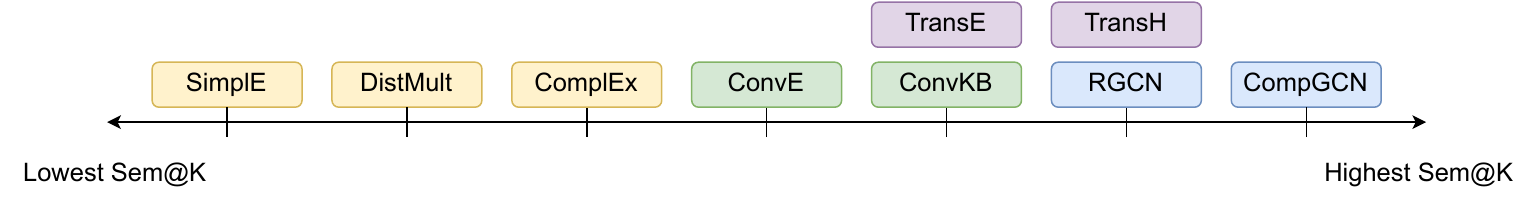}
        \caption{YAGO3-37K}
        \label{subfig:yardstick-yago3}
        \vspace*{7mm}
    \end{subfigure}
    \begin{subfigure}[c]{1.0\textwidth}
        \centering
        \includegraphics[width=\textwidth]{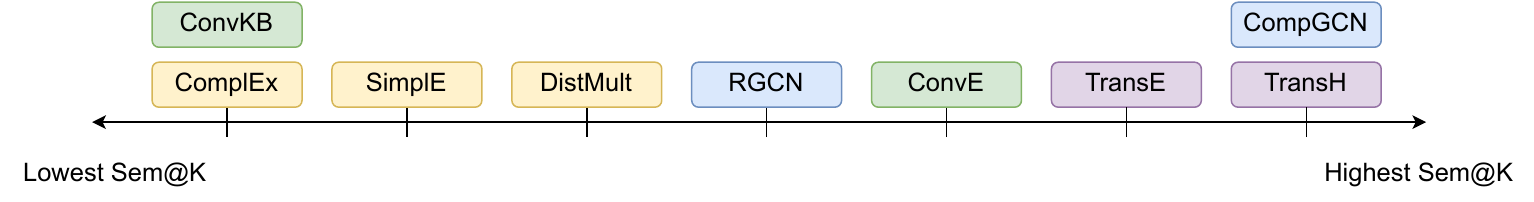}
        \caption{YAGO4-19K}
        \label{subfig:yardstick-yago4}
    \end{subfigure}
    \caption{Sem@$K$[base] comparisons between KGEMs on the 4 benchmarked schema-defined KGs. Colors indicate the family of models: blue, purple, green, and yellow cells denote GNNs, translational, convolutional, and semantic matching models, respectively. Regarding Sem@$K$, the relative hierarchy of models is consistent across KGs and we can clearly see that KGEMs are grouped by families of models}
    \label{fig:yardsticks-schema-defined}
\end{figure}

\begin{figure}
    \centering
    \begin{subfigure}[c]{0.875\textwidth}
        \centering
        \includegraphics[width=\textwidth]{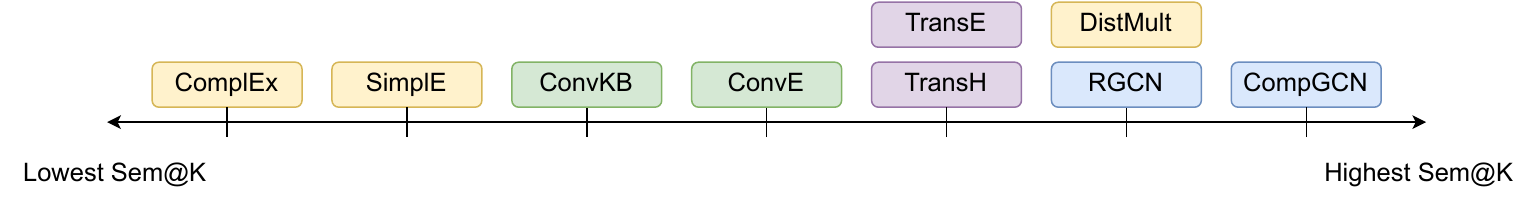}
        \caption{Codex-S}
        \label{subfig:yardstick-codex-s}
        \vspace*{7mm}
    \end{subfigure}
    \begin{subfigure}[c]{1.0\textwidth}
        \centering
        \includegraphics[width=\textwidth]{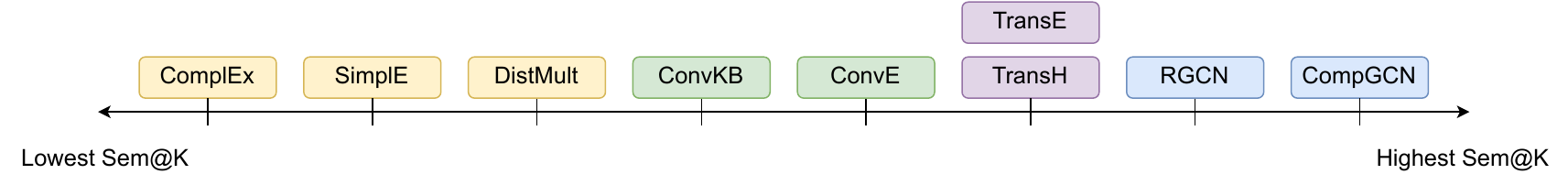}
        \caption{Codex-M}
        \label{subfig:yardstick-codex-m}
        \vspace*{7mm}
    \end{subfigure}
    \begin{subfigure}[c]{0.875\textwidth}
        \centering
        \includegraphics[width=\textwidth]{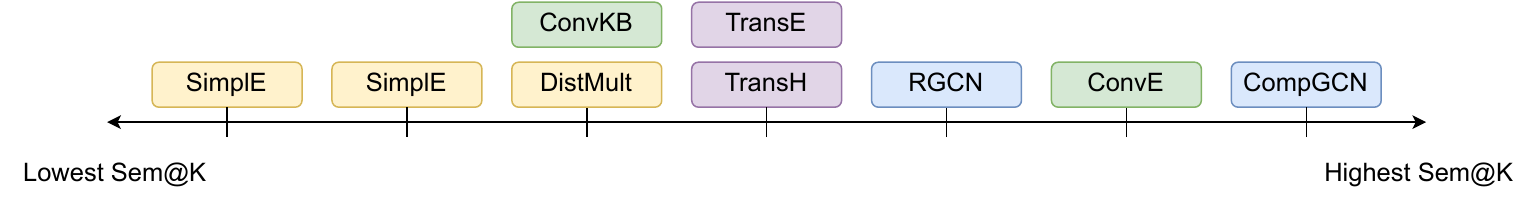}
        \caption{WN18RR}
        \label{subfig:yardstick-wn18rr}
    \end{subfigure}
    \caption{Sem@$K$[ext] comparisons between KGEMs on the 3 benchmarked schemaless KGs. Colors indicate the family of models: blue, purple, green, and yellow cells denote GNNs, translational, convolutional, and semantic matching models, respectively. Regarding Sem@$K$, the relative hierarchy of models is consistent across KGs and we can clearly see that KGEMs are grouped by families of models.}
    \label{fig:yardsticks-schemaless}
\end{figure}

From a coarse-grained viewpoint, conclusions about the relative superiority of models with the distinct consideration of rank-based metrics and semantic awareness can be generalized at the level of models families. For example, semantic matching models (DistMult, ComplEx, SimplE) globally achieve better MRR and Hits@$K$ values while their semantic capabilities are in most cases lower than the ones of translational models (TransE, TransH) -- see Table~\ref{tab:schema-defined-results-compiled} and Table~\ref{tab:schemaless-results-compiled} for detailed results w.r.t. rank-based and semantic-oriented metrics. A condensed view of the comparison between MRR and Sem@10 is also reported in Fig.~\ref{fig:barplots-schemadefined} and Fig.~\ref{fig:barplots-schemaless}. The respective hierarchies of such models for the benchmarked schema-defined and schemaless KGs are depicted in Fig.~\ref{fig:yardsticks-schema-defined} and Fig.~\ref{fig:yardsticks-schemaless}, respectively. It is clearly visible that KGEMs are grouped by family. In particular, GNNs and translational models showcase very promising semantic capabilities. GNNs are almost always the best regarding Sem@$K$[ext] values -- not only for schemaless KGs (Fig.~\ref{fig:yardsticks-schemaless}), but also for schema-defined KGs (see Table~\ref{tab:schema-defined-results-compiled} for full results, and Fig.~\ref{fig:barplots-schemadefined} for a quick glimpse). This means GNNs are more capable of predicting entities that have been observed as head (resp. tail) of a given relation. Translational models are very competitive in terms of Sem@$K$[base]. In Fig.~\ref{fig:yardsticks-schema-defined}, we clearly see that they consistently rank among the best performing models regarding Sem@$K$[base]. Interestingly, the semantic matching models DistMult, ComplEx and SimplE perform relatively poorly. This observation holds regardless of the nature of the KG, as they systematically rank among the worst performing models for schema-defined (Fig.~\ref{fig:yardsticks-schema-defined}) and schemaless (Fig.~\ref{fig:yardsticks-schemaless}) KGs.

Therefore, it seems that translational models are better able at recovering the semantics of entities and relations to properly predict entities that are in the domain (resp. range) of a given relation, while semantic matching models might sometimes be better at predicting entities already observed in the domain (resp. range) of a given relation (\textit{e.g.} DistMult reaches very high Sem@$K$[ext] values on Codex-S, as evidenced in Fig.~\ref{subfig:yardstick-codex-s}). In cases when translational and semantic matching models provide similar results in terms of rank-based metrics, the nature of the dataset at hand -- whether it is schema-defined or schemaless -- might thus strongly influence the final choice of a KGEM.

Interestingly, CompGCN which is by far the most recent and sophisticated model used in our experiments, outperforms all the other models in terms of semantic awareness as, with very limited exceptions, it is the best in terms of Sem@$K$[base], Sem@$K$[ext] and Sem@$K$[wup]. In addition, it should be noted that R-GCN provides satisfying results as well. Except in a very few cases (\textit{e.g.} outperformed by ComplEx on Codex-S and WN18RR in terms of Sem@1), R-GCN showcases better semantic awareness than semantic matching models. Most of the time, R-GCN also provides comparable or even higher semantic capabilities than translational models. In particular, Sem@$K$ values achieved with R-GCN are similar to the ones achieved with TransE and TransH (\textit{e.g.} on FB15K237-ET and YAGO3-37K, see Tables~\ref{tab:schema-defined-results-fb15k237} and \ref{tab:schema-defined-results-yago3}) while the latter models are actually outperformed by R-GCN in terms of Sem@$K$[ext]. It appears clearly on Codex-M and WN18RR (Tables~\ref{tab:schemaless-results-codex-m} and \ref{tab:schemaless-results-wn18rr}), although the conclusion holds for all datasets.

Our experimental results suggest that the structure of GNNs seems to be able to encode the latent semantics of the entities and relations of the graph. This ability may be attributed to the fact that, contrary to translational models which only model the local neighborhood of each triple, GNNs update entity embeddings (and relation embeddings in the case of CompGCN) based on the information found in the extended, h-hop neighborhood of the central node. While translational and semantic matching models treat each triple independently, GNNs model interactions between entities on a large range of semantic relations. It is likely that this extended neighborhood comprises signals or patterns that help the model infer the classes of entities, thus providing very promising semantic capabilities in all experimental conditions.

\subsection{Dynamic appraisal of KGEM semantic awareness}

\begin{figure}
    \centering
    \begin{subfigure}{0.45\textwidth}
        \centering
        \includegraphics[width=1.\linewidth]{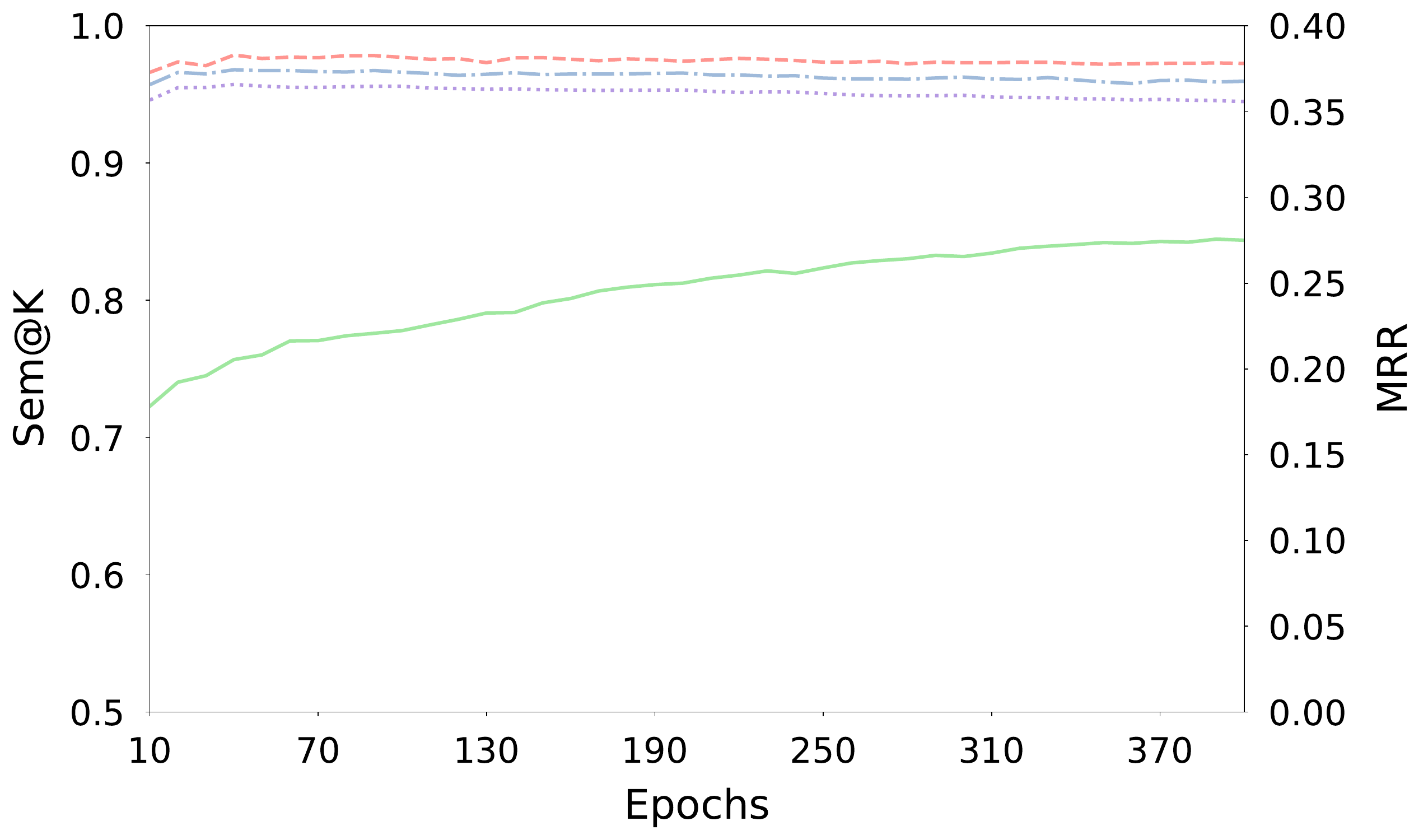}
        \caption{TransE -- FB15K237-ET}
        \label{subfig:fb-transe-body}
    \end{subfigure}
    \qquad
    \begin{subfigure}{0.45\textwidth}
        \centering
        \includegraphics[width=1.\linewidth]{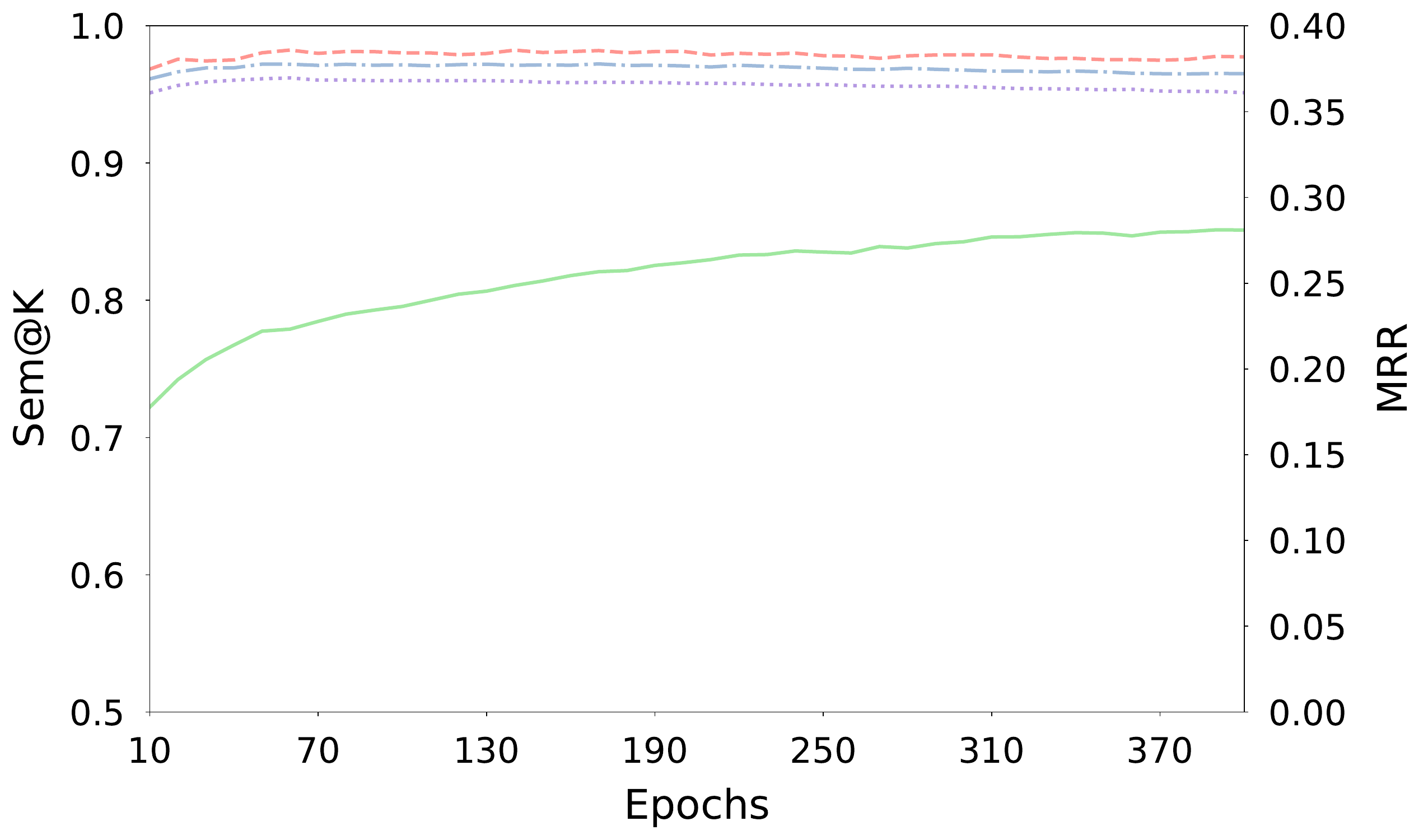}
        \caption{TransH -- FB15K237-ET}
        \label{subfig:fb-transh-body}
    \end{subfigure}\par\medskip
    
    \begin{subfigure}{0.45\textwidth}
        \centering
        \includegraphics[width=1.\linewidth]{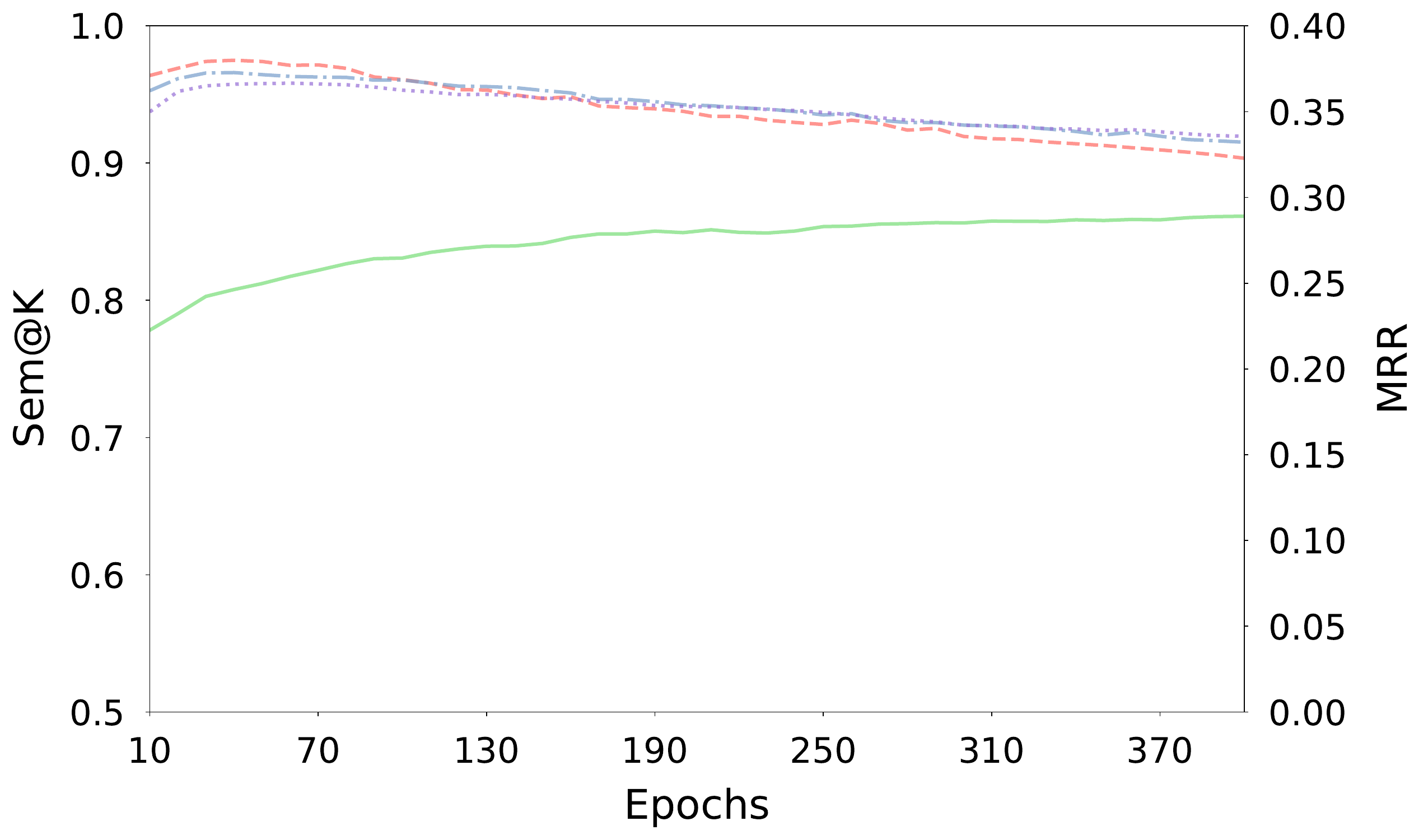}
        \caption{DistMult -- FB15K237-ET}
        \label{subfig:fb-distmult-body}
    \end{subfigure}
    \qquad
    \begin{subfigure}{0.45\textwidth}
        \centering
        \includegraphics[width=1.\linewidth]{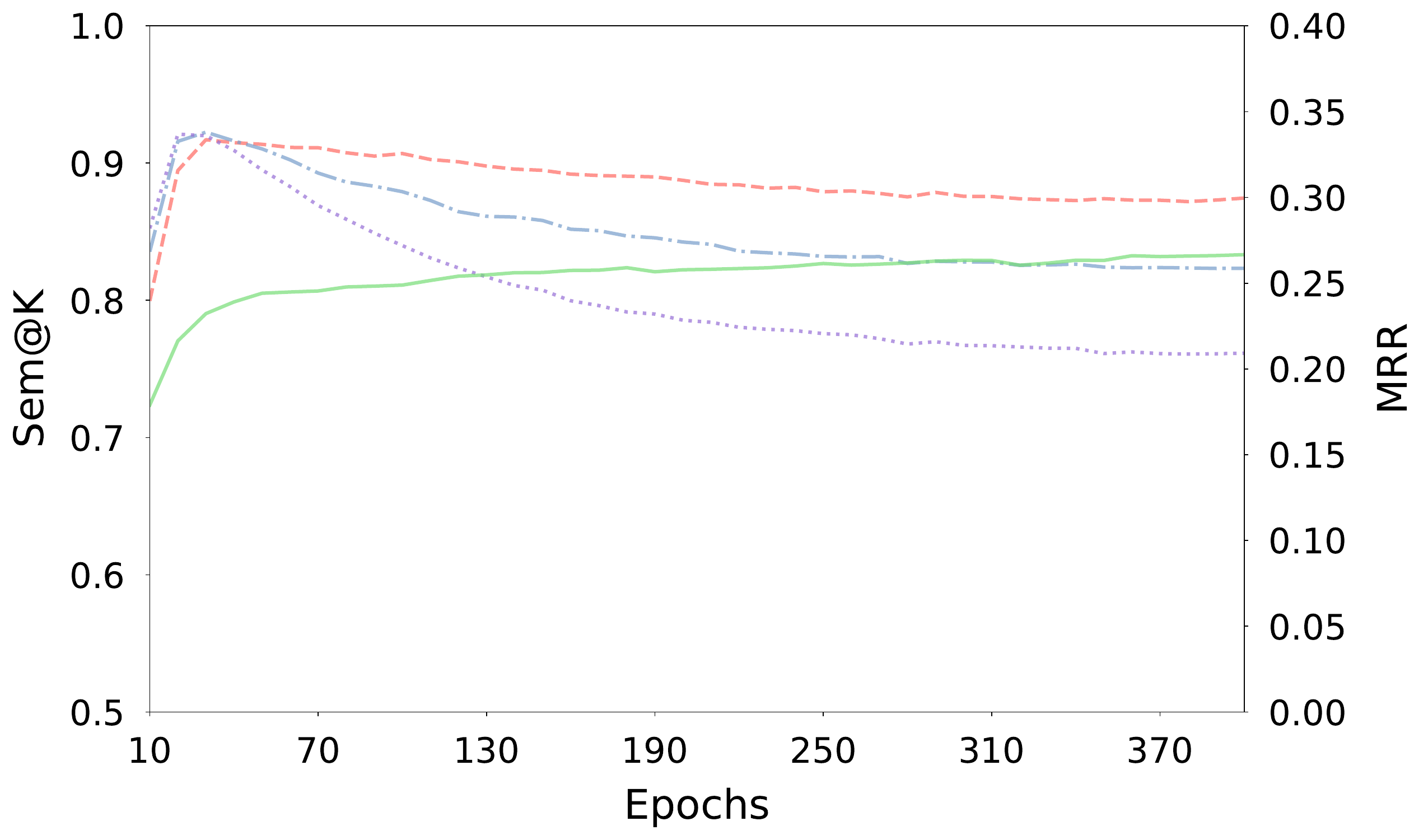}
        \caption{ComplEx -- FB15K237-ET}
        \label{subfig:fb-complex-body}
    \end{subfigure}
    \caption[]{Evolution of MRR (\raisebox{2pt}{\begin{tikzpicture}[scale=0.5]
        \draw[color=pastelgreen, solid, line width=1pt] (0,0) -- (0.7,0);
    \end{tikzpicture}}), Sem@1 (\raisebox{2pt}{\begin{tikzpicture}[scale=0.5]
        \draw[color=pastelred, dashed, line width=1pt] (0,0) -- (0.65,0);
    \end{tikzpicture}}), Sem@3 (\raisebox{2pt}{\begin{tikzpicture}[scale=0.5]
        \draw[color=darkpastelblue, dash dot, line width=1pt] (0,0) -- (0.7,0);
    \end{tikzpicture}}), and Sem@10 (\raisebox{2pt}{\begin{tikzpicture}[scale=0.5]
        \draw[color=darkpastelpurple, dotted, line width=1pt] (0,0) -- (0.7,0);
    \end{tikzpicture}}) for translational (TransE, TransH) and semantic matching models (DistMult, ComplEx) on FB15K237-ET}
    \label{fig:fb15-plots}
\end{figure}

For certain models, rank-based metrics performance and semantic capabilities improve jointly. For others, the enhancement of their performance in terms of rank-based metrics comes at the expense of their semantic awareness. Interestingly, trends emerge relatively to families of models. First, we observe that a trade-off exists for semantic matching models. Results are particularly striking on FB15K237-ET (see Fig.~\ref{fig:fb15-plots}), where it is obvious that after reaching the best Sem@$K$ values after a few epochs, Sem@$K$ values of DistMult and ComplEx quickly drop while MRR continues rising. Conversely, translational models are more robust to Sem@$K$ degradation throughout the epochs. Even though the best achieved Sem@$K$ values are also reported in the very first epochs, once these values are reached they remain stable for the remaining epochs of training. This might be due to the geometric nature of such KGEMs, which will organize the representation space so as to $h + r$ falls in a region of the space where neighboring entities of the ground-truth tail $t$ all are entities of the expected type. This is highly related to the block structure property, which is a common statistical pattern found in KGs~\cite{nickel2016}. It refers to the fact that entities can naturally belong to different groups (blocks), such that all the entities of a given group are linked to entities of another group through the same relationship. In this case, each group comprises entities of the same class. Translational models will naturally group entities of the same class in the same region of the representation space, as this is determined by the translation vector in that space.

\begin{figure}
    \centering
    \includegraphics[scale=0.5]{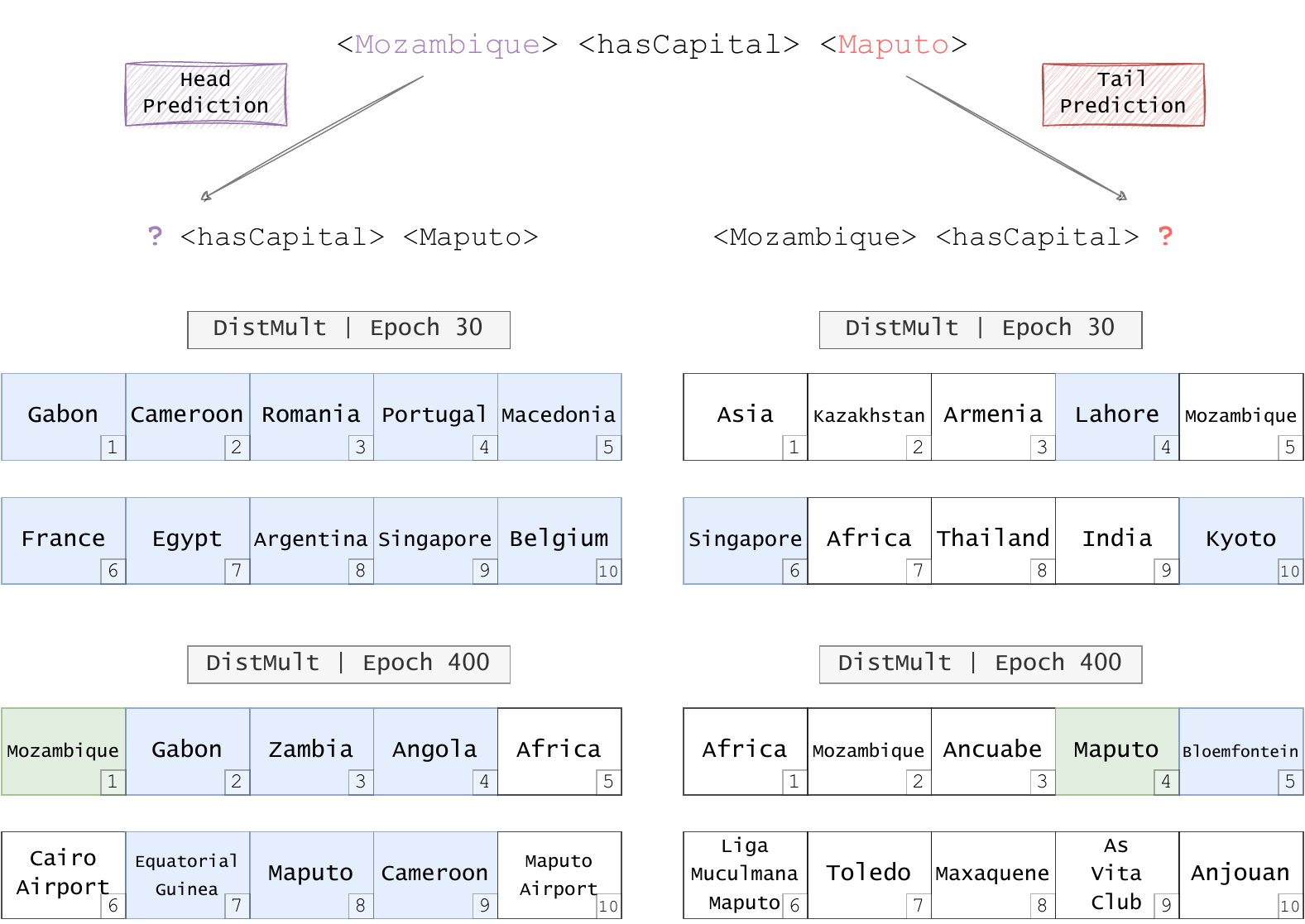}
    \caption{Top-ten ranked entities for head and tail predictions at epochs 30 and 400 for a sample triple from YAGO3-37K. Green, blue and white cells respectively denote the ground-truth entity, entities other than the ground-truth and semantically valid, and entities other than the ground-truth and semantically invalid. In this case, semantic validity is based on the domain and range of the relation \texttt{<hasCapital>}}
    \label{fig:distmult}
\end{figure}

Plots of the joint evolution of MRR and Sem@$K$ values show that most of the KGEMs reaches their best Sem@$K$ values after a few number of epochs. This means that predictions get semantically valid in the early stages of training. As previously mentioned, Sem@$K$ then usually start to decrease, as it has been noted for semantic matching models in particular. To this respect, an excerpt of the head and tail predictions of DistMult on YAGO3-37K is depicted in Fig.~\ref{fig:distmult}. Even though the ground-truth entity does not show up in neither the head nor the tail top-K list, we clearly see that after only 30 epochs of training, predictions made by DistMult are more meaningful than after 400 epochs of training. This relative trade-off between making semantically valid predictions and predictions that comprise the ground-truth entity higher in the top-$K$ list calls for finding a compromise in terms of training. The LP task is usually addressed in terms of rank-based metrics only, hence the choice of performing more and more training epochs so as to find the optimal KGEM in terms of MRR and Hits@K. However, as discussed in the present work, adding training steps may improve KGEM performance at the expense of its semantic awareness. In many cases, rank-based metrics values only slightly increase, whereas Sem@$K$ values drastically drop. For instance, comparing MRR vs. Sem@$K$ evolution of DistMult on FB15K237-ET (Fig.~\ref{subfig:fb-distmult}), we clearly see that after a moderate number of epochs, any additional epoch of training only provides a very slight improvement in terms of MRR, while it is very detrimental to Sem@$K$ values. Depending on the use case, such a decline in the semantic capabilities of the model is not desirable, and a compromise is to be found between training more to increase KGEM predictive performance and stopping training early enough so as not to deteriorate its semantic awareness too much.

\subsection{On the use of Sem@$K$ for different kinds of KGs}

As reported in Table~\ref{tab:typology}, KGs based upon a schema and a class hierarchy are candidates for the computation of all the versions of Sem@$K$. For the schema-defined KGs used in our experiments, we choose to report values regarding all these metrics so as to enable multi-view comparisons across models. From Table~\ref{tab:schema-defined-results-compiled} it can be clearly seen that the relative superiority of models is consistent throughout the different Sem@$K$ definitions. From a higher perspective, this means that even for schema-defined KGs with a hierarchy class, Sem@$K$[ext] is already a good proxy. This may be a good option to only rely on the Sem@$K$[ext] whenever the computation of Sem@$K$[base] is too expensive, due to the entity type checking part. This is even more true for Sem@$K$[wup], which requires an additional step of semantic relatedness computation.

\section{Discussion}
\label{discussion}
Three major research questions have been formulated in Section~\ref{intro}. Based on the analysis presented in Section~\ref{results}, we discuss each research question individually. We ultimately discuss the potential for further considerations of semantics into KGEMs.

\subsection{RQ1: how semantic-aware agnostic KGEMs are?}

From a coarse-grained viewpoint, we noted that KGEMs trained in an agnostic way prove capable of giving higher scores to semantically valid triples. However, disparities exist between models. Interestingly, these disparities seem to derive from the family of such models. Globally, translational models and GNNs -- represented by R-GCN and CompGCN in this work -- provide promising results.
It appears that the two aforementioned families of KGEMs are better able than semantic matching models (DistMult, ComplEx, SimplE) at recovering the semantics of entities and relations to give higher score to semantically valid triples. In fact, semantic matching models are almost systematically the worst performing models in terms of semantic awareness.
From a dynamic standpoint, it is worth noting the high semantic capabilities of KGEMs reached during the first epochs of training. In most cases, this is even during the first epochs that the optimal semantic awareness is attained.

\subsection{RQ2: how KGEM semantic awareness' evaluation should adapt to the typology of KGs?}

Drawing on the initial version of Sem@$K$ as presented in~\cite{dl4kg} -- referred to herein as Sem@$K$[base] -- an issue is quickly encountered when it comes to schemaless KGs, which do not contain any \texttt{rdfs:domain} (resp. \texttt{rdfs:range}) clause to indicate the class that candidate heads (resp. tails) should belong to. Our work introduces Sem@$K$[ext] -- a new version of Sem@$K$ that overcomes the aforementioned limitation.
In addition, even with schema-defined KGs, Sem@$K$[base] is not necessarily sufficient in itself. This metric can be further enriched whenever a KG comes with a class hierarchy. In Section~\ref{sematk-wup}, we integrate class hierarchy into Sem@$K$ by means of a similarity measure between concepts. We subsequently provide an example using the Wu-Palmer similarity score. The resulting Sem@$K$[wup] is used in the experiments in Section~\ref{results} and provide a finer-grained measure of KGEMs semantic awareness.

\subsection{RQ3: does the evaluation of KGEM semantic awareness offers different conclusions on the relative superiority of some KGEMs?}

A major finding is that models performing well with respect to rank-based metrics are not necessarily the most competitive regarding their semantic capabilities.
We previously noted that translational models globally showcase better Sem@$K$ values compared to semantic matching models. Considering MRR and Hits@$K$, the opposite conclusion is often drawn. Hence, the performance of KGEMs in terms of rank-based metrics is not indicative of their semantic capabilities. The only exception that might exist is for GNNs that perform well both in terms of rank-based metrics and semantic-oriented measures.

The answers provided to the research questions also lead to consider new matters.
As evidenced in Table~\ref{tab:schema-defined-results-compiled} and Table~\ref{tab:schemaless-results-compiled}, some KGs are more challenging with regard to Sem@$K$ results. Due to its tailored extraction strategy that purposedly favored difficult relations to feature in the validation and test sets, YAGO4-19K is the schema-defined KG with the lowest achieved Sem@$K$. This observation raises a deeper question: what characteristics of a KG make it inherently challenging for KGEMs to recover the semantics of entities and relations? An extensive study of the influence of KGs characteristics on the semantic capabilities of KGEMs would require to benchmark them on a broad set of KGs with varying dimensions, so as to determine those that are the most prevalent. Such characteristics can be the total number of relations, the average number of instances per class, or a combination of different factors. We leave this experimental study for future work.

Recall that this work is motivated by the possibility of going beyond a mere assessment of KGEM performance regarding rank-based metrics. We showed that these metrics only evaluate one aspect of such models, somehow providing a partial view on the quality of KGEMs. Our proposal for further assessing KGEM semantic capabilities aims at diving deeper into their predictive expressiveness and measuring to what extent their predictions are semantically valid. However, this second evaluation component does not shed full light on the respective KGEM peculiarities. Other evaluation components may be added, such as the storage and computational requirements of KGEMs~\cite{lightkgc,rdf2vec-light} and the environmental impact of their training and testing procedure~\cite{procrustes}. Furthermore, the explainability of KGEMs is another dimension that deserves great attention~\cite{xtranse,kelpie}.

\subsection{Towards further considerations of semantics in knowledge graph embeddings models}

The Sem@$K$ metric presented in this work allows for a more comprehensive evaluation of KGEMs. Based on domains and ranges of relations, Sem@$K$ assesses to what extent the predictions of a model are semantically valid. The present work constitutes one of several stepping stones toward the further consideration of ontological and semantic information in KGEM design and evaluation.

It should be noted that due to the only consideration of domains and ranges, Sem@$K$ cannot indicate whether predictions are logically consistent with other constraints posed by the ontology. This is in contrast with Inc@$K$ presented in~\cite{jain_iswc} that takes a broader set of ontological axioms into account. However, Inc@$K$ and Sem@$K$ intrinsically assess distinct dimensions of predictions. While the former is concerned with the logical consistency of predictions, the latter focuses on whether these predictions are semantically valid.
For instance, an ontology can specify that \texttt{City} is the range of \texttt{livesIn}, that \texttt{Seattle} is a \texttt{City} but not a \texttt{Capital}, and that entities of type \texttt{President} should be linked to a \texttt{Capital} through the relation \texttt{livesIn}.
Hence, it would still be meaningful and semantically correct to predict that \texttt{(BarackObama,livesIn,Seattle)}. However, this prediction is not logically consistent w.r.t. ontology specifications.
Inc@$K$ and Sem$K$ thus consider triples at different levels: while a given triple can be meaningful and semantically correct on its own, its combination with other triples may not be semantically valid or consistent with the ontology.
In future work, we will consider more expressive ontologies and see how the broad collection of axioms that constitute them can be incorporated into Sem@$K$.

KGEMs evaluated in this work are all agnostic to ontological information in their design. 
However, some models that consider or ensure specific ontological or logical properties exist.
For example, HAKE~\cite{hake} is constructed with the purpose of preserving hierarchies, Logic Tensor Networks~\cite{badreddineGSS22} are designed to ensure logical reasoning, and the training of TransOWL and TransROWL~\cite{damato2021} is enriched with additional triples deduced from, \textit{e.g.}, inverse predicates or equivalent classes.
Because of the integration of semantic information in their design or training, one could wonder if they present improved semantic awareness compared to agnostic models. 
Additionally, KGEMs can also be used to predict triples that represent class instantiations. 
A possible extension of the present work thus consists in studying whether predicted links and class instantiations are consistent and lead to increased Sem@$K$ values.
This would further qualify and highlight the semantic awareness and the consistency of predictions of KGEMs.
We leave these questions for future work.

\section{Conclusion}
\label{conclusion}
In this work, we consider the link prediction task and extend our previously introduced Sem@$K$ metric to measure the ability of KGEMs to
assign higher scores to triples that are semantically valid.
In particular, to adapt to different types of KGs (\textit{e.g.}, schemaless, class hierarchy), we introduce Sem@$K$[base], Sem@$K$[ext], or Sem@$K$[wup]. 
Compared with the traditional evaluation approach that solely relies on rank-based metrics, we show that the evaluation procedure is enhanced with the addition of semantic-oriented metrics that bring an additional perspective on KGEM quality.  Our experiments with different types of KGs highlight that there is no clear correlation between the performance of KGEMs in terms of traditional rank-based metrics versus their performance regarding semantic-oriented ones. In some cases, however, a trade-off does exist. Consequently, this calls for monitoring KGEM training under more scrutiny. Our experiments also point out that most of the conclusions that have been drawn actually hold at the level of families of models.

In future work, we will conduct experiments considering a broader array of KGEM families (\textit{e.g.}, KGEMs that include semantics) and propose evaluation metrics that consider additional and more expressive ontological constraints.

\section*{Acknowledgements}
This work is supported by the AILES PIA3 project (see \url{https://www.projetailes.com/}).
Experiments presented in this paper were carried out using the Grid'5000 testbed, supported by a scientific interest group hosted by Inria and including CNRS, RENATER and several Universities as well as other organizations (see \url{https://www.grid5000.fr}). 

\begin{appendix}
\label{appendix}

\section{Hyperparameters}
\label{appendix:hyperparameters}
For datasets with no reported optimal hyperparameters, grid-search based on curated hyperpamaters were performed. The full hyperparameter space is provided in Table~\ref{tab:searchspace}.
Chosen hyperparameters for each pair of KGEM and dataset are provided in Table~\ref{tab:hyperparams}.

\begin{table*}[h]
\centering
\caption{Hyperparameter search space}\label{tab:searchspace}
\begin{tabular}{ll}
\toprule Hyperparameters & Range \\
\midrule Batch Size & $\{128,256,512,1024,2048\}$ \\
Embedding Dimension & $\{50,100,150,200\}$ \\
Regularizer Type & $\{$None, $L1$, $L2$$\}$ \\
Regularizer Weight $(\lambda)$ & $\{1\mathrm{e}^{-1}, 1\mathrm{e}^{-2}, 1\mathrm{e}^{-3}, 1\mathrm{e}^{-4}, 1\mathrm{e}^{-5}\}$ \\
Learning Rate $(l r)$ & $\{0.005,0.003,0.001,0.0005,0.0003,0.0001\}$ \\
\bottomrule
\end{tabular}
\end{table*}

\begin{table*}[h]
\centering
\caption{Chosen hyperparameters for schema-defined and schemaless KGs used in the experiments. $|\mathcal{B}|$, $d$, $lr$, and $\lambda$ denote the batch size, embedding dimension, learning rate, and regularization weight, respectively. 
We experimentally found that $L2$ regularizer systematically worked the best. 
We therefore decide not to refer to it in the table. For ConvKB, the number of filters $\mathcal{T}$ was set to $128$. 
Parameters specific to R-GCN and CompGCN were left as defined in the original implementations.}\label{tab:hyperparams}
\renewcommand{\arraystretch}{1.0}
\setlength{\tabcolsep}{3pt}
\begin{tabular}{rrrrrrrrr}
\toprule 
 \multicolumn{1}{c}{} & \multicolumn{1}{c}{Hyperparameter} & \multicolumn{1}{c}{FB15K237-ET} & \multicolumn{1}{c}{DB93K} & \multicolumn{1}{c}{YAGO3-37K} & \multicolumn{1}{c}{YAGO4-19K} & \multicolumn{1}{c}{Codex-S} & \multicolumn{1}{c}{Codex-M} & \multicolumn{1}{c}{WN18RR} \\
\midrule
\multirow{4}{*}{TransE} & $|\mathcal{B}|$ & $512$&$256$&$256$&$ 512$&$128$&$128$&$512$\\
& $d$ & $100$&$200$&$150$&$100$&$100$&$100$&$100$\\
& $lr$ &$1\mathrm{e}^{-3}$&$5\mathrm{e}^{-3}$&$5\mathrm{e}^{-3}$&$5\mathrm{e}^{-2}$&$1\mathrm{e}^{-3}$&$1\mathrm{e}^{-3}$&$1\mathrm{e}^{-3}$\\
& $\lambda$ &$1\mathrm{e}^{-5}$&$1\mathrm{e}^{-5}$&$1\mathrm{e}^{-5}$&$1\mathrm{e}^{-5}$&$1\mathrm{e}^{-5}$&$1\mathrm{e}^{-5}$&$1\mathrm{e}^{-5}$\\
\midrule
\multirow{4}{*}{TransH}& $|\mathcal{B}|$ & $512$&$256$&$256$&$ 512$&$128$&$128$&$512$\\
& $d$ & $100$&$200$&$150$&$100$&$100$&$100$&$100$\\
& $lr$ &$1\mathrm{e}^{-3}$&$5\mathrm{e}^{-3}$&$5\mathrm{e}^{-3}$&$5\mathrm{e}^{-2}$&$1\mathrm{e}^{-3}$&$1\mathrm{e}^{-3}$&$1\mathrm{e}^{-3}$\\
& $\lambda$ &$1\mathrm{e}^{-5}$&$1\mathrm{e}^{-5}$&$1\mathrm{e}^{-5}$&$1\mathrm{e}^{-5}$&$1\mathrm{e}^{-5}$&$1\mathrm{e}^{-5}$&$1\mathrm{e}^{-5}$\\
\midrule
\multirow{4}{*}{DistMult} & $|\mathcal{B}|$ & $1024$&$1024$&$1024$&$1024$&$1024$&$1024$&$1024$\\
& $d$ & $100$&$200$&$150$&$100$&$100$&$100$&$100$\\
& $lr$ &$1\mathrm{e}^{-3}$&$1\mathrm{e}^{-3}$&$5\mathrm{e}^{-3}$&$5\mathrm{e}^{-2}$&$1\mathrm{e}^{-3}$&$1\mathrm{e}^{-3}$&$1\mathrm{e}^{-3}$\\
& $\lambda$ & $1\mathrm{e}^{-5}$&$1\mathrm{e}^{-5}$&$1\mathrm{e}^{-5}$&$0$&$0$&$0$&$0$\\
\midrule
\multirow{4}{*}{ComplEx} & $|\mathcal{B}|$ & $1024$&$1024$&$1024$&$1024$&$1024$&$1024$&$1024$\\
& $d$ & $100$&$200$&$150$&$100$&$100$&$100$&$100$\\
& $lr$ &$1\mathrm{e}^{-3}$&$5\mathrm{e}^{-3}$&$1\mathrm{e}^{-3}$&$1\mathrm{e}^{-3}$&$1\mathrm{e}^{-3}$&$1\mathrm{e}^{-3}$&$1\mathrm{e}^{-3}$\\
& $\lambda$ & $1\mathrm{e}^{-2}$&$1\mathrm{e}^{-1}$&$1\mathrm{e}^{-5}$&$1\mathrm{e}^{-2}$&$1\mathrm{e}^{-2}$&$1\mathrm{e}^{-2}$&$1\mathrm{e}^{-2}$\\
\midrule
\multirow{4}{*}{SimplE} & $|\mathcal{B}|$ & $1024$&$1024$&$1024$&$1024$&$1024$&$1024$&$1024$\\
& $d$ & $100$&$200$&$150$&$100$&$100$&$100$&$100$\\
& $lr$ &$1\mathrm{e}^{-3}$&$5\mathrm{e}^{-3}$&$1\mathrm{e}^{-3}$&$1\mathrm{e}^{-3}$&$1\mathrm{e}^{-3}$&$1\mathrm{e}^{-3}$&$1\mathrm{e}^{-3}$\\
& $\lambda$ & $1\mathrm{e}^{-2}$&$1\mathrm{e}^{-1}$&$1\mathrm{e}^{-5}$&$0$&$1\mathrm{e}^{-2}$&$1\mathrm{e}^{-2}$&$0$\\
\midrule
\multirow{4}{*}{ConvE} & $|\mathcal{B}|$ & $512$&$256$&$256$&$512$&$512$&$512$&$512$\\
& $d$ & $200$&$200$&$200$&$200$&$200$&$200$&$200$\\
& $lr$ &$1\mathrm{e}^{-3}$&$5\mathrm{e}^{-3}$&$5\mathrm{e}^{-3}$&$1\mathrm{e}^{-3}$&$1\mathrm{e}^{-3}$&$1\mathrm{e}^{-3}$&$1\mathrm{e}^{-3}$\\
& $\lambda$ & $1\mathrm{e}^{-5}$&$1\mathrm{e}^{-5}$&$1\mathrm{e}^{-5}$&$1\mathrm{e}^{-5}$&$1\mathrm{e}^{-5}$&$1\mathrm{e}^{-5}$&$1\mathrm{e}^{-5}$\\
\midrule
\multirow{4}{*}{ConvKB} & $|\mathcal{B}|$ & $512$&$256$&$256$&$512$&$512$&$512$&$512$\\
& $d$ & $100$&$100$&$100$&$100$&$100$&$100$&$100$\\
& $lr$ &$1\mathrm{e}^{-3}$&$1\mathrm{e}^{-3}$&$5\mathrm{e}^{-3}$&$1\mathrm{e}^{-3}$&$1\mathrm{e}^{-3}$&$1\mathrm{e}^{-3}$&$1\mathrm{e}^{-3}$\\
& $\lambda$ & $1\mathrm{e}^{-5}$&$1\mathrm{e}^{-5}$&$1\mathrm{e}^{-5}$&$1\mathrm{e}^{-5}$&$1\mathrm{e}^{-5}$&$1\mathrm{e}^{-5}$&$1\mathrm{e}^{-5}$\\
\midrule
\multirow{3}{*}{R-GCN} & $d$ & $500$&$500$&$500$&$500$&$500$&$500$&$500$\\
& $lr$ &$1\mathrm{e}^{-2}$&$1\mathrm{e}^{-2}$&$1\mathrm{e}^{-2}$&$1\mathrm{e}^{-2}$&$1\mathrm{e}^{-2}$&$1\mathrm{e}^{-2}$&$1\mathrm{e}^{-2}$\\
& $\lambda$ & $1\mathrm{e}^{-2}$&$1\mathrm{e}^{-2}$&$1\mathrm{e}^{-2}$&$1\mathrm{e}^{-2}$&$1\mathrm{e}^{-2}$&$1\mathrm{e}^{-2}$&$1\mathrm{e}^{-2}$\\
\midrule
\multirow{4}{*}{CompGCN} & $|\mathcal{B}|$ & $1024$&$1024$&$1024$&$1024$&$1024$&$1024$&$1024$\\
& $d$ & $200$&$200$&$200$&$200$&$200$&$200$&$200$\\
& $lr$ &$1\mathrm{e}^{-3}$&$1\mathrm{e}^{-3}$&$1\mathrm{e}^{-3}$&$1\mathrm{e}^{-3}$&$1\mathrm{e}^{-3}$&$1\mathrm{e}^{-3}$&$1\mathrm{e}^{-3}$\\
& $\lambda$ & $0$&$0$&$0$&$0$&$0$&$0$&$0$\\
\bottomrule
\end{tabular}
\end{table*}

\section{Results achieved with the best reported hyperparameters}
\label{appendix:results}

Results achieved with the best reported hyperparameters are presented in Tables~\ref{tab:schema-defined-results-compiled} and~\ref{tab:schemaless-results-compiled}. 

\begin{table*}[h]
        \caption{Rank-based and semantic-based results on the schema-defined knowledge graphs. Bold fonts indicate which model performs best with respect to a given metric.}
	\label{tab:schema-defined-results-compiled}
	\begin{subtable}[h]{1.0\textwidth}
	\centering
	\caption{FB15K237-ET}
	\label{tab:schema-defined-results-fb15k237}
                \vspace{0.5mm}
                \begin{adjustbox}{max width={\textwidth},totalheight={0.2\textheight},keepaspectratio}%
			\begin{tabular}{cc|cccc|ccc|ccc|ccc}
                    
                    \toprule
				& & \multicolumn{4}{c|}{Rank-based} & \multicolumn{3}{c|}{Sem@$K$[base]} & \multicolumn{3}{c|}{Sem@$K$[wup]} & \multicolumn{3}{c}{Sem@$K$[ext]} \\
				Model Family & Model & MRR & H@1 & H@3 & H@10 & 
                    S@1 & S@3 & S@10 &
                    S@1 & S@3 & S@10 &
                    S@1 & S@3 & S@10 \\
				\midrule
				\multirow{2}*{Geometric}& TransE &
    $.273$&$.182$&$.302$&$.453$& $.978$&$.964$&$.949$& $.983$&$.972$&$.961$& $.888$&$.870$&$.845$\\
				& TransH &
    $.281$&$.187$&$.306$&$.458$& $.981$&$.968$&$.955$& $.986$&$.976$&$.966$& $.896$&$.873$&$.855$ \\
				\midrule
				\multirow{3}*{Semantic Matching}& DistMult &
    $.289$&$.206$&$.313$&$.456$& $.907$&$.919$&$.922$& $.919$&$.930$&$.934$& $.842$&$.853$&$.849$ \\
				& ComplEx &
    $.267$&$.185$&$.292$&$.431$& $.874$&$.823$&$.761$& $.897$&$.856$&$.804$& $.822$&$.770$&$.7005$  \\
    			& SimplE &
    $.257$&$.180$&$.275$&$.416$& $.895$&$.878$&$.848$& $.911$&$.896$&$.869$& $.839$&$.817$&$.776$  \\
				\midrule
				\multirow{4}*{Convolutional}& ConvE &
    $.310$&$.211$&$.337$&$.490$& $.972$&$.970$&$.961$& $.977$&$.973$&$.968$& $.883$&$.880$&$.870$ \\
				& ConvKB &
    $.251$&$.159$&$.261$&$.411$& $.918$&$.897$&$.872$& $.935$&$.918$&$.899$& $.842$&$.812$&$.781$  \\
    			& R-GCN &
    $.238$&$.153$&$.257$&$.412$& $.980$&$.970$&$.953$& $.985$&$.976$&$.961$& $.902$&$.889$&$.861$  \\
        		& CompGCN &
    $\mathbf{.337}$&$\mathbf{.227}$&$\mathbf{.365}$&$\mathbf{.530}$& $\mathbf{.998}$&$\mathbf{.996}$&$\mathbf{.992}$& $\mathbf{.999}$&$\mathbf{.997}$&$\mathbf{.995}$& $\mathbf{.993}$&$\mathbf{.986}$&$\mathbf{.976}$  \\
    \bottomrule
    \vspace{0.5mm}
			\end{tabular}
                \end{adjustbox}
	\end{subtable}

	\begin{subtable}[h]{1.0\textwidth}
	\centering
	\caption{DB93K}
	\label{tab:schema-defined-results-db93k}
                \vspace{0.5mm}
                \begin{adjustbox}{max width={\textwidth},totalheight={0.2\textheight},keepaspectratio}%
			\begin{tabular}{cc|cccc|ccc|ccc|ccc}
                    \toprule
				& & \multicolumn{4}{c|}{Rank-based} & \multicolumn{3}{c|}{Sem@$K$[base]} & \multicolumn{3}{c|}{Sem@$K$[wup]} & \multicolumn{3}{c}{Sem@$K$[ext]} \\
				Model Family & Model & MRR & H@1 & H@3 & H@10 & 
                    S@1 & S@3 & S@10 &
                    S@1 & S@3 & S@10 &
                    S@1 & S@3 & S@10 \\
				\midrule
				\multirow{2}*{Geometric}& TransE &
    $.233$&$.145$&$.275$&$.397$& $.985$&$.975$&$.961$& $.991$&$.987$&$.979$& $.767$&$.730$&$.689$\\
				& TransH &
    $.236$&$.147$&$.278$&$.399$& $\mathbf{.988}$&$\mathbf{.979}$&$\mathbf{.968}$& $\mathbf{.993}$&$\mathbf{.990}$&$\mathbf{.981}$& $.769$&$.739$&$.694$ \\
				\midrule
				\multirow{3}*{Semantic Matching}& DistMult &
    $.261$&$.202$&$.287$&$.369$& $.865$&$.831$&$.790$& $.890$&$.865$&$.833$& $.716$&$.670$&$.617$\\
				& ComplEx &
    $.287$&$.213$&$.325$&$.417$& $.941$&$.917$&$.877$& $.955$&$.937$&$.907$& $.815$&$.767$&$.698$ \\
    			& SimplE &
    $.252$&$.202$&$.274$&$.339$& $.896$&$.871$&$.837$& $.915$&$.895$&$.867$& $.774$&$.719$&$.659$ \\
				\midrule
				\multirow{4}*{Convolutional}& ConvE &
    $.256$&$.183$&$.289$&$.392$& $.871$&$.862$&$.862$& $.882$&$.874$&$.874$& $.764$&$.750$&$.734$\\
				& ConvKB &
    $.178$&$.121$&$.199$&$.283$& $.883$&$.895$&$.881$& $.902$&$.895$&$.881$& $.742$&$.702$&$.659$ \\
    			& R-GCN &
    $.208$&$.144$&$.233$&$.319$& $.969$&$.959$&$.945$& $.979$&$.972$&$.961$& $.819$&$.785$&$.740$ \\
        		& CompGCN &
    $\mathbf{.319}$&$\mathbf{.249}$&$\mathbf{.351}$&$\mathbf{.446}$& $\mathbf{.988}$&$.976$&$.967$& $.991$&$.982$&$.975$& $\mathbf{.953}$&$\mathbf{.918}$&$\mathbf{.878}$ \\
    \bottomrule
    \vspace{0.5mm}
			\end{tabular}
                \end{adjustbox}
	\end{subtable}

	\begin{subtable}[h]{1.0\textwidth}
	\centering
	\caption{YAGO3-37K}
	\label{tab:schema-defined-results-yago3}
                \vspace{0.5mm}
                \begin{adjustbox}{max width={\textwidth},totalheight={0.2\textheight},keepaspectratio}%
			\begin{tabular}{cc|cccc|ccc|ccc|ccc}
                    \toprule
				& & \multicolumn{4}{c|}{Rank-based} & \multicolumn{3}{c|}{Sem@$K$[base]} & \multicolumn{3}{c|}{Sem@$K$[wup]} & \multicolumn{3}{c}{Sem@$K$[ext]} \\
				Model Family & Model & MRR & H@1 & H@3 & H@10 & 
                    S@1 & S@3 & S@10 &
                    S@1 & S@3 & S@10 &
                    S@1 & S@3 & S@10 \\
				\midrule
				\multirow{2}*{Geometric}& TransE &
    $.184$&$.080$&$.198$&$.408$& $.989$&$.988$&$.988$& $.993$&$.994$&$.995$& $.897$&$.904$&$.911$\\
				& TransH &
    $.187$&$.091$&$.199$&$.415$& $.995$&$.993$&$.993$& $.995$&$.997$&$.997$& $.901$&$.911$&$.925$ \\
				\midrule
				\multirow{3}*{Semantic Matching}& DistMult &
    $.225$&$.112$&$.251$&$.465$& $.885$&$.940$&$.959$& $.921$&$.962$&$.980$& $.795$&$.851$&$.886$\\
				& ComplEx &
    $.430$&$.250$&$.551$&$.780$& $.740$&$.820$&$.859$& $.911$&$.935$&$.950$& $.662$&$.735$&$.794$ \\
    			& SimplE &
    $.311$&$.081$&$.468$&$.749$& $.377$&$.733$&$.866$& $.786$&$.901$&$.944$& $.310$&$.640$&$.777$ \\
				\midrule
				\multirow{4}*{Convolutional}& ConvE &
    $\mathbf{.493}$&$\mathbf{.350}$&$\mathbf{.578}$&$\mathbf{.775}$& $.933$&$.923$&$.910$& $.977$&$.977$&$.975$& $.893$&$.879$&$.871$\\
				& ConvKB &
    $.305$&$.162$&$.352$&$.631$& $.979$&$.979$&$.978$& $.990$&$.990$&$.990$& $.899$&$.896$&$.904$ \\
    			& R-GCN &
    $.115$&$.046$&$.110$&$.254$& $.993$&$.994$&$.993$& $.995$&$.996$&$.997$& $.933$&$.932$&$.928$ \\
        		& CompGCN &
    $.399$&$.269$&$.464$&$.663$& $\mathbf{.998}$&$\mathbf{.997}$&$\mathbf{.996}$& $\mathbf{.999}$&$\mathbf{.999}$&$\mathbf{.998}$& $\mathbf{.998}$&$\mathbf{.994}$&$\mathbf{.982}$ \\
    \bottomrule
    \vspace{0.5mm}
			\end{tabular}
                \end{adjustbox}
	\end{subtable}

 \begin{subtable}[h]{1.0\textwidth}
	\centering
	\caption{YAGO4-19K}
 	\label{tab:schema-defined-results-yago4}
                \vspace{0.5mm}
                \begin{adjustbox}{max width={\textwidth},totalheight={0.2\textheight},keepaspectratio}%
			\begin{tabular}{cc|cccc|ccc|ccc|ccc}
                    \toprule
				& & \multicolumn{4}{c|}{Rank-based} & \multicolumn{3}{c|}{Sem@$K$[base]} & \multicolumn{3}{c|}{Sem@$K$[wup]} & \multicolumn{3}{c}{Sem@$K$[ext]} \\
				Model Family & Model & MRR & H@1 & H@3 & H@10 & 
                    S@1 & S@3 & S@10 &
                    S@1 & S@3 & S@10 &
                    S@1 & S@3 & S@10 \\
				\midrule
				\multirow{2}*{Geometric}& TransE &
    $.749$&$.651$&$.833$&$.901$& $.975$&$.890$&$.843$& $.980$&$.914$&$.870$& $.844$&$.647$&$.547$\\
				& TransH &
    $.752$&$.656$&$.829$&$.898$& $.979$&$\mathbf{.909}$&$\mathbf{.868}$& $\mathbf{.982}$&$\mathbf{.931}$&$\mathbf{.891}$& $.848$&$.656$&$.564$ \\
				\midrule
				\multirow{3}*{Semantic Matching}& DistMult &
    $.863$&$.839$&$.881$&$.898$& $.931$&$.687$&$.534$& $.933$&$.686$&$.551$& $.901$&$.556$&$.376$\\
				& ComplEx &
    $.890$&$.881$&$.897$&$.903$& $.923$&$.627$&$.479$& $.923$&$.635$&$.496$& $.909$&$.524$&$.349$ \\
    			& SimplE &
    $.808$&$.757$&$.848$&$.883$& $.891$&$.646$&$.500$& $.894$&$.654$&$.516$& $.840$&$.530$&$.367$ \\
				\midrule
				\multirow{4}*{Convolutional}& ConvE &
    $.901$&$.891$&$\mathbf{.908}$&$.913$& $.977$&$.858$&$.844$& $.977$&$.869$&$.854$& $.948$&$.731$&$.709$\\
				& ConvKB &
    $.828$&$.772$&$.883$&$.908$& $.865$&$.614$&$.564$& $.870$&$.639$&$.590$& $.793$&$.460$&$.361$ \\
    			& R-GCN &
    $.893$&$.885$&$.896$&$.903$& $.969$&$.832$&$.727$& $.970$&$.841$&$.741$& $.930$&$.670$&$.522$ \\
        		& CompGCN &
    $\mathbf{.907}$&$\mathbf{.903}$&$\mathbf{.908}$&$\mathbf{.918}$& $\mathbf{.980}$&$.906$&$.862$& $.981$&$.912$&$.870$& $\mathbf{.968}$&$\mathbf{.815}$&$\mathbf{.725}$ \\
    \bottomrule
    \vspace{0.5mm}
			\end{tabular}
                \end{adjustbox}
	\end{subtable}
\end{table*}

\begin{table*}[h]
        \caption{Rank-based and semantic-based results on the schemaless knowledge graphs. Bold fonts indicate which model performs best with respect to a given metric.}
	\label{tab:schemaless-results-compiled}
	\begin{subtable}[h]{1.0\textwidth}
	\centering
	\caption{Codex-S}
	\label{tab:schemaless-results-codex-s}
                \vspace{0.5mm}
			\begin{tabular}{cc|cccc|ccc}
                    \toprule
				& & \multicolumn{4}{c|}{Rank-based} & \multicolumn{3}{c}{Sem@$K$[ext]} \\
				Model Family & Model & MRR & H@1 & H@3 & H@10 & 
                    S@1 & S@3 & S@10  \\
				\midrule
				\multirow{2}*{Geometric}& TransE &
    $.354$&$.223$&$.409$&$.620$& $.927$&$.900$&$.873$\\
				& TransH &
    $.355$&$.225$&$.410$&$.621$& $.928$&$.901$&$.875$\\
				\midrule
				\multirow{3}*{Semantic Matching}& DistMult &
    $.442$&$.336$&$.487$&$.661$& $.935$&$.919$&$.887$\\
				& ComplEx &
    $\mathbf{.511}$&$\mathbf{.426}$&$\mathbf{.552}$&$\mathbf{.668}$& $.872$&$.786$&$.694$\\
    			& SimplE &
    $.464$&$.368$&$.514$&$.641$& $.876$&$.789$&$.727$\\
				\midrule
				\multirow{4}*{Convolutional}& ConvE &
    $.379$&$.259$&$.439$&$.610$& $.912$&$.913$&$.886$\\
				& ConvKB &
    $.453$&$.361$&$.493$&$.632$& $.858$&$.806$&$.769$\\
    			& R-GCN &
    $.335$&$.225$&$.381$&$.556$& $.936$&$.921$&$.889$\\
        		& CompGCN &
    $.380$&$.268$&$.427$&$.597$& $\mathbf{.993}$&$\mathbf{.986}$&$\mathbf{.974}$\\
    \bottomrule
    \vspace{0.5mm}
			\end{tabular}
	\end{subtable}

	\begin{subtable}[h]{1.0\textwidth}
	\centering
	\caption{Codex-M}
	\label{tab:schemaless-results-codex-m}
                \vspace{0.5mm}
			\begin{tabular}{cc|cccc|ccc}
                    \toprule
				& & \multicolumn{4}{c|}{Rank-based} & \multicolumn{3}{c}{Sem@$K$[ext]} \\
				Model Family & Model & MRR & H@1 & H@3 & H@10 & 
                    S@1 & S@3 & S@10  \\
				\midrule
				\multirow{2}*{Geometric}& TransE &
    $.262$&$.188$&$.284$&$.405$& $.889$&$.855$&$.839$\\
				& TransH &
    $.266$&$.190$&$.291$&$.415$& $.894$&$.862$&$.842$\\
				\midrule
				\multirow{3}*{Semantic Matching}& DistMult &
    $.268$&$.199$&$.291$&$.402$& $.726$&$.749$&$.765$\\
				& ComplEx &
    $.274$&$.213$&$.298$&$.386$& $.768$&$.687$&$.608$\\
    			& SimplE &
    $.269$&$.212$&$.289$&$.378$& $.799$&$.713$&$.629$\\
				\midrule
				\multirow{4}*{Convolutional}& ConvE &
    $.255$&$.184$&$.280$&$.395$& $.863$&$.839$&$.814$\\
				& ConvKB &
    $.230$&$.156$&$.250$&$.376$& $.797$&$.771$&$.770$\\
    			& R-GCN &
    $.185$&$.110$&$.201$&$.340$& $.944$&$.920$&$.887$\\
        		& CompGCN &
    $\mathbf{.314}$&$\mathbf{.237}$&$\mathbf{.346}$&$\mathbf{.460}$& $\mathbf{.996}$&$\mathbf{.992}$&$\mathbf{.983}$\\
    \bottomrule
    \vspace{0.5mm}
			\end{tabular}
	\end{subtable}

	\begin{subtable}[h]{1.0\textwidth}
	\centering
	\caption{WN18RR}
	\label{tab:schemaless-results-wn18rr}
                \vspace{0.5mm}
			\begin{tabular}{cc|cccc|ccc}
                    \toprule
				& & \multicolumn{4}{c|}{Rank-based} & \multicolumn{3}{c}{Sem@$K$[ext]} \\
				Model Family & Model & MRR & H@1 & H@3 & H@10 & 
                    S@1 & S@3 & S@10  \\
				\midrule
				\multirow{2}*{Geometric}& TransE &
    $.186$&$.032$&$.303$&$.455$& $.770$&$.756$&$.715$\\
				& TransH &
    $.191$&$.034$&$.306$&$.458$& $.773$&$.760$&$.717$\\
				\midrule
				\multirow{3}*{Semantic Matching}& DistMult &
    $.399$&$.372$&$.413$&$.444$& $.795$&$.732$&$.688$\\
				& ComplEx &
    $.430$&$.400$&$.446$&$.487$& $.715$&$.669$&$.673$\\
    			& SimplE &
    $.397$&$.375$&$.406$&$.438$& $.690$&$.598$&$.673$\\
				\midrule
				\multirow{4}*{Convolutional}& ConvE &
    $.406$&$.375$&$.422$&$.459$& $.865$&$.854$&$.871$\\
				& ConvKB &
    $.356$&$.302$&$.391$&$.444$& $.712$&$.713$&$.732$\\
    			& R-GCN &
    $.382$&$.345$&$.402$&$.441$& $.836$&$.797$&$.752$\\
        		& CompGCN &
    $\mathbf{.471}$&$\mathbf{.437}$&$\mathbf{.483}$&$\mathbf{.536}$& $\mathbf{.970}$&$\mathbf{.946}$&$\mathbf{.927}$\\
    \bottomrule
    \vspace{0.5mm}
			\end{tabular}
	\end{subtable}
\end{table*}

\section{Evolution of MRR and Sem@$K$ values with respect to the number of epochs}
\label{appendix:plots}

\begin{figure}[h]
    \centering
    \begin{subfigure}[c]{0.49\textwidth}
        \centering
        \includegraphics[width=\textwidth]{fig/FB15K237-ET/FB15K237-TransE.pdf}
        \caption{TransE -- FB15K237-ET}
        \label{subfig:fb-transe}
    \end{subfigure}
    \hfill
    \begin{subfigure}[c]{0.49\textwidth}
        \centering
        \includegraphics[width=\textwidth]{fig/FB15K237-ET/FB15K237-TransH.pdf}
        \caption{TransH -- FB15K237-ET}
        \label{subfig:fb-transh}
    \end{subfigure}
    
    \begin{subfigure}[c]{0.49\textwidth}
        \centering
        \includegraphics[width=\textwidth]{fig/FB15K237-ET/FB15K237-DistMult.pdf}
        \caption{DistMult -- FB15K237-ET}
        \label{subfig:fb-distmult}
    \end{subfigure}
    \hfill
    \begin{subfigure}[c]{0.49\textwidth}
        \centering
        \includegraphics[width=\textwidth]{fig/FB15K237-ET/FB15K237-ComplEx.pdf}
        \caption{ComplEx -- FB15K237-ET}
        \label{subfig:fb-complex}
    \end{subfigure}
    
    \begin{subfigure}[c]{0.49\textwidth}
        \centering
        \includegraphics[width=\textwidth]{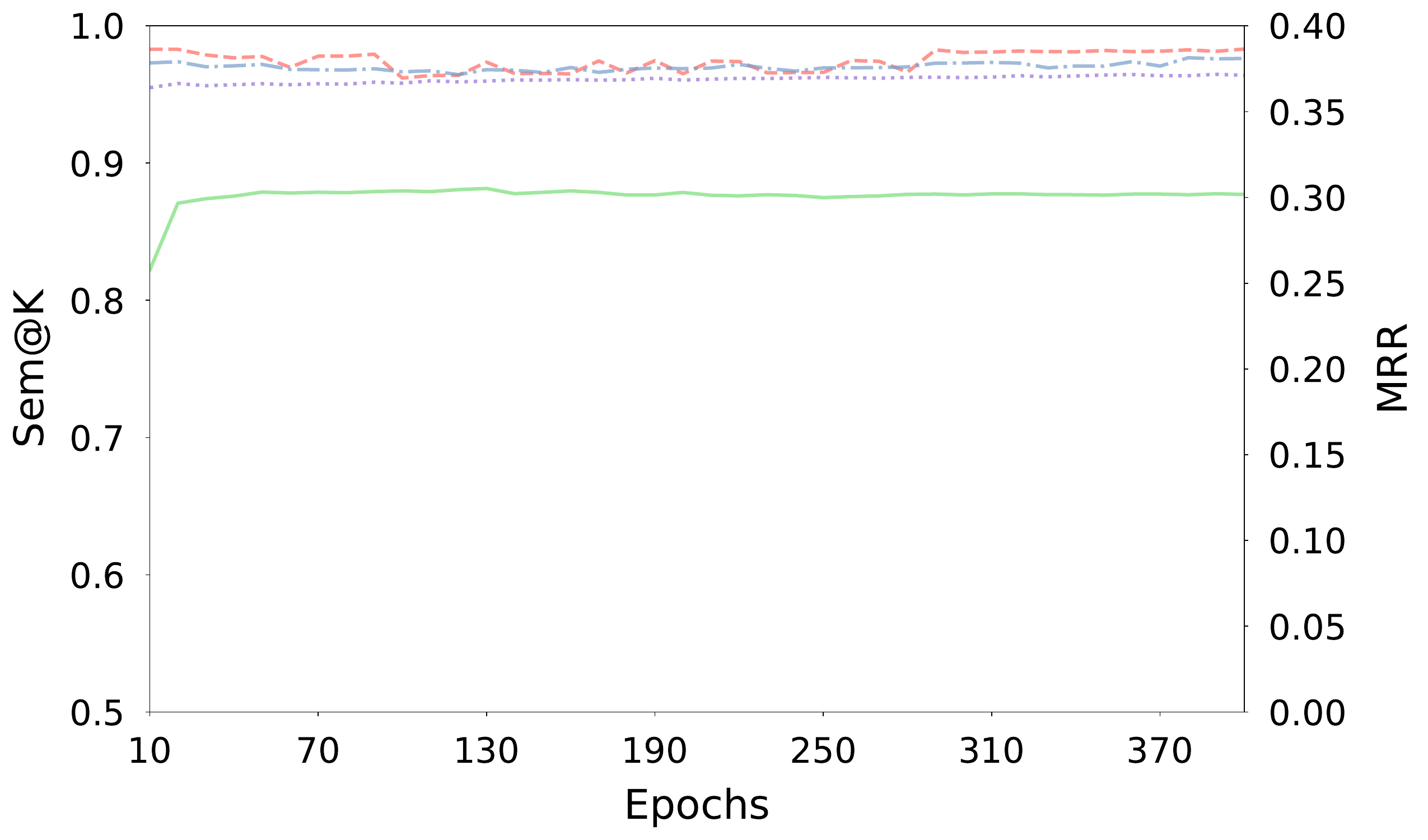}
        \caption{ConvE -- FB15K237-ET}
        \label{subfig:fb-conve}
    \end{subfigure}
    \hfill
    \begin{subfigure}[c]{0.49\textwidth}
        \centering
        \includegraphics[width=\textwidth]{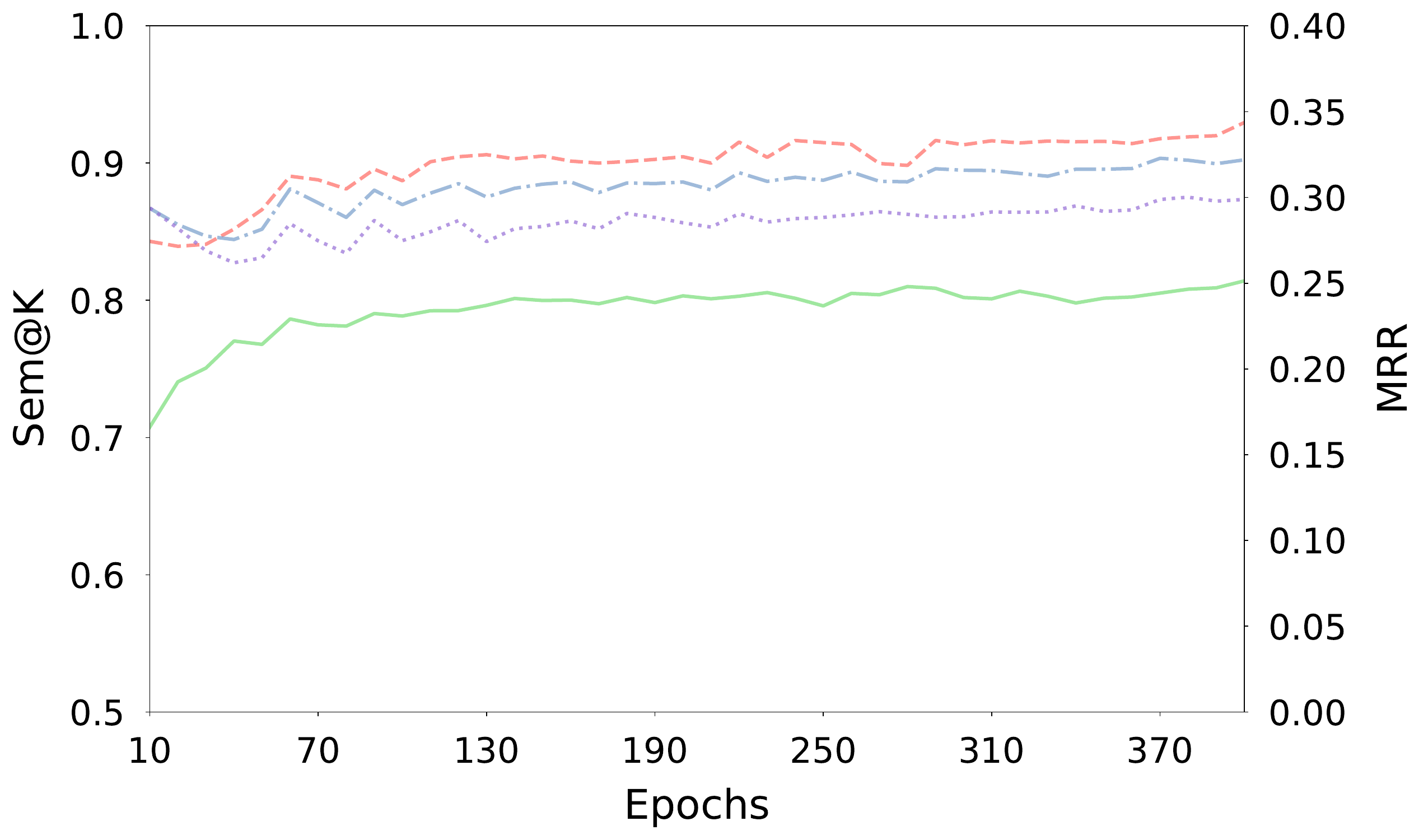}
        \caption{ConvKB -- FB15K237-ET}
        \label{subfig:fb-convkb}
    \end{subfigure}

    \begin{subfigure}[c]{0.49\textwidth}
        \centering
        \includegraphics[width=\textwidth]{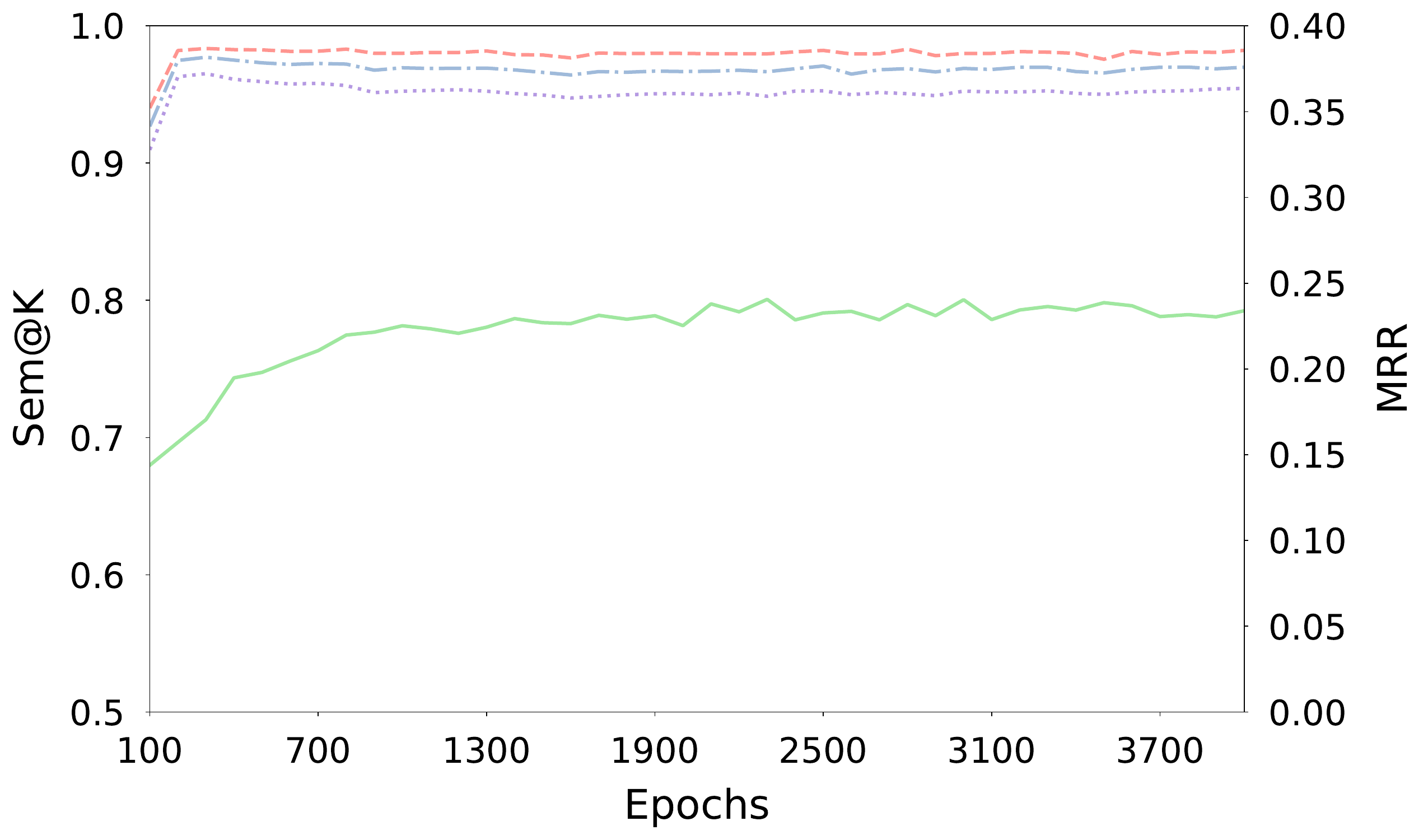}
        \caption{R-GCN -- FB15K237-ET}
        \label{subfig:fb-rgcn}
    \end{subfigure}
    \hfill
    \begin{subfigure}[c]{0.49\textwidth}
        \centering
        \includegraphics[width=\textwidth]{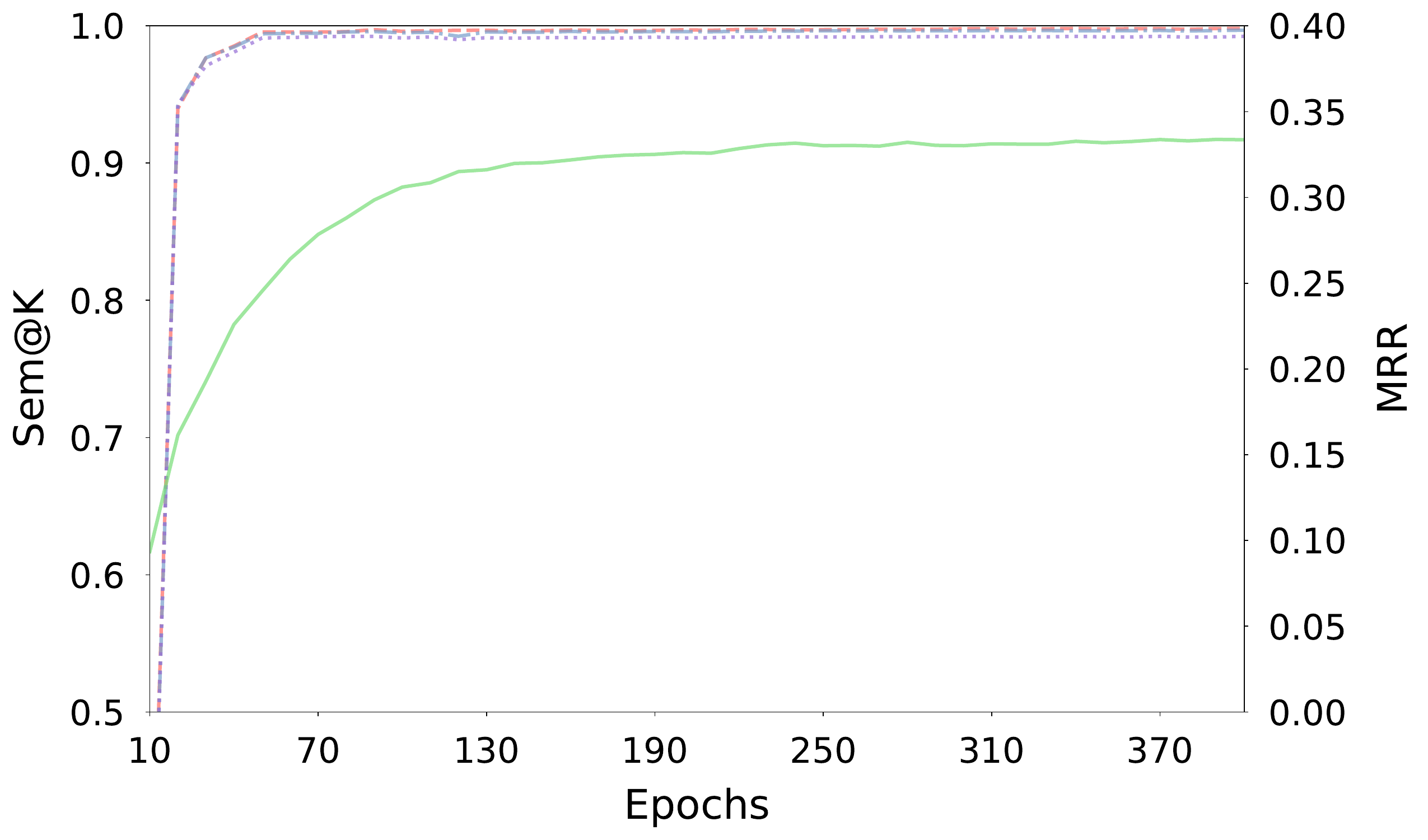}
        \caption{CompGCN -- FB15K237-ET}
        \label{subfig:fb-compgcn}
    \end{subfigure}

    \caption[]{Evolution of MRR (\raisebox{2pt}{\begin{tikzpicture}[scale=0.5]
        \draw[color=pastelgreen, solid, line width=1pt] (0,0) -- (0.7,0);
    \end{tikzpicture}}), Sem@1 (\raisebox{2pt}{\begin{tikzpicture}[scale=0.5]
        \draw[color=pastelred, dashed, line width=1pt] (0,0) -- (0.65,0);
    \end{tikzpicture}}), Sem@3 (\raisebox{2pt}{\begin{tikzpicture}[scale=0.5]
        \draw[color=darkpastelblue, dash dot, line width=1pt] (0,0) -- (0.7,0);
    \end{tikzpicture}}), and Sem@10 (\raisebox{2pt}{\begin{tikzpicture}[scale=0.5]
        \draw[color=darkpastelpurple, dotted, line width=1pt] (0,0) -- (0.7,0);
    \end{tikzpicture}}) on FB15K237-ET}
    \label{fig:results}
\end{figure}

\begin{figure}[h]
    \centering
    \begin{subfigure}[c]{0.49\textwidth}
        \centering
        \includegraphics[width=\textwidth]{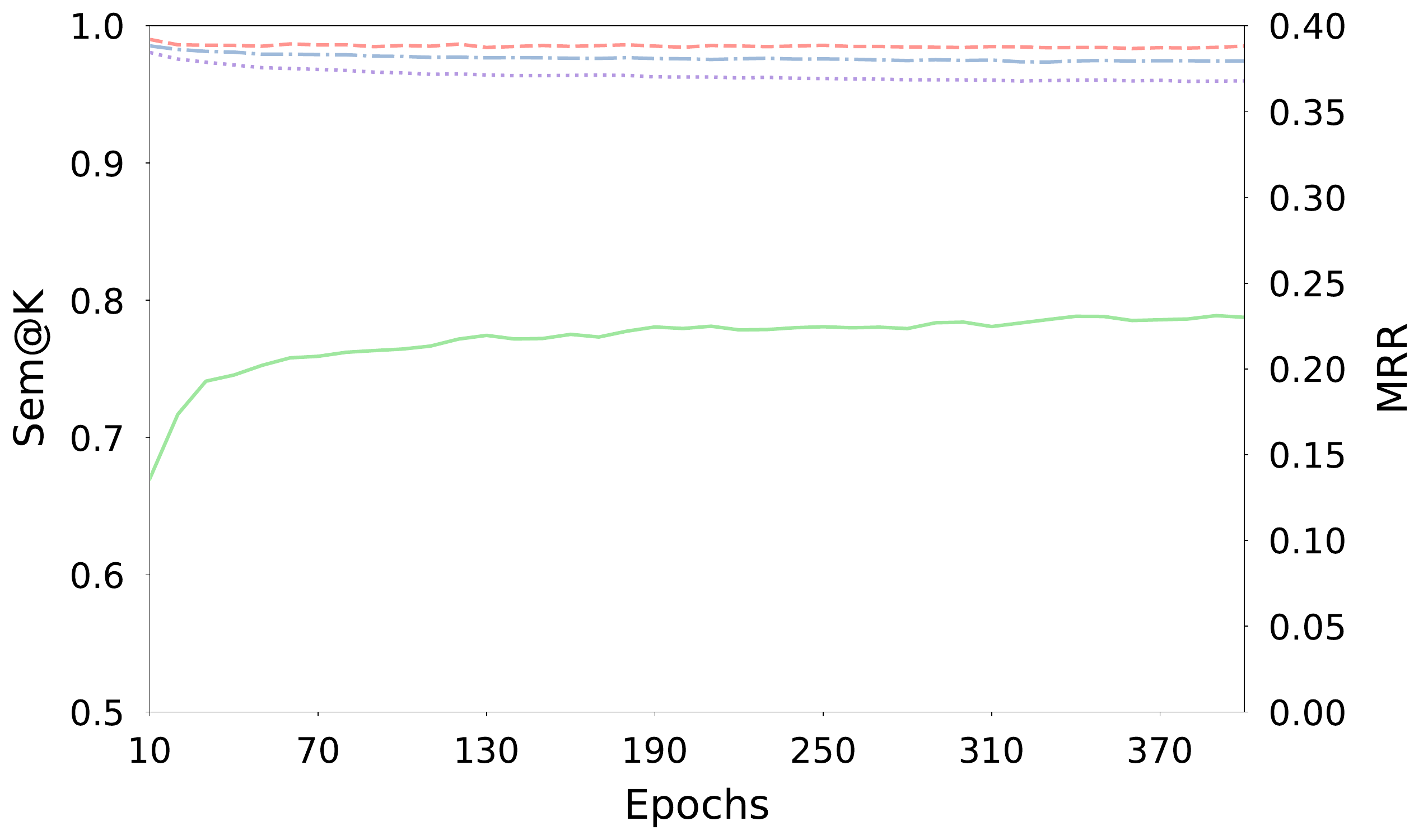}
        \caption{TransE -- DB93K}
        \label{subfig:db-transe}
    \end{subfigure}
    \hfill
    \begin{subfigure}[c]{0.49\textwidth}
        \centering
        \includegraphics[width=\textwidth]{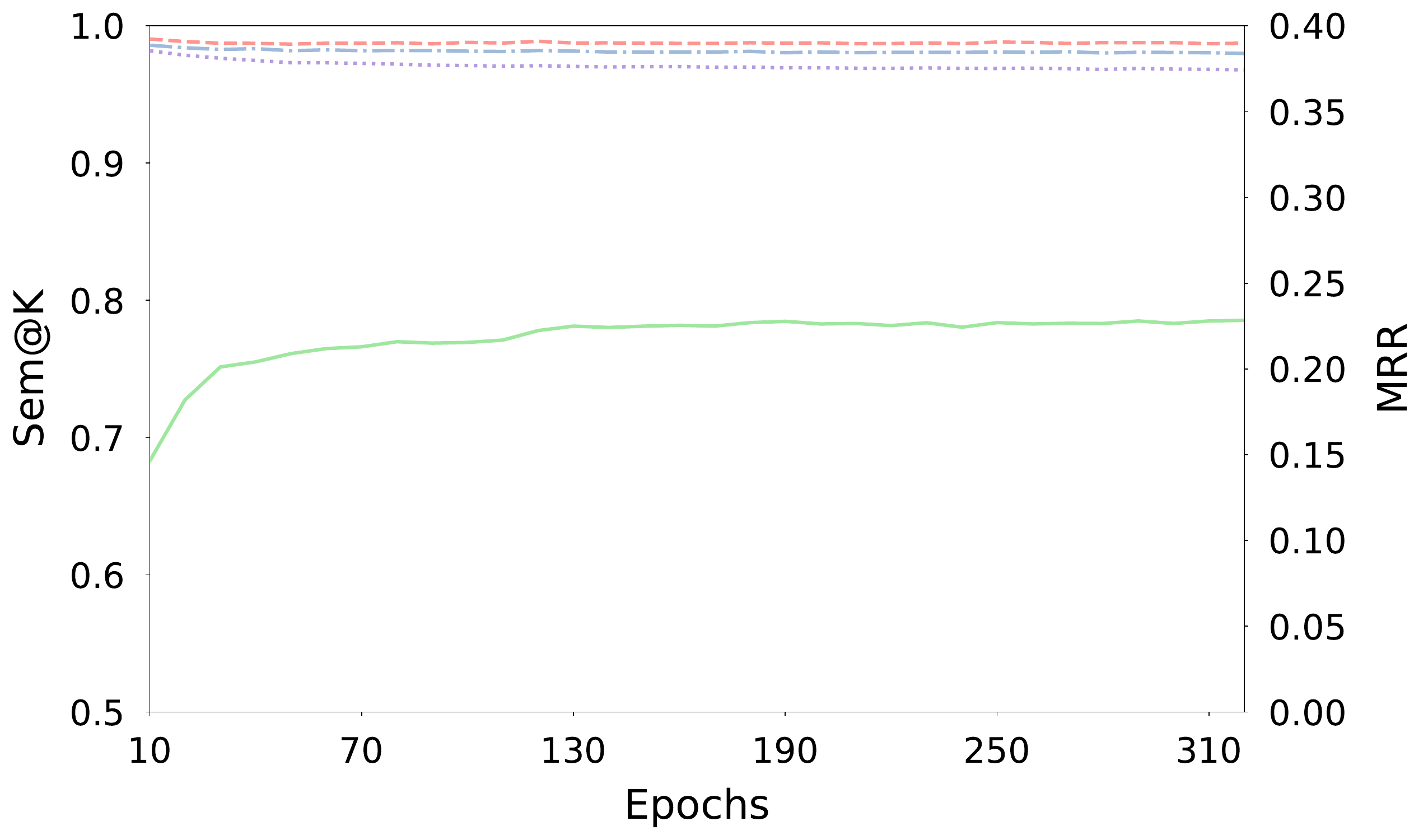}
        \caption{TransH -- DB93K}
        \label{subfig:db-transh}
    \end{subfigure}
    
    \begin{subfigure}[c]{0.49\textwidth}
        \centering
        \includegraphics[width=\textwidth]{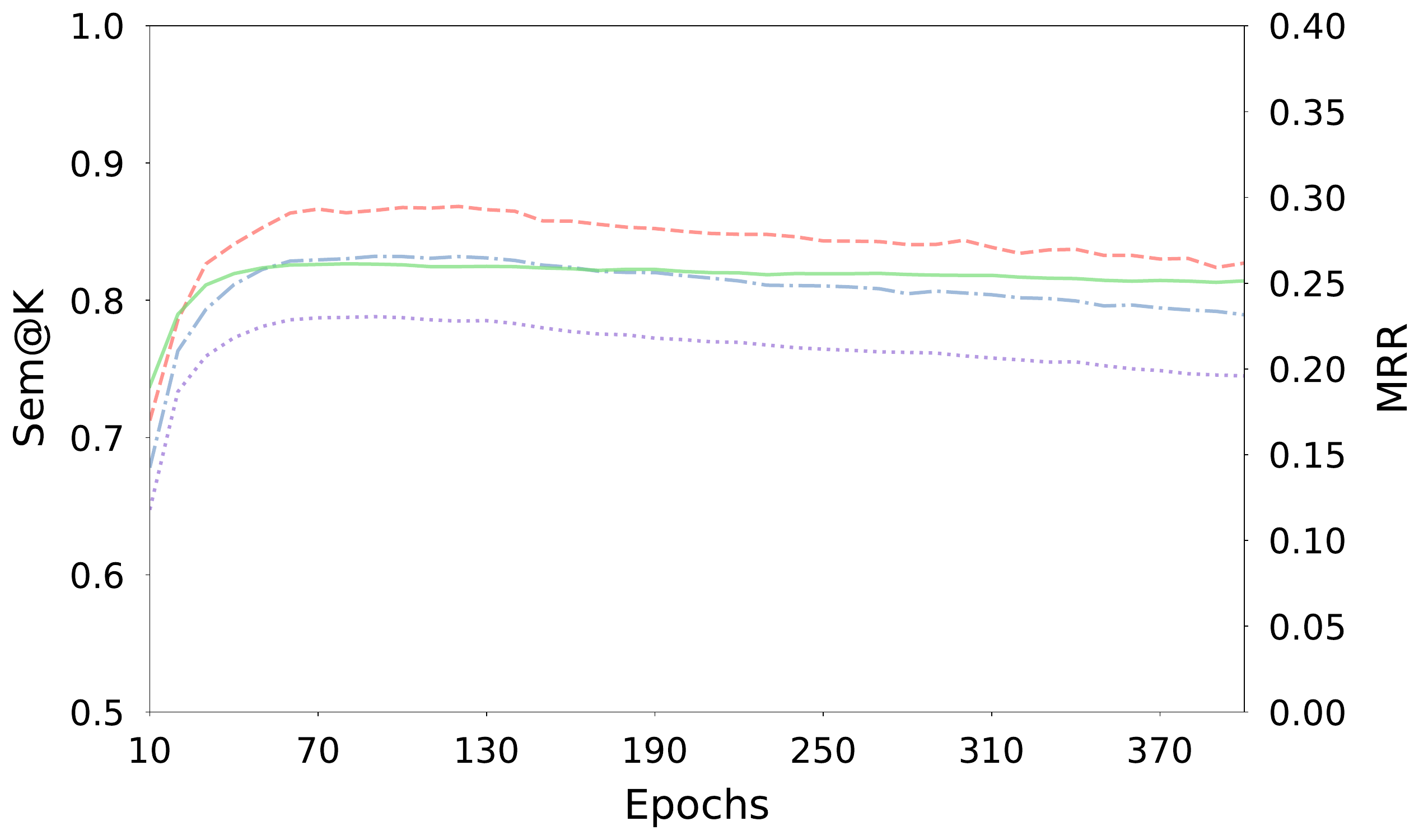}
        \caption{DistMult -- DB93K}
        \label{subfig:db-distmult}
    \end{subfigure}
    \hfill
    \begin{subfigure}[c]{0.49\textwidth}
        \centering
        \includegraphics[width=\textwidth]{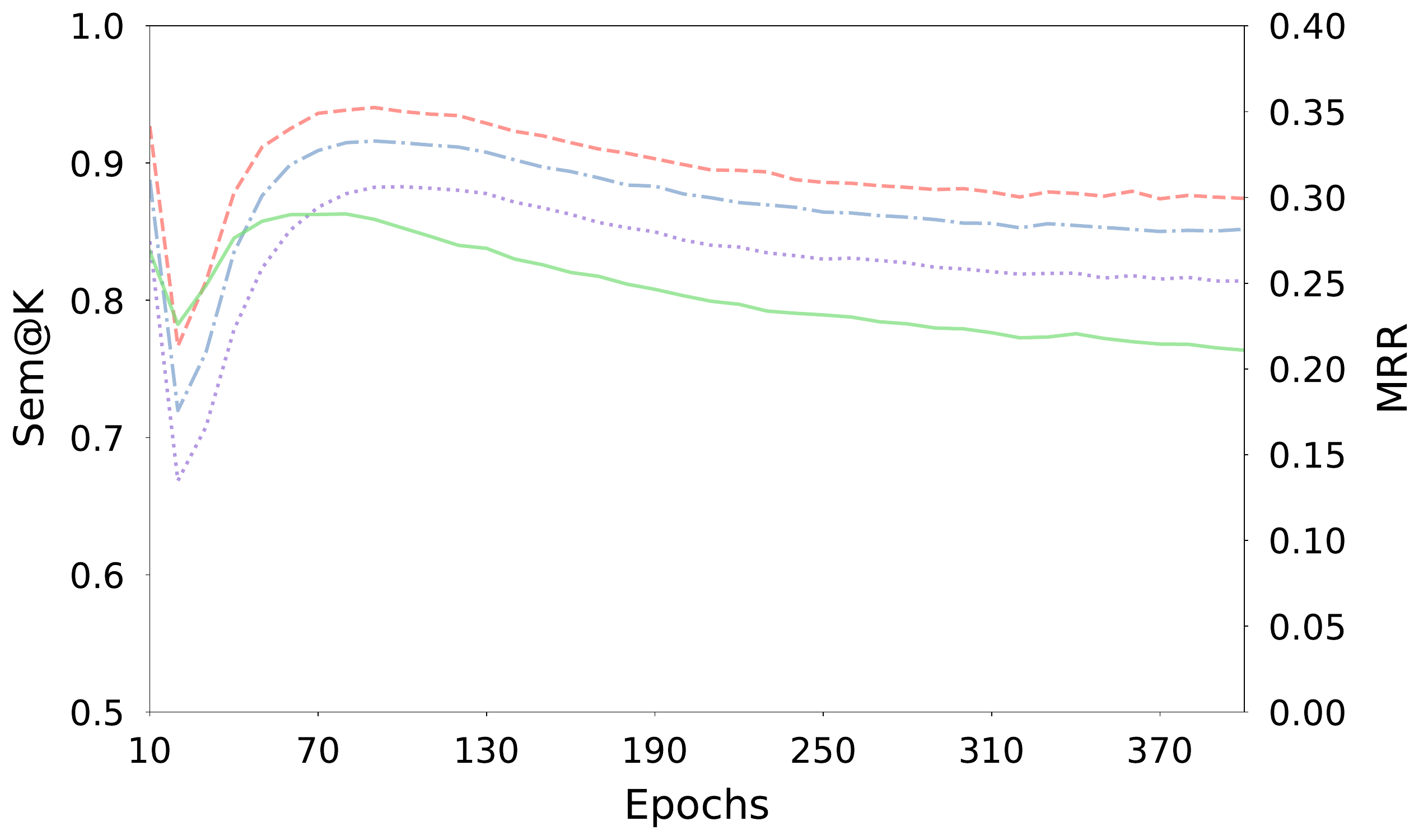}
        \caption{ComplEx -- DB93K}
        \label{subfig:db-complex}
    \end{subfigure}
    
    \begin{subfigure}[c]{0.49\textwidth}
        \centering
        \includegraphics[width=\textwidth]{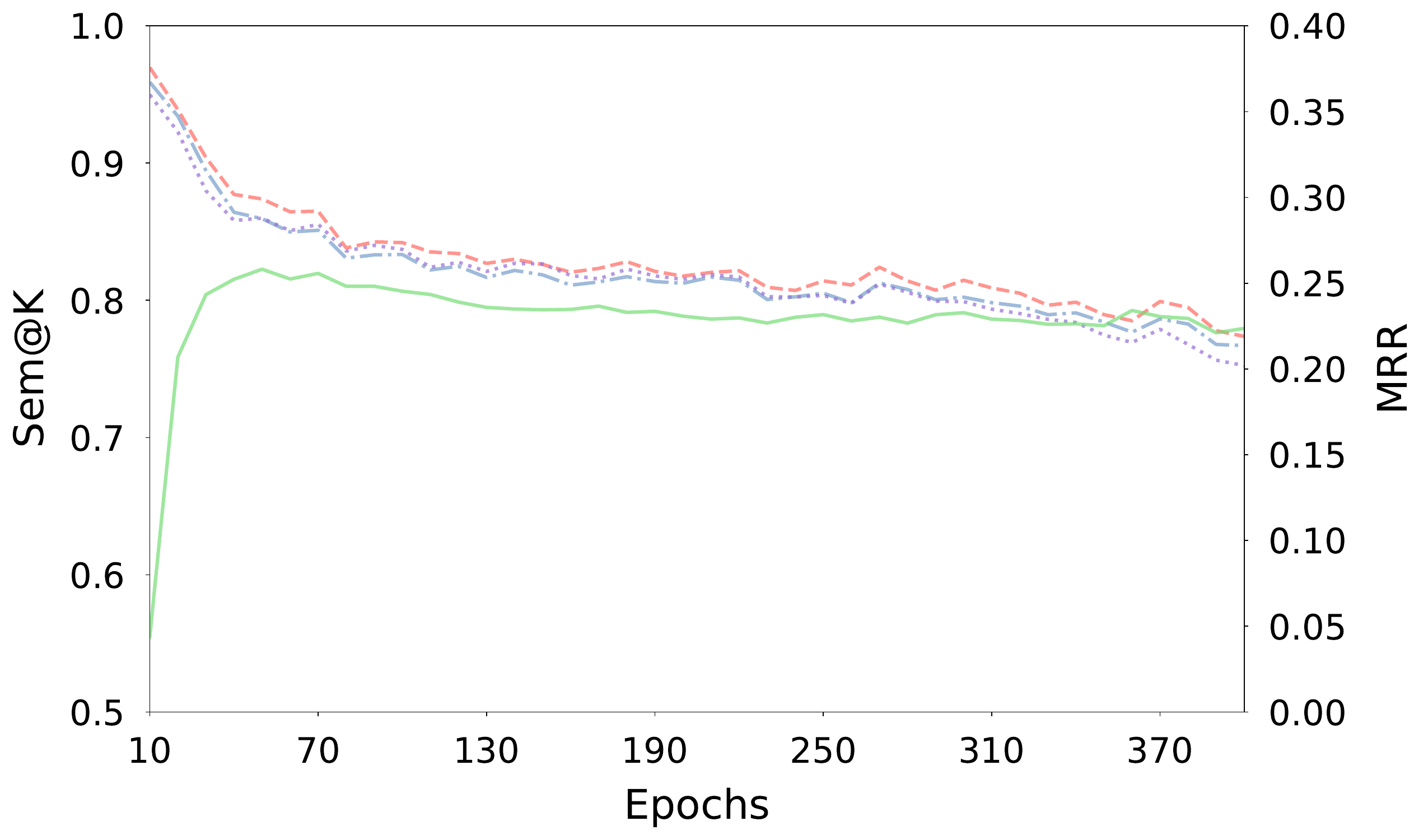}
        \caption{ConvE -- DB93K}
        \label{subfig:db-conve}
    \end{subfigure}
    \hfill
    \begin{subfigure}[c]{0.49\textwidth}
        \centering
        \includegraphics[width=\textwidth]{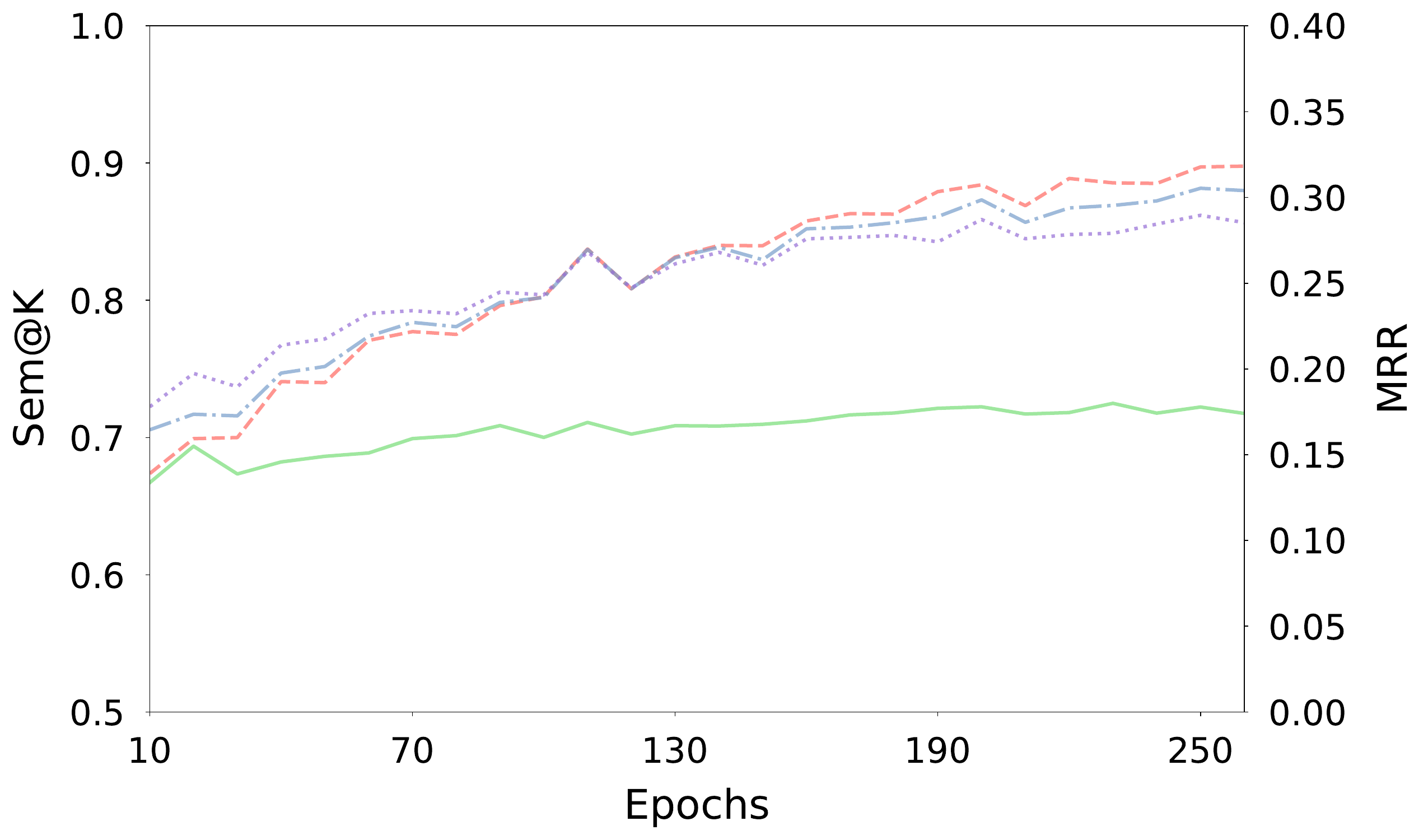}
        \caption{ConvKB -- DB93K}
        \label{subfig:db-convkb}
    \end{subfigure}

    \begin{subfigure}[c]{0.49\textwidth}
        \centering
        \includegraphics[width=\textwidth]{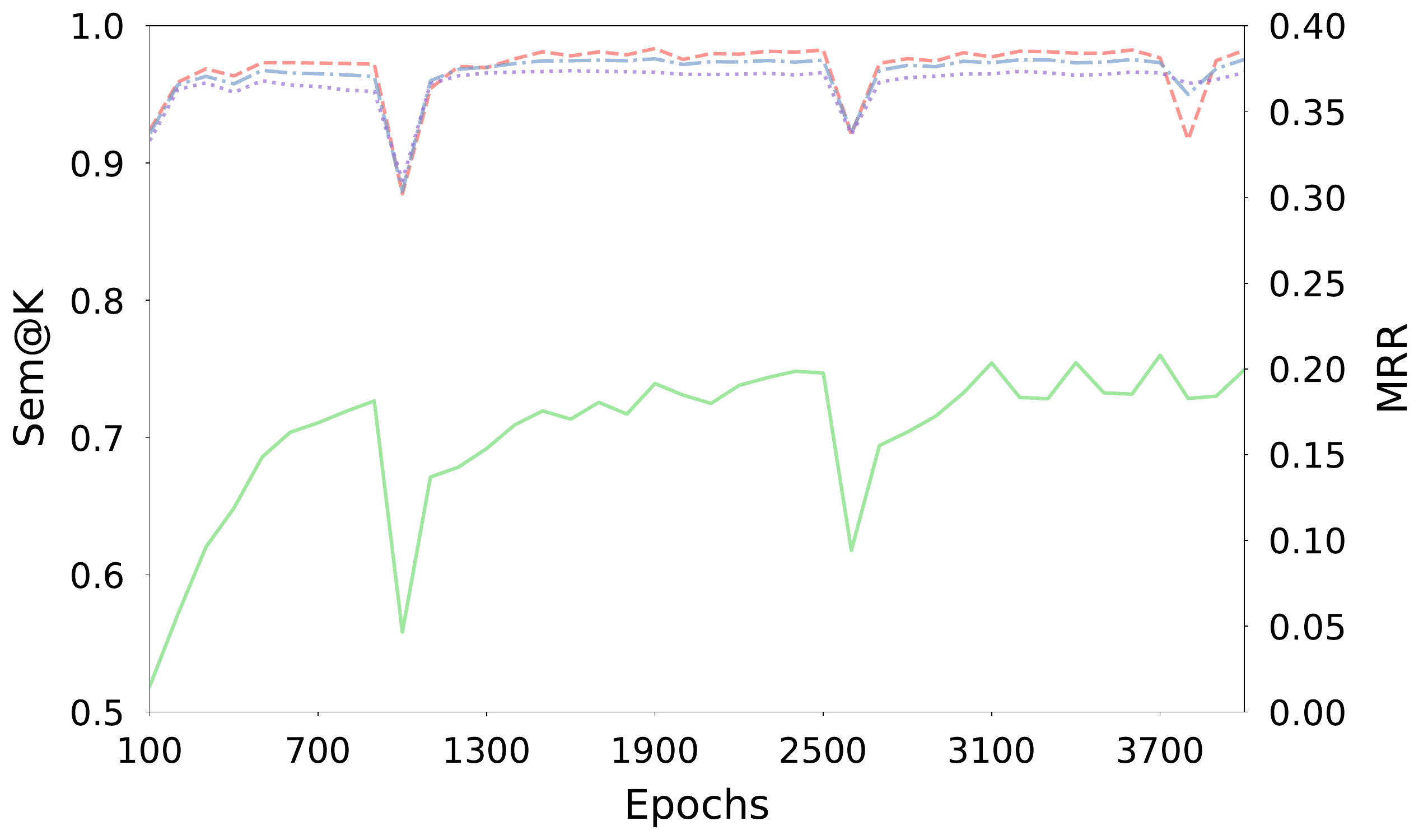}
        \caption{R-GCN -- DB93K}
        \label{subfig:db-rgcn}
    \end{subfigure}
    \hfill
    \begin{subfigure}[c]{0.49\textwidth}
        \centering
        \includegraphics[width=\textwidth]{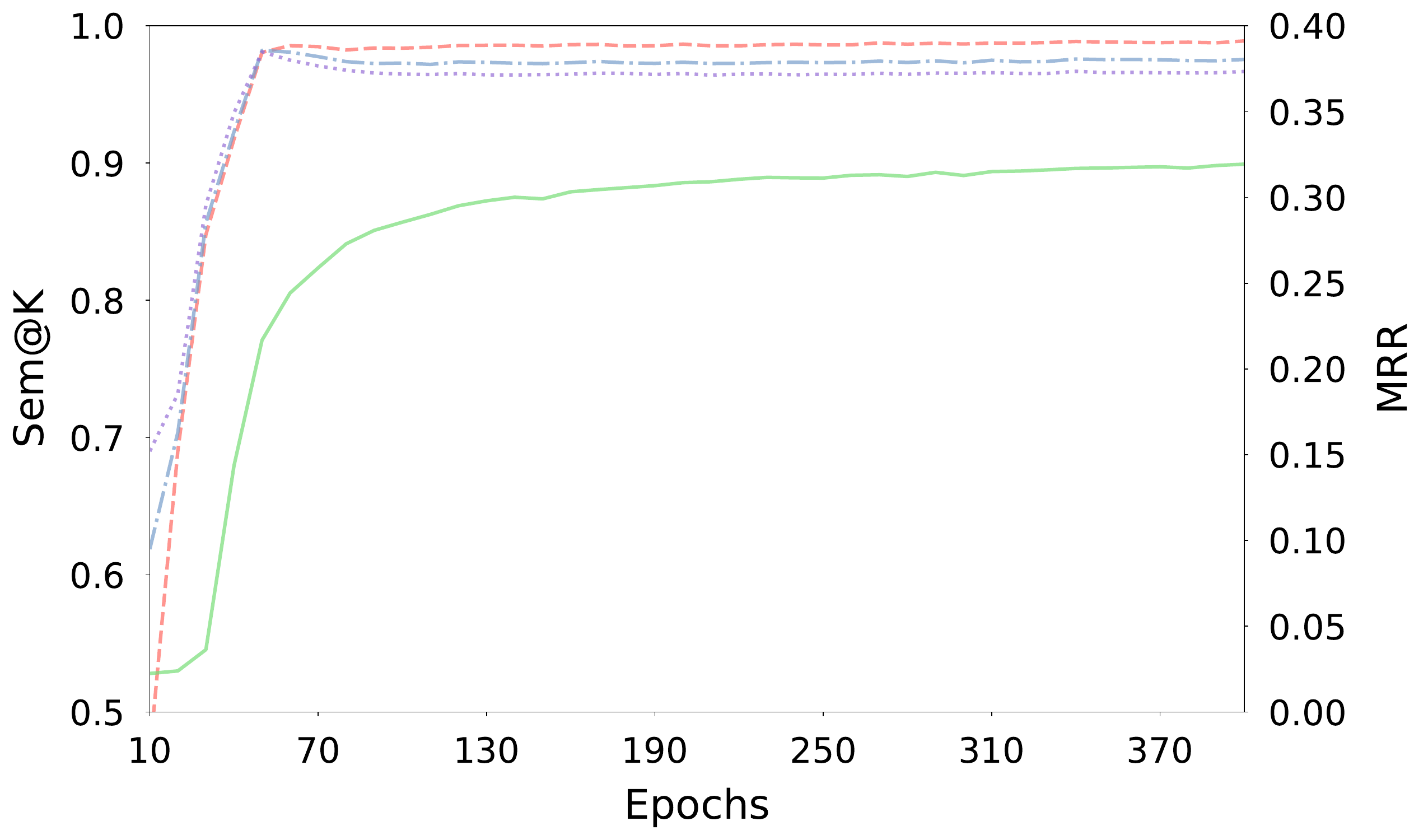}
        \caption{CompGCN -- DB93K}
        \label{subfig:db-compgcn}
    \end{subfigure}

    \caption[]{Evolution of MRR (\raisebox{2pt}{\begin{tikzpicture}[scale=0.5]
        \draw[color=pastelgreen, solid, line width=1pt] (0,0) -- (0.7,0);
    \end{tikzpicture}}), Sem@1 (\raisebox{2pt}{\begin{tikzpicture}[scale=0.5]
        \draw[color=pastelred, dashed, line width=1pt] (0,0) -- (0.65,0);
    \end{tikzpicture}}), Sem@3 (\raisebox{2pt}{\begin{tikzpicture}[scale=0.5]
        \draw[color=darkpastelblue, dash dot, line width=1pt] (0,0) -- (0.7,0);
    \end{tikzpicture}}), and Sem@10 (\raisebox{2pt}{\begin{tikzpicture}[scale=0.5]
        \draw[color=darkpastelpurple, dotted, line width=1pt] (0,0) -- (0.7,0);
    \end{tikzpicture}}) on DB93K}
    \label{fig:results2}
\end{figure}

\begin{figure}[h]
    \centering
    \begin{subfigure}[c]{0.49\textwidth}
        \centering
        \includegraphics[width=\textwidth]{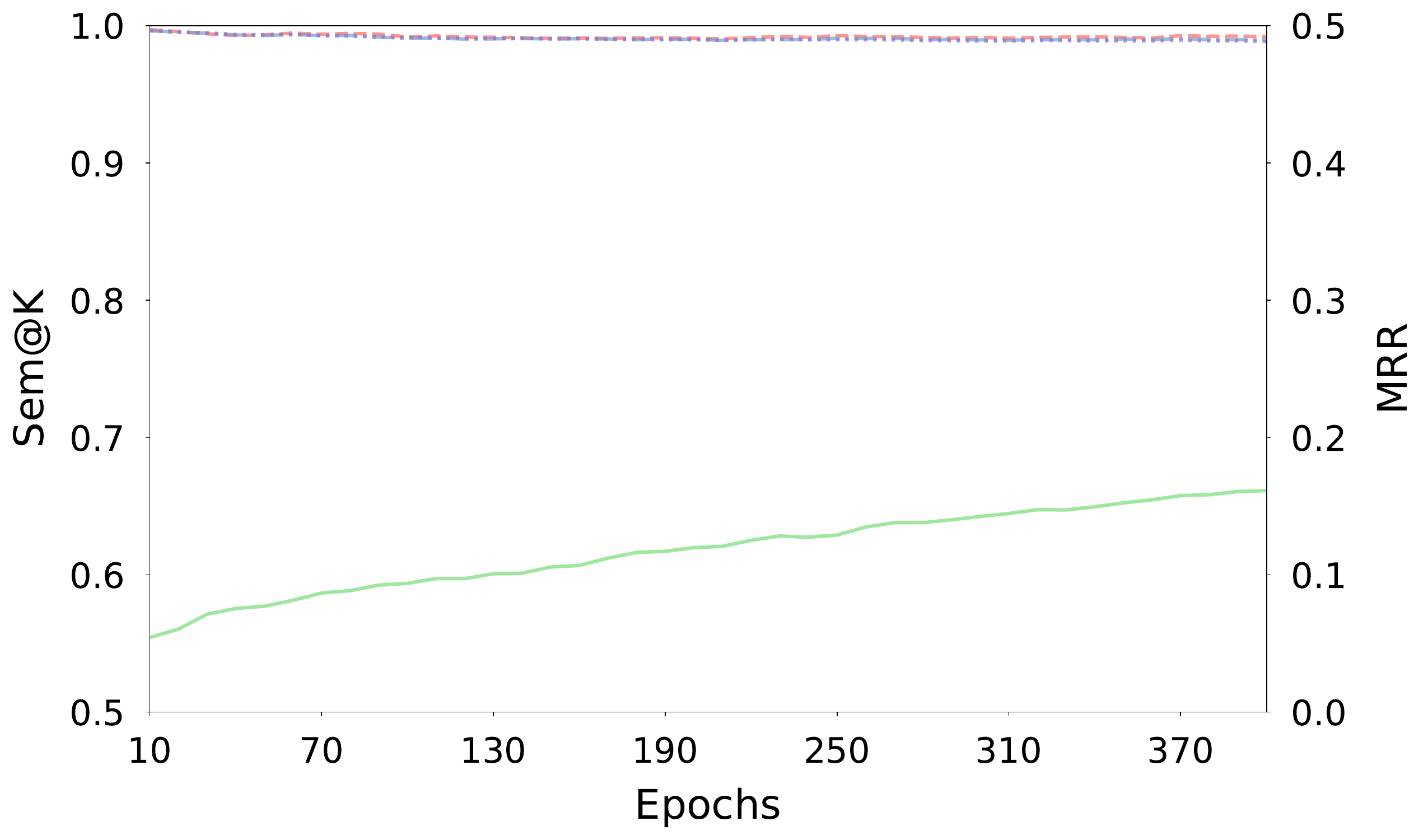}
        \caption{TransE -- YAGO3-37K}
        \label{subfig:yago3-transe}
    \end{subfigure}
    \hfill
    \begin{subfigure}[c]{0.49\textwidth}
        \centering
        \includegraphics[width=\textwidth]{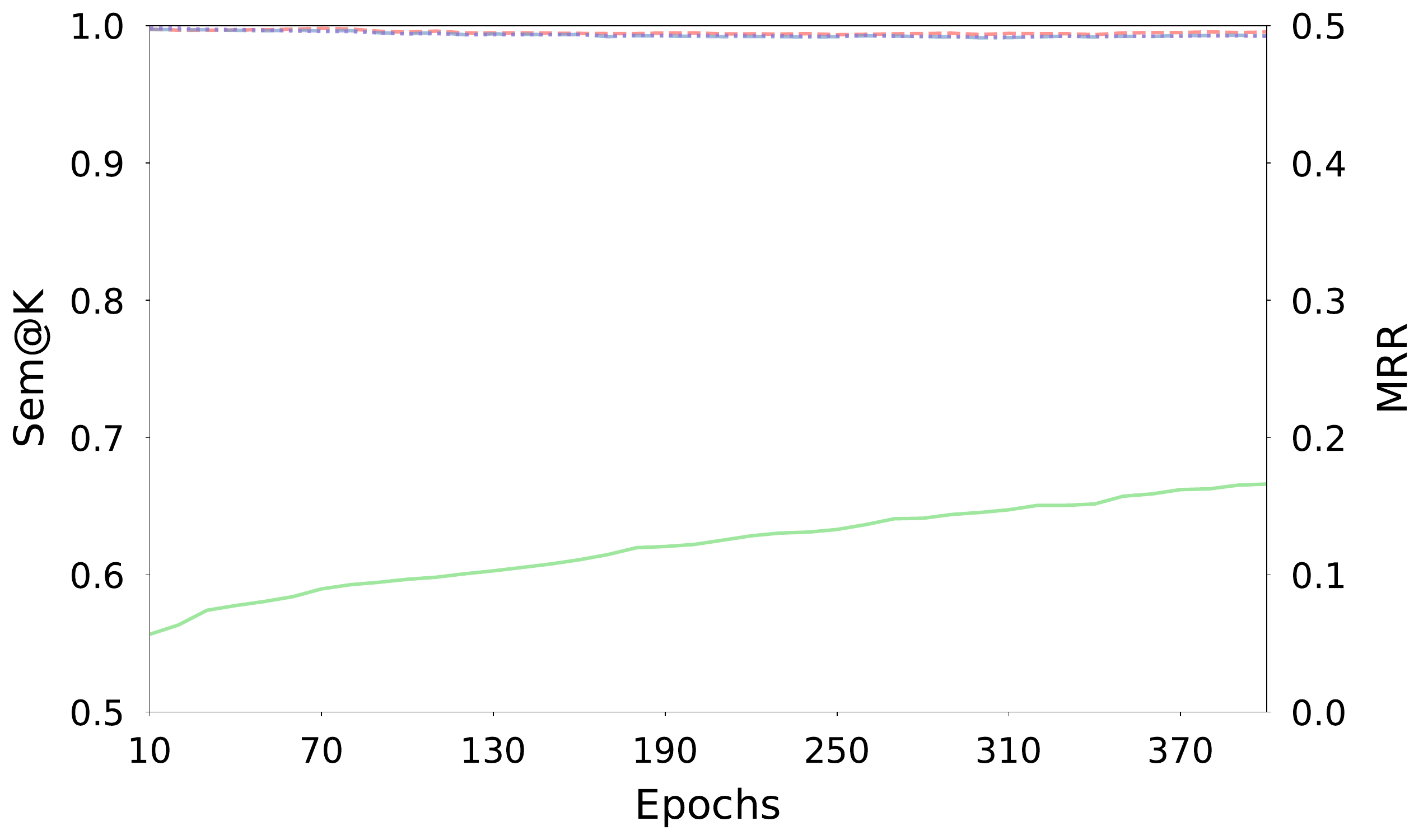}
        \caption{TransH -- YAGO3-37K}
        \label{subfig:yago3-transh}
    \end{subfigure}
    
    \begin{subfigure}[c]{0.49\textwidth}
        \centering
        \includegraphics[width=\textwidth]{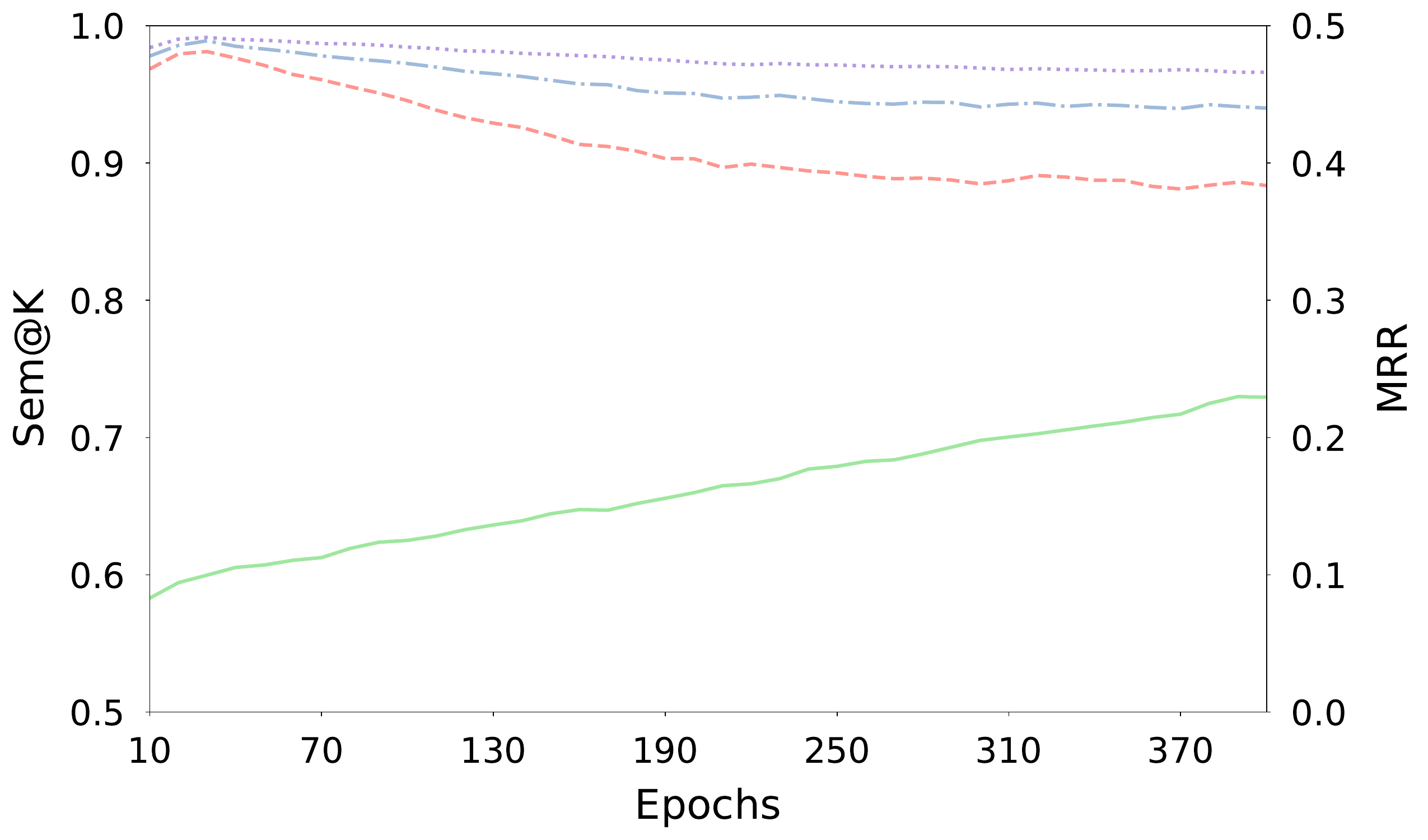}
        \caption{DistMult -- YAGO3-37K}
        \label{subfig:yago3-distmult}
    \end{subfigure}
    \hfill
    \begin{subfigure}[c]{0.49\textwidth}
        \centering
        \includegraphics[width=\textwidth]{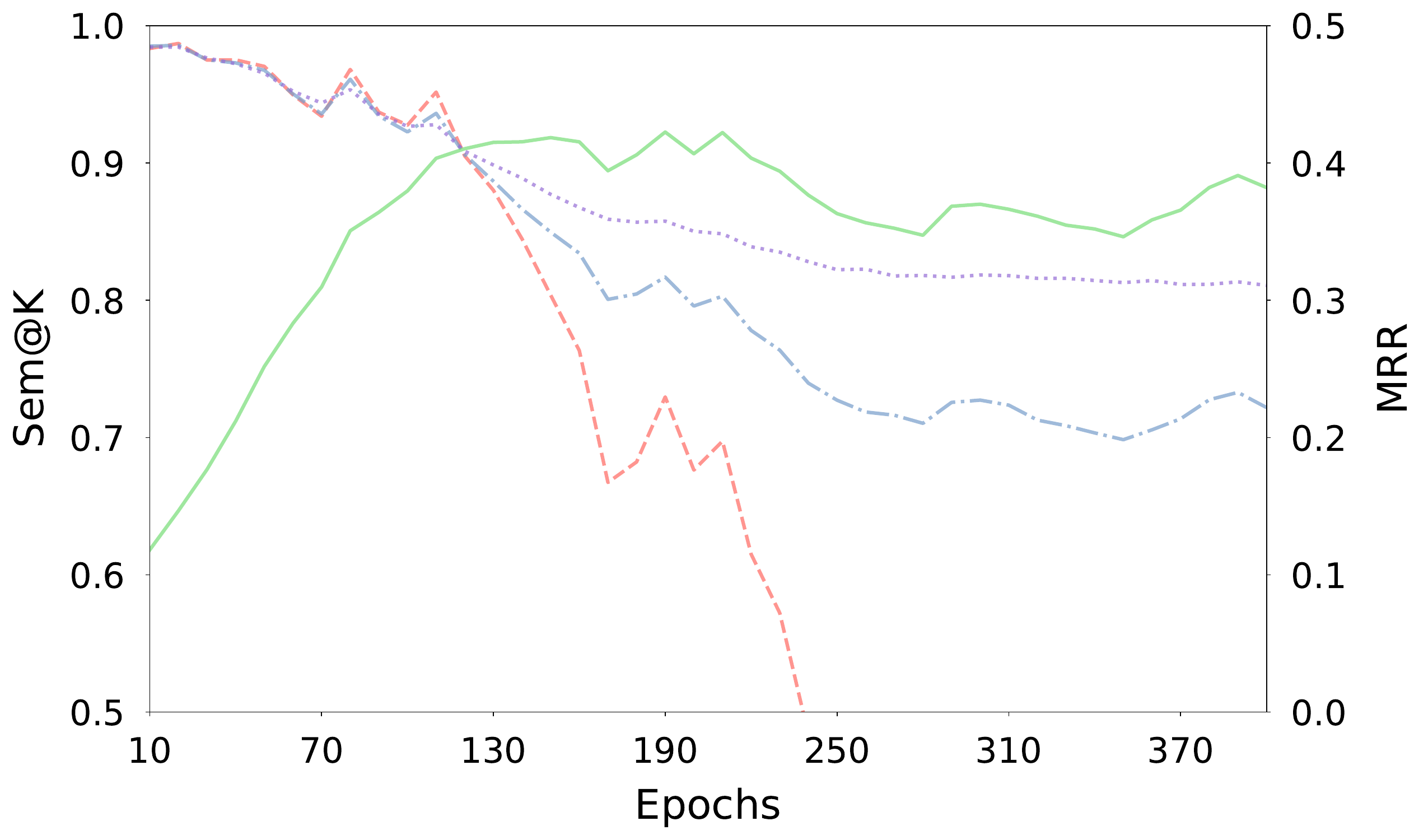}
        \caption{ComplEx -- YAGO3-37K}
        \label{subfig:yago3-complex}
    \end{subfigure}
    
    \begin{subfigure}[c]{0.49\textwidth}
        \centering
        \includegraphics[width=\textwidth]{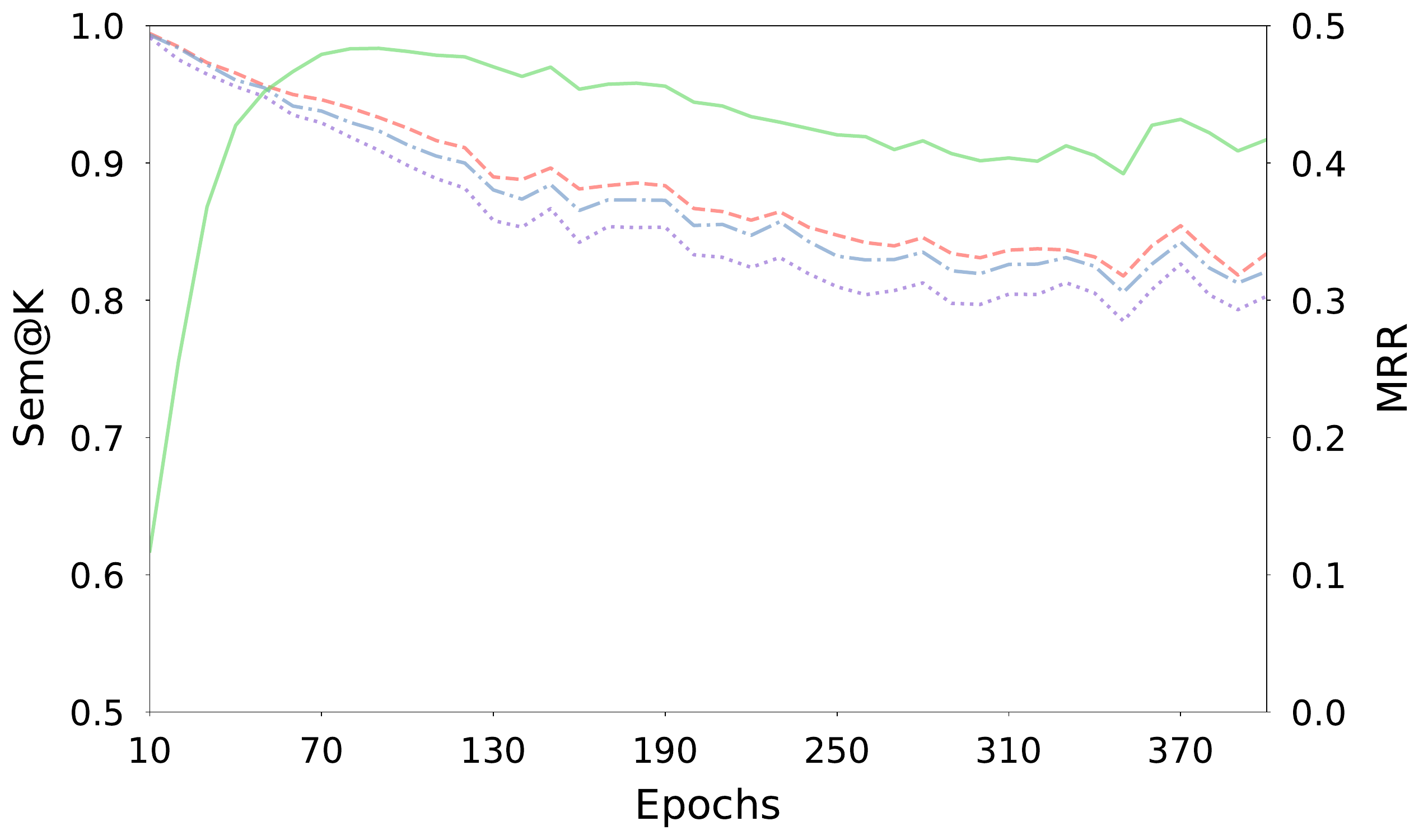}
        \caption{ConvE -- YAGO3-37K}
        \label{subfig:yago3-conve}
    \end{subfigure}
    \hfill
    \begin{subfigure}[c]{0.49\textwidth}
        \centering
        \includegraphics[width=\textwidth]{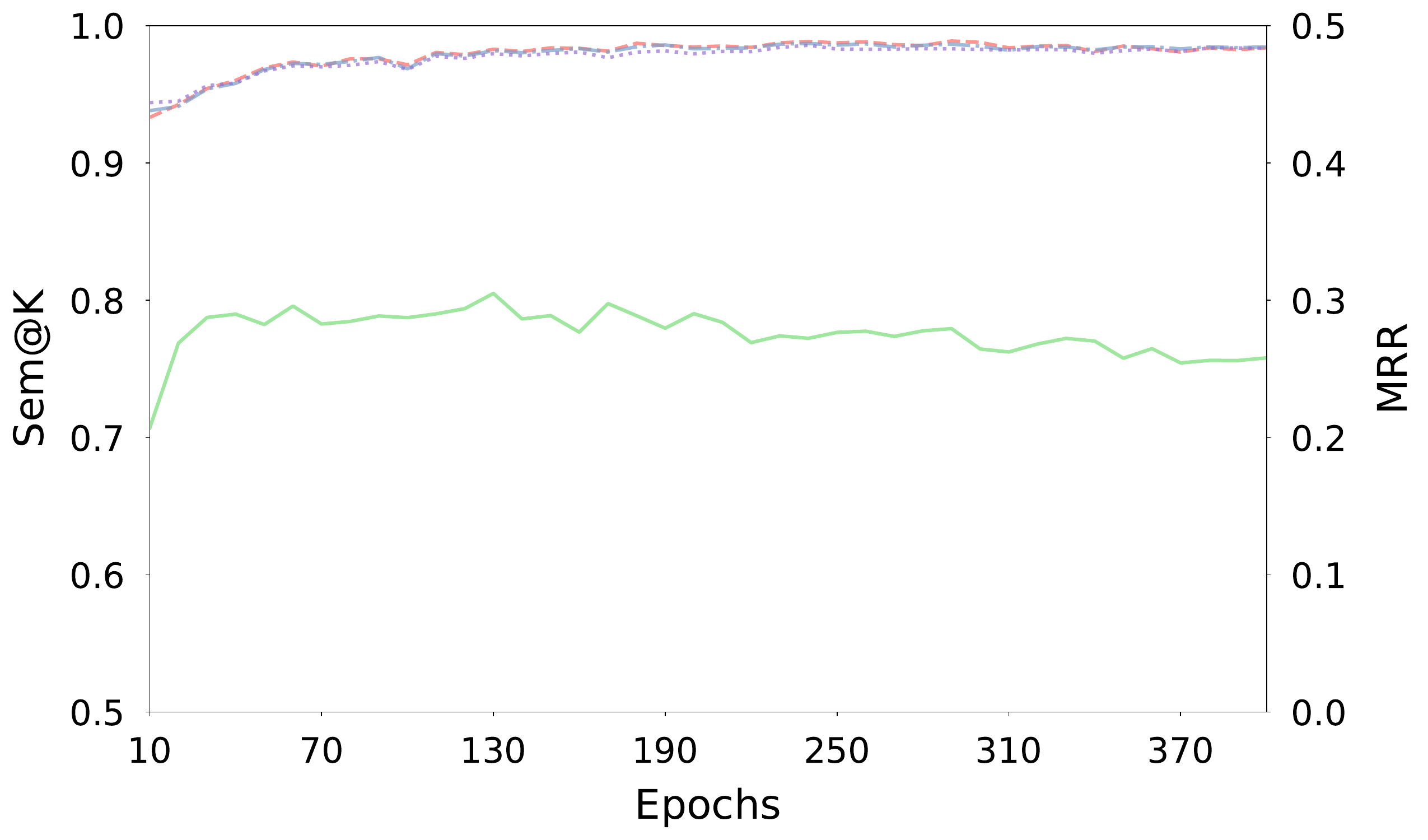}
        \caption{ConvKB -- YAGO3-37K}
        \label{subfig:yago3-convkb}
    \end{subfigure}

    \begin{subfigure}[c]{0.49\textwidth}
        \centering
        \includegraphics[width=\textwidth]{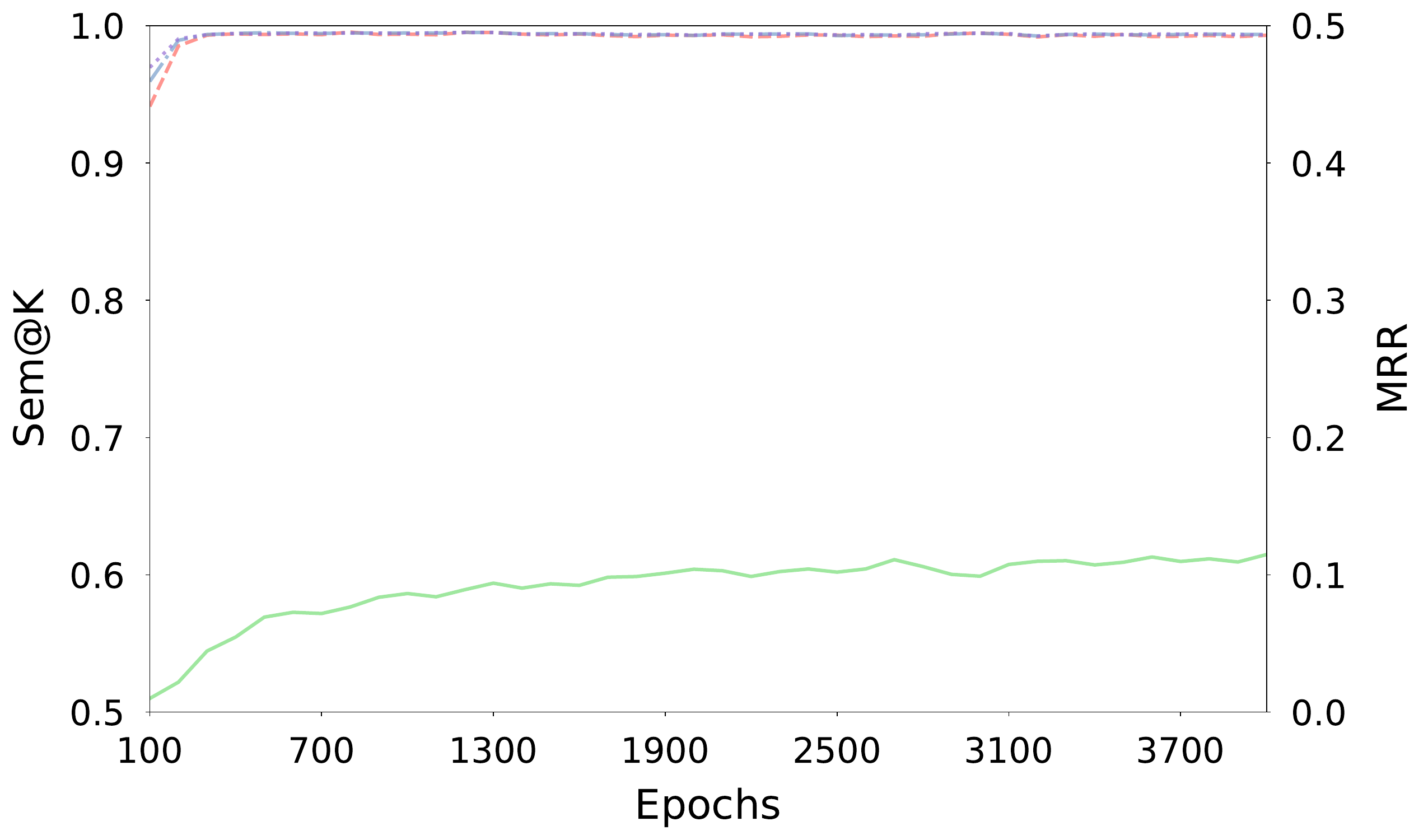}
        \caption{R-GCN -- YAGO3-37K}
        \label{subfig:yago3-rgcn}
    \end{subfigure}
    \hfill
    \begin{subfigure}[c]{0.49\textwidth}
        \centering
        \includegraphics[width=\textwidth]{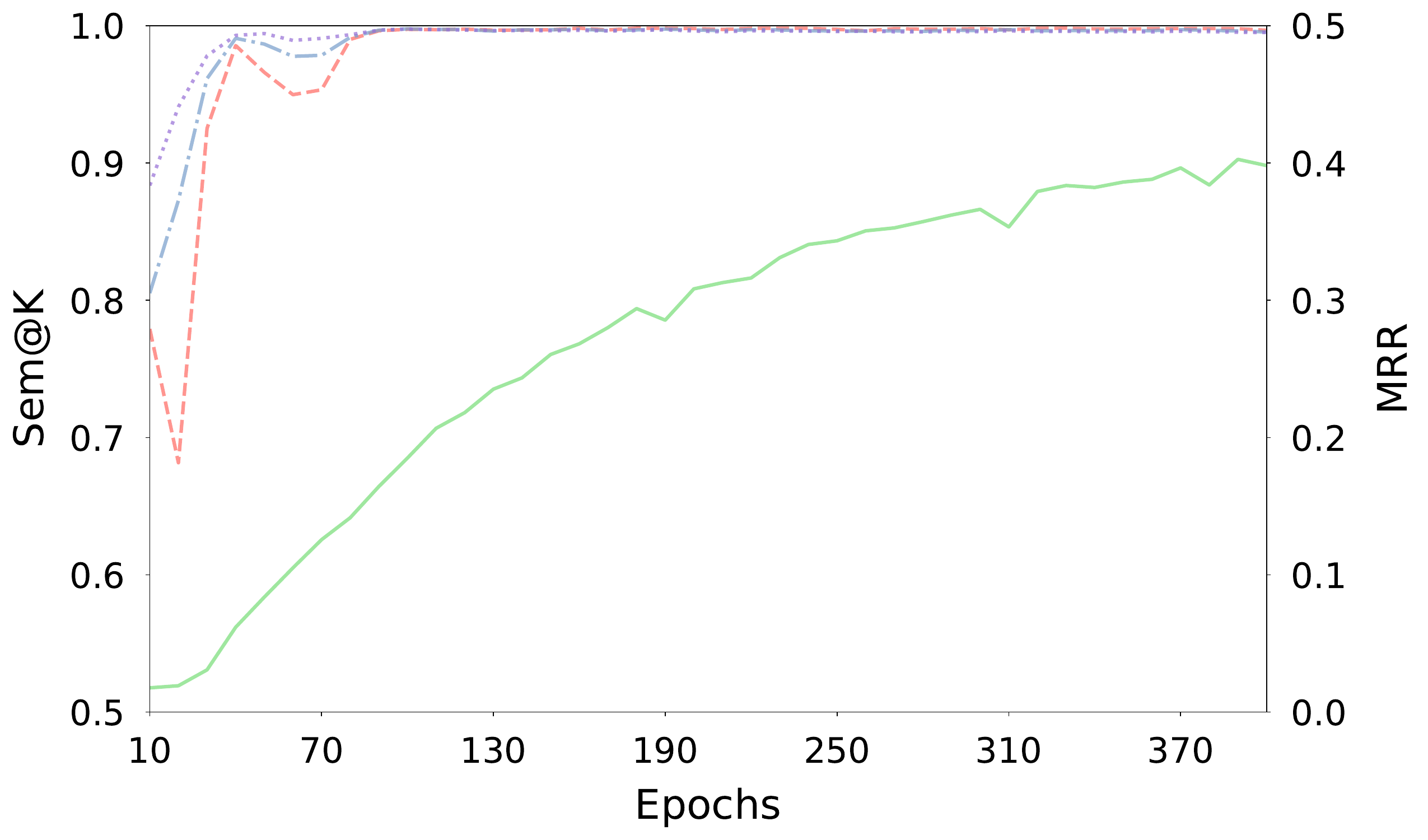}
        \caption{CompGCN -- YAGO3-37K}
        \label{subfig:yago3-compgcn}
    \end{subfigure}

    \caption[]{Evolution of MRR (\raisebox{2pt}{\begin{tikzpicture}[scale=0.5]
        \draw[color=pastelgreen, solid, line width=1pt] (0,0) -- (0.7,0);
    \end{tikzpicture}}), Sem@1 (\raisebox{2pt}{\begin{tikzpicture}[scale=0.5]
        \draw[color=pastelred, dashed, line width=1pt] (0,0) -- (0.65,0);
    \end{tikzpicture}}), Sem@3 (\raisebox{2pt}{\begin{tikzpicture}[scale=0.5]
        \draw[color=darkpastelblue, dash dot, line width=1pt] (0,0) -- (0.7,0);
    \end{tikzpicture}}), and Sem@10 (\raisebox{2pt}{\begin{tikzpicture}[scale=0.5]
        \draw[color=darkpastelpurple, dotted, line width=1pt] (0,0) -- (0.7,0);
    \end{tikzpicture}}) on YAGO3-37K}
    \label{fig:results3}
\end{figure}

\begin{figure}[h]
    \centering
    \begin{subfigure}[c]{0.49\textwidth}
        \centering
        \includegraphics[width=\textwidth]{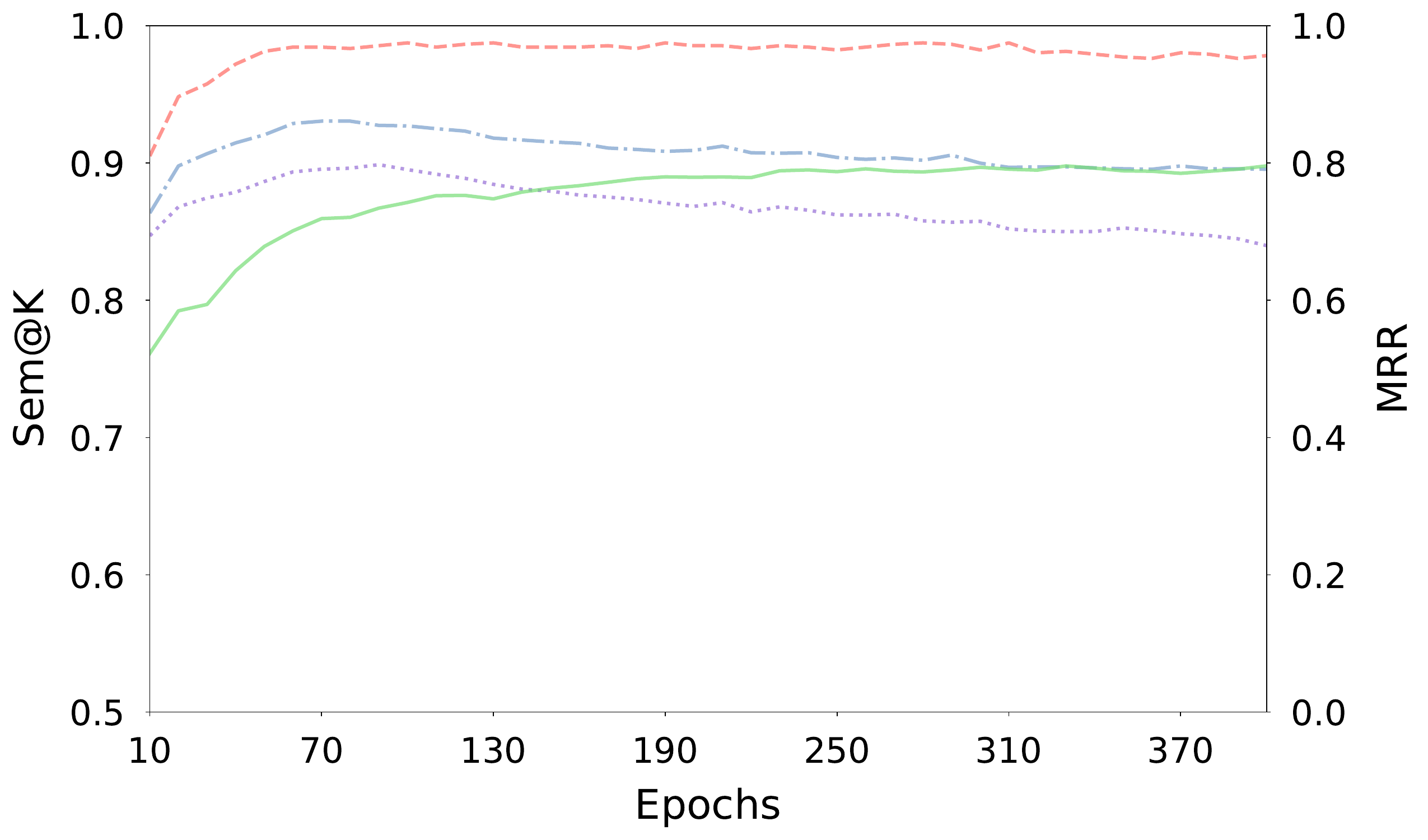}
        \caption{TransE -- YAGO4-19K}
        \label{subfig:yago4-transe}
    \end{subfigure}
    \hfill
    \begin{subfigure}[c]{0.49\textwidth}
        \centering
        \includegraphics[width=\textwidth]{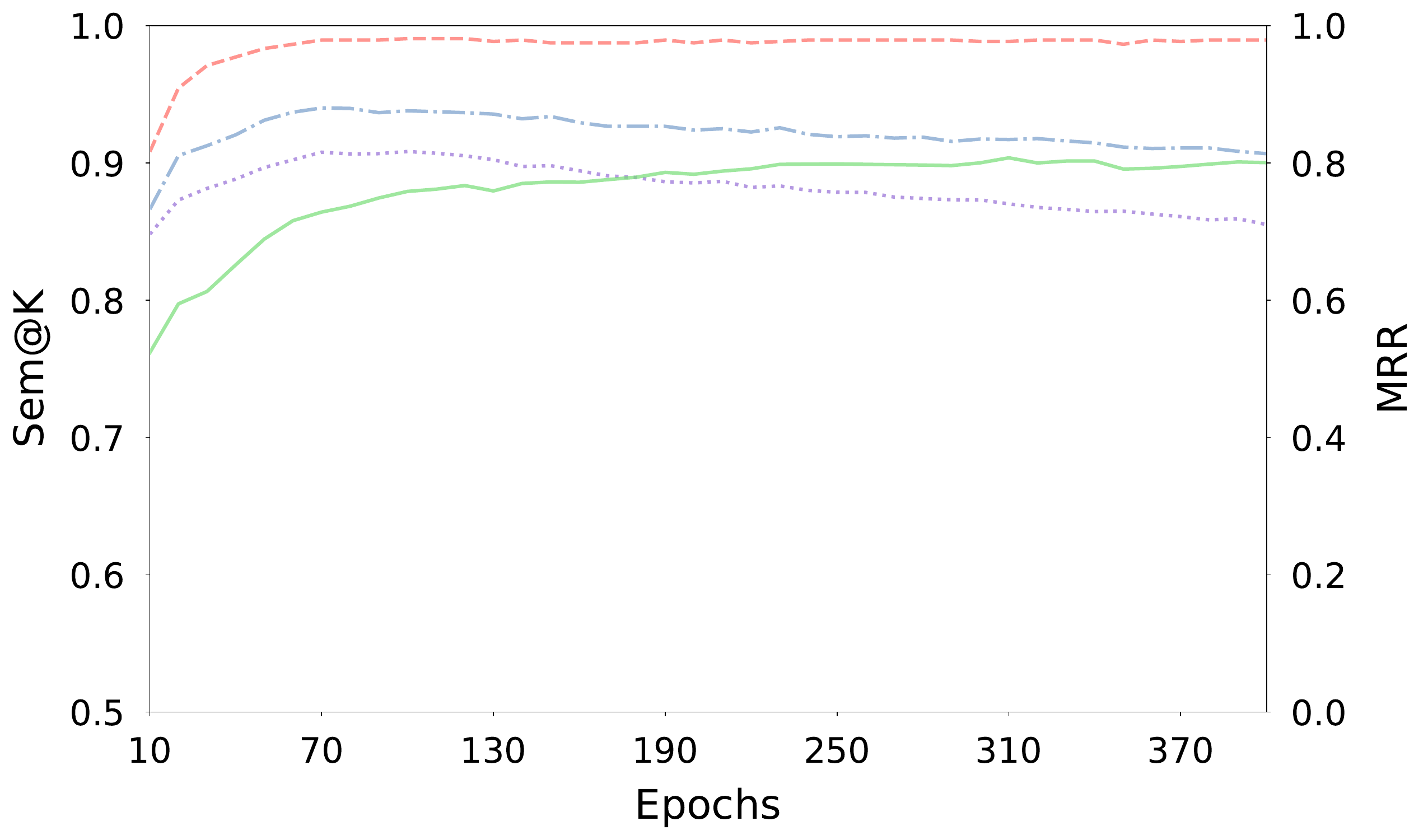}
        \caption{TransH -- YAGO4-19K}
        \label{subfig:yago4-transh}
    \end{subfigure}
    
    \begin{subfigure}[c]{0.49\textwidth}
        \centering
        \includegraphics[width=\textwidth]{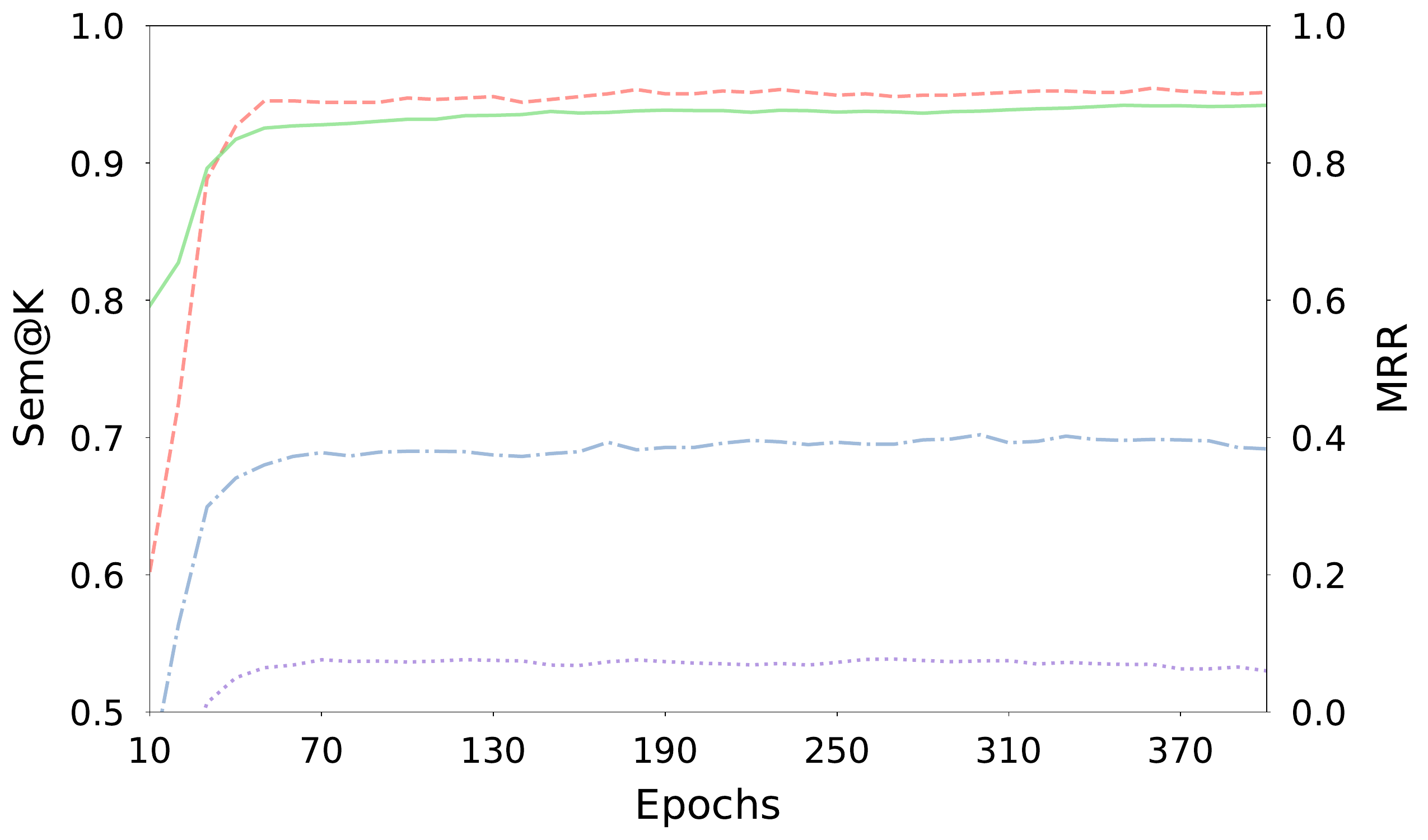}
        \caption{DistMult -- YAGO4-19K}
        \label{subfig:yago4-distmult}
    \end{subfigure}
    \hfill
    \begin{subfigure}[c]{0.49\textwidth}
        \centering
        \includegraphics[width=\textwidth]{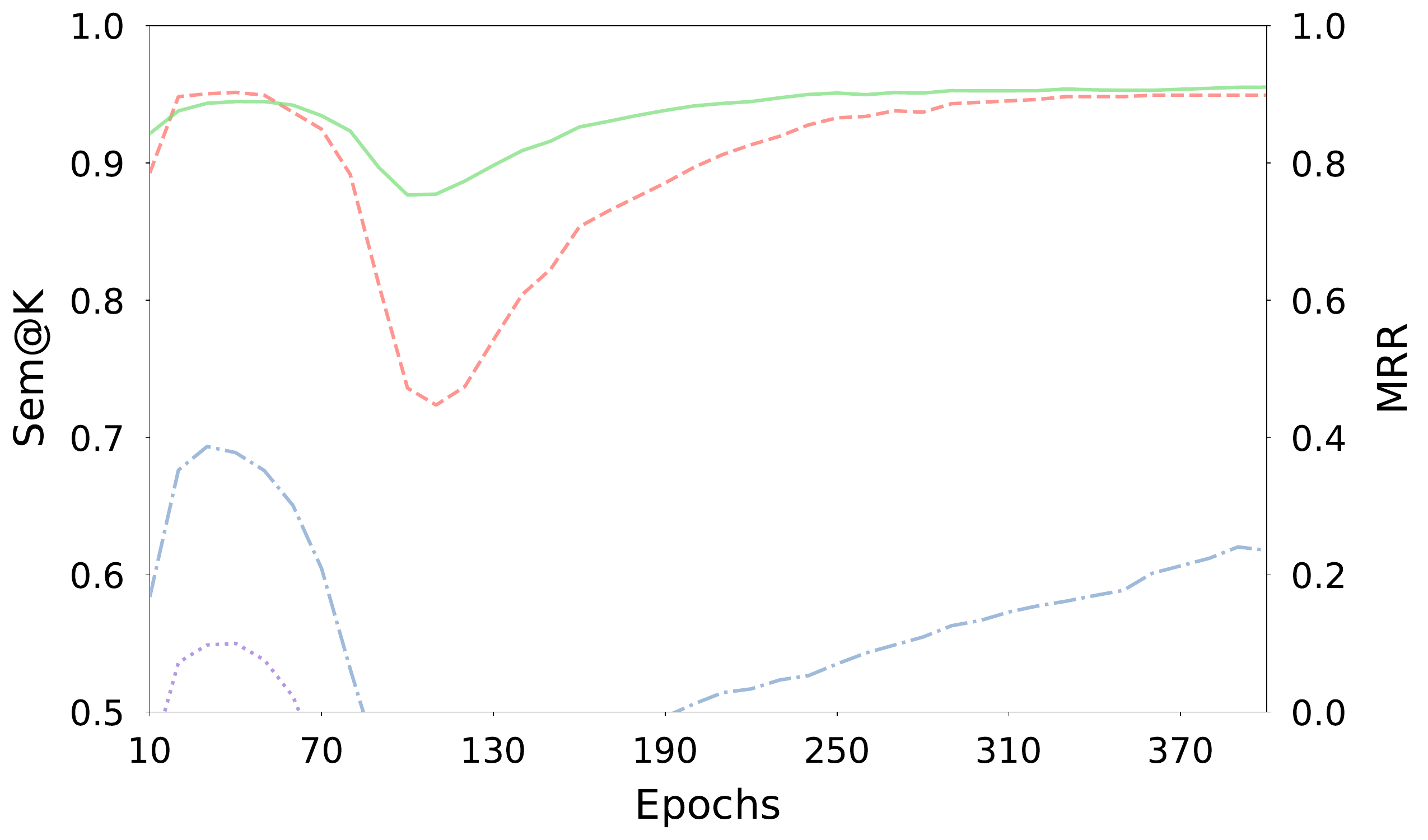}
        \caption{ComplEx -- YAGO4-19K}
        \label{subfig:yago4-complex}
    \end{subfigure}
    
    \begin{subfigure}[c]{0.49\textwidth}
        \centering
        \includegraphics[width=\textwidth]{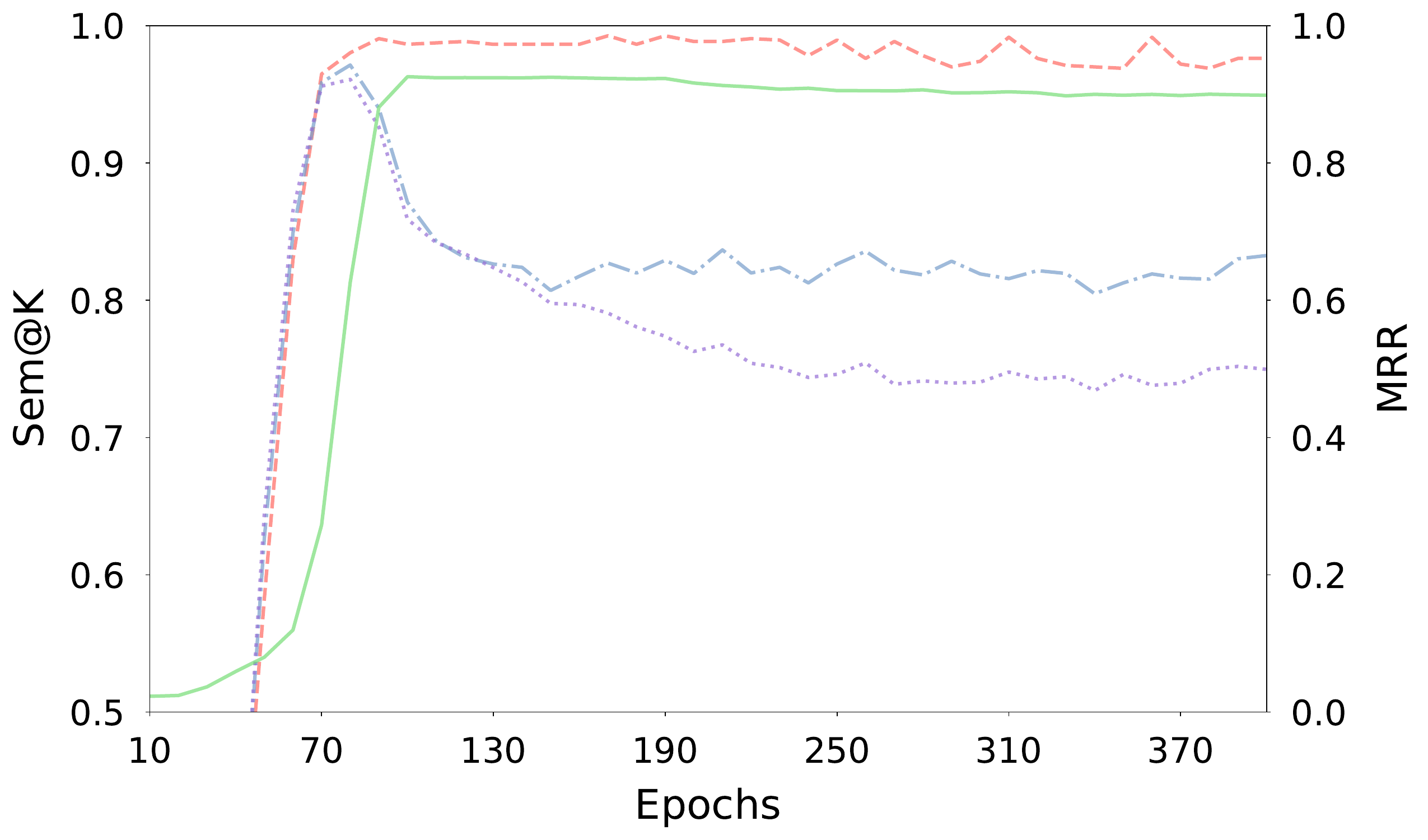}
        \caption{ConvE -- YAGO4-19K}
        \label{subfig:yago4-conve}
    \end{subfigure}
    \hfill
    \begin{subfigure}[c]{0.49\textwidth}
        \centering
        \includegraphics[width=\textwidth]{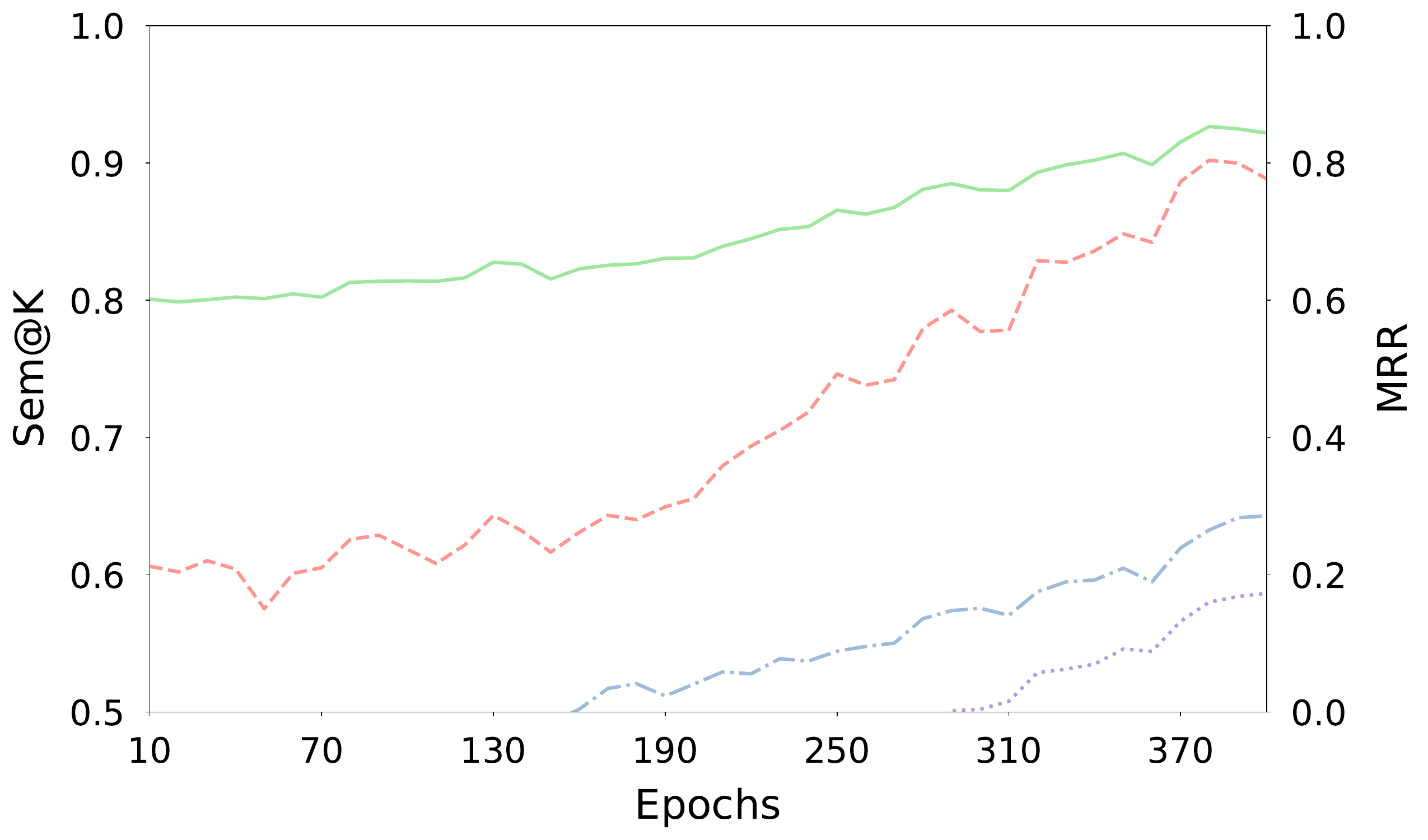}
        \caption{ConvKB -- YAGO4-19K}
        \label{subfig:yago4-convkb}
    \end{subfigure}

    \begin{subfigure}[c]{0.49\textwidth}
        \centering
        \includegraphics[width=\textwidth]{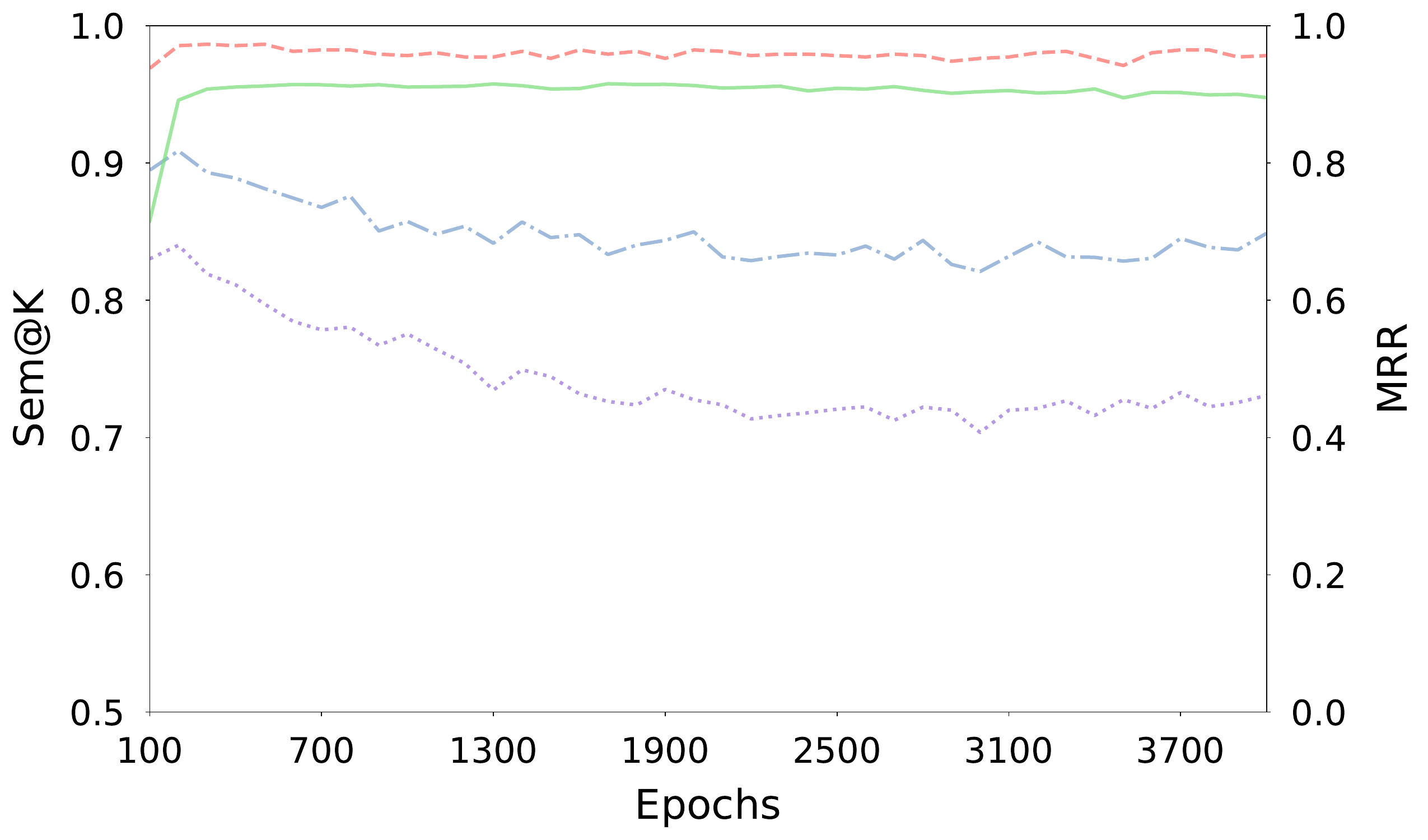}
        \caption{R-GCN -- YAGO4-19K}
        \label{subfig:yago4-rgcn}
    \end{subfigure}
    \hfill
    \begin{subfigure}[c]{0.49\textwidth}
        \centering
        \includegraphics[width=\textwidth]{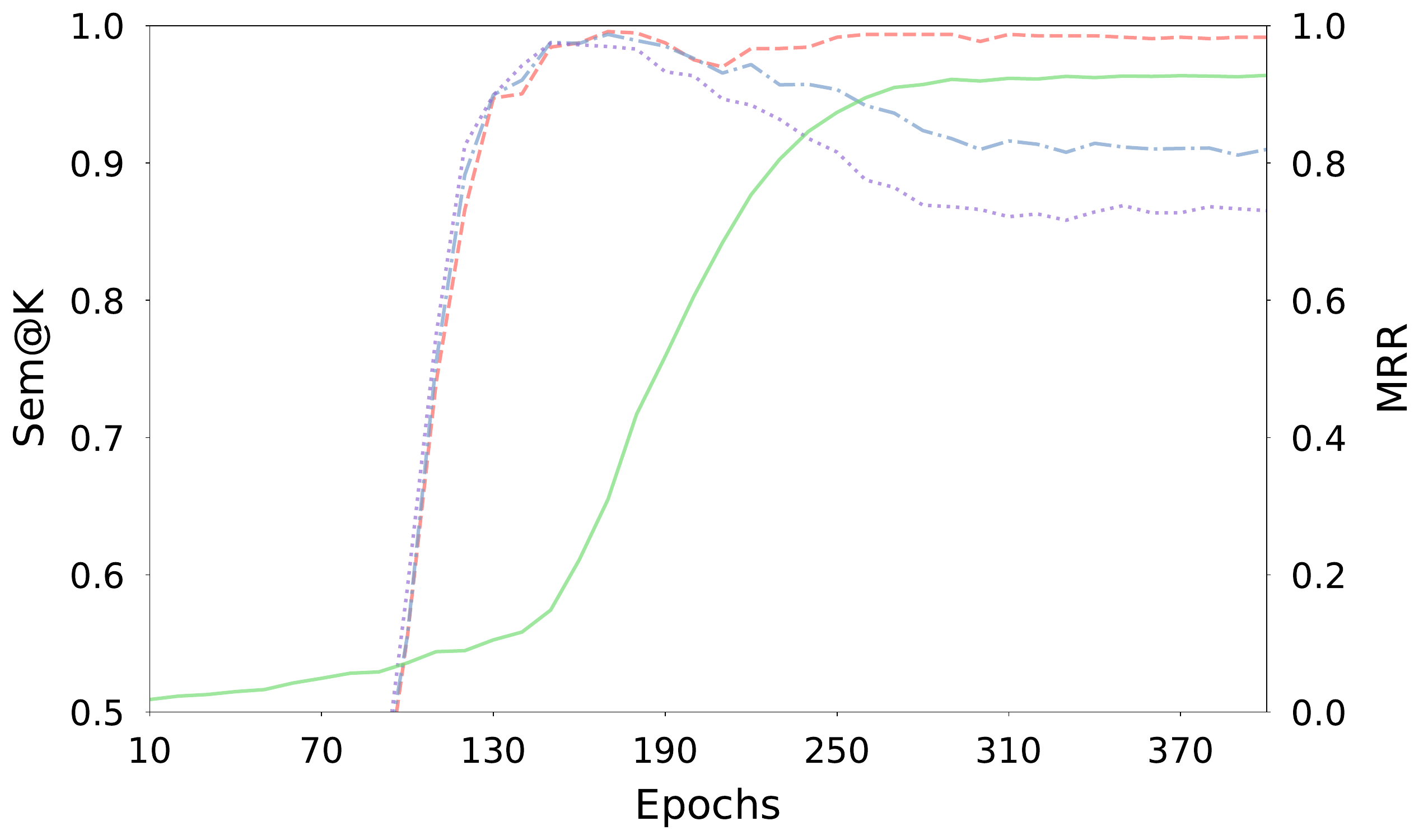}
        \caption{CompGCN -- YAGO4-19K}
        \label{subfig:yago4-compgcn}
    \end{subfigure}

    \caption[]{Evolution of MRR (\raisebox{2pt}{\begin{tikzpicture}[scale=0.5]
        \draw[color=pastelgreen, solid, line width=1pt] (0,0) -- (0.7,0);
    \end{tikzpicture}}), Sem@1 (\raisebox{2pt}{\begin{tikzpicture}[scale=0.5]
        \draw[color=pastelred, dashed, line width=1pt] (0,0) -- (0.65,0);
    \end{tikzpicture}}), Sem@3 (\raisebox{2pt}{\begin{tikzpicture}[scale=0.5]
        \draw[color=darkpastelblue, dash dot, line width=1pt] (0,0) -- (0.7,0);
    \end{tikzpicture}}), and Sem@10 (\raisebox{2pt}{\begin{tikzpicture}[scale=0.5]
        \draw[color=darkpastelpurple, dotted, line width=1pt] (0,0) -- (0.7,0);
    \end{tikzpicture}}) on YAGO4-19K}
    \label{fig:results4}
\end{figure}

\begin{figure}[h]
    \centering
    \begin{subfigure}[c]{0.49\textwidth}
        \centering
        \includegraphics[width=\textwidth]{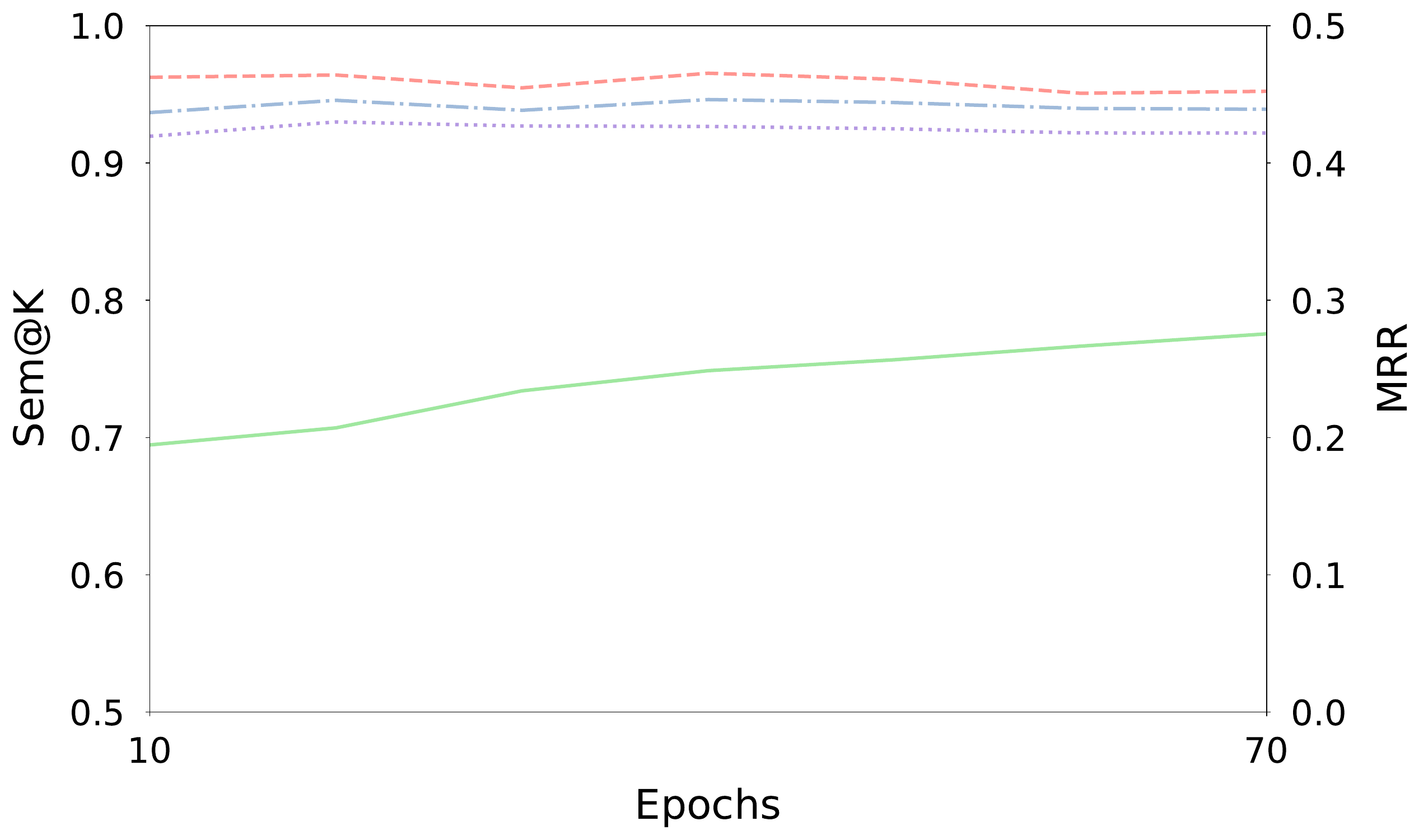}
        \caption{TransE -- Codex-S}
        \label{subfig:codexs-transe}
    \end{subfigure}
    \hfill
    \begin{subfigure}[c]{0.49\textwidth}
        \centering
        \includegraphics[width=\textwidth]{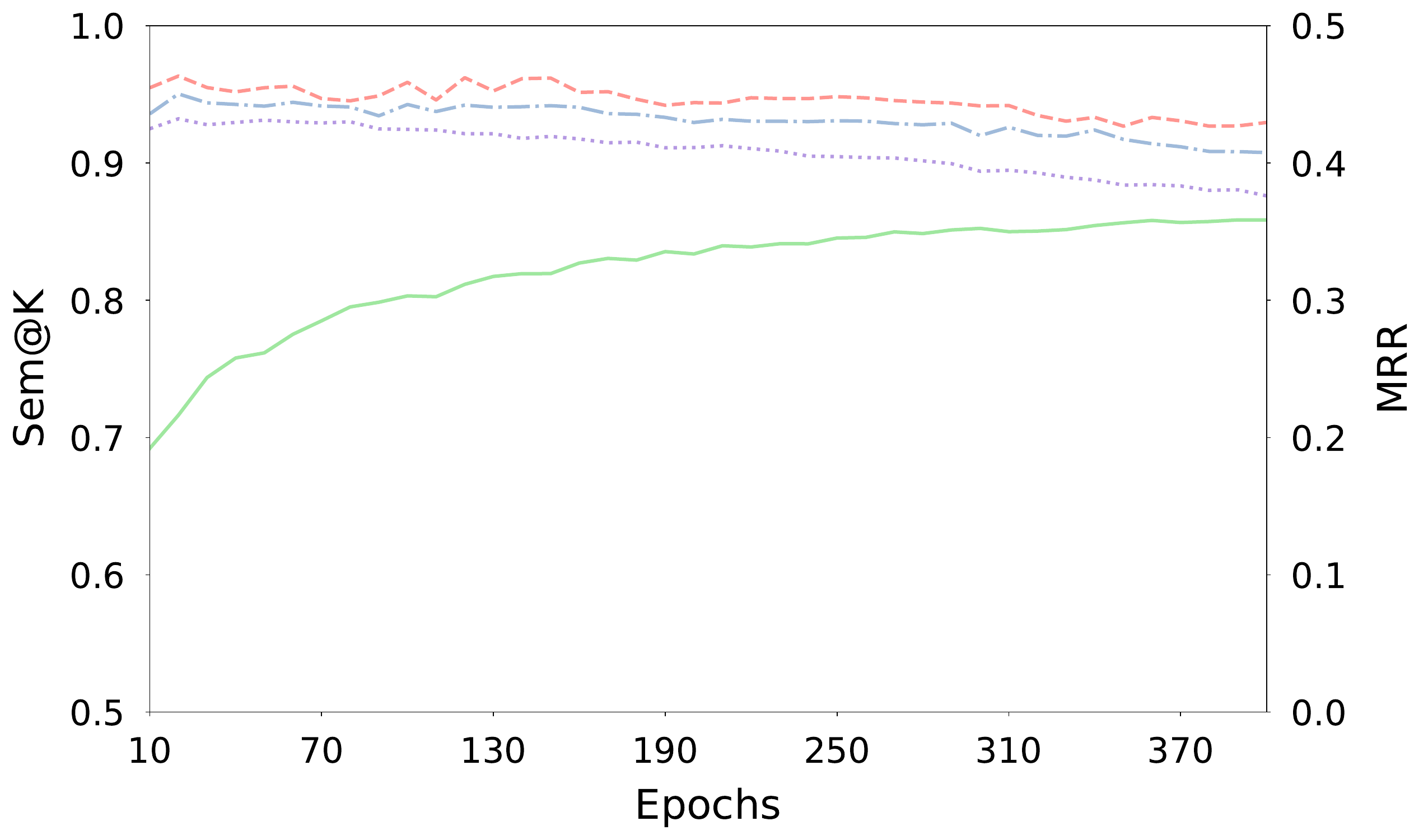}
        \caption{TransH -- Codex-S}
        \label{subfig:codexs-transh}
    \end{subfigure}
    
    \begin{subfigure}[c]{0.49\textwidth}
        \centering
        \includegraphics[width=\textwidth]{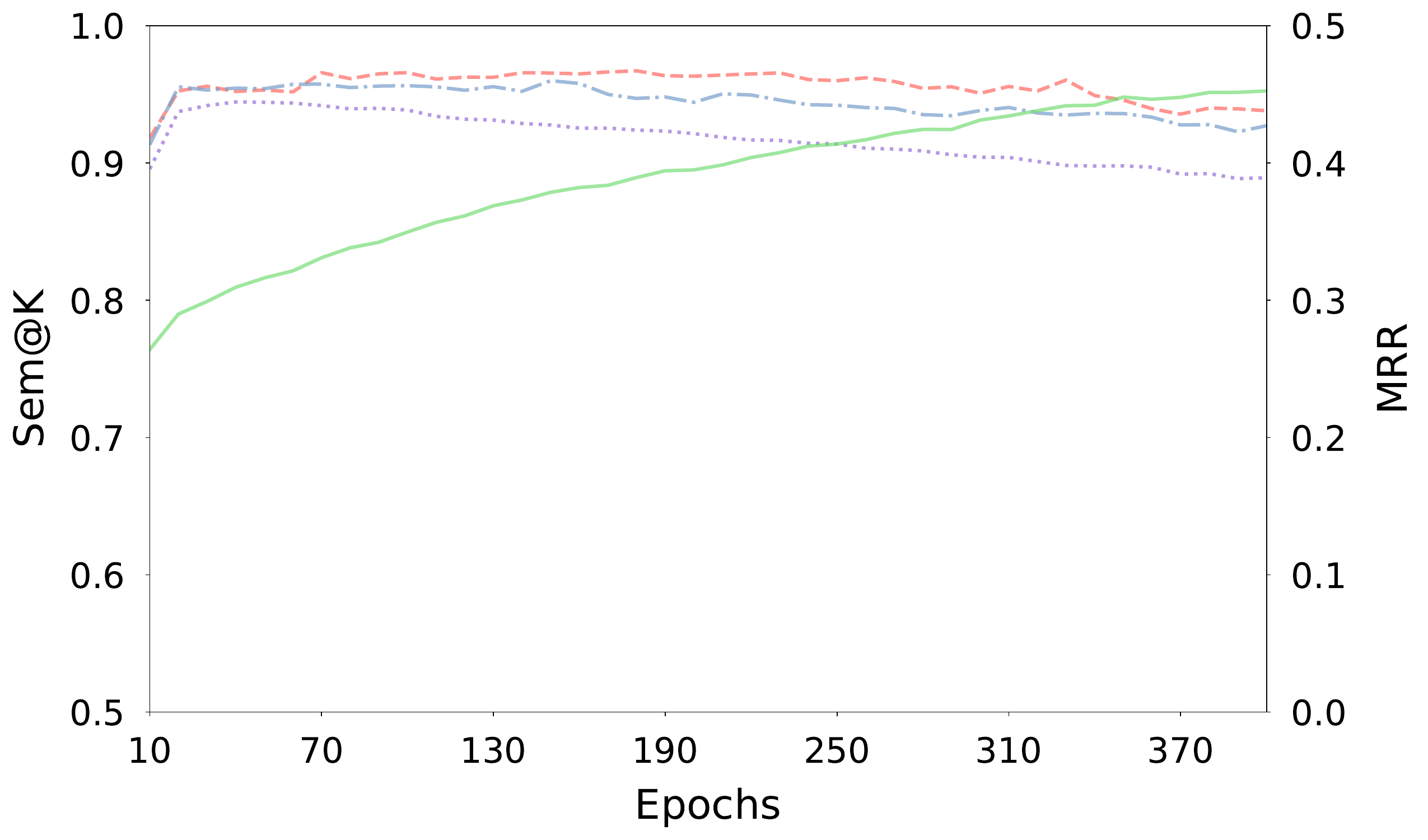}
        \caption{DistMult -- Codex-S}
        \label{subfig:codexs-distmult}
    \end{subfigure}
    \hfill
    \begin{subfigure}[c]{0.49\textwidth}
        \centering
        \includegraphics[width=\textwidth]{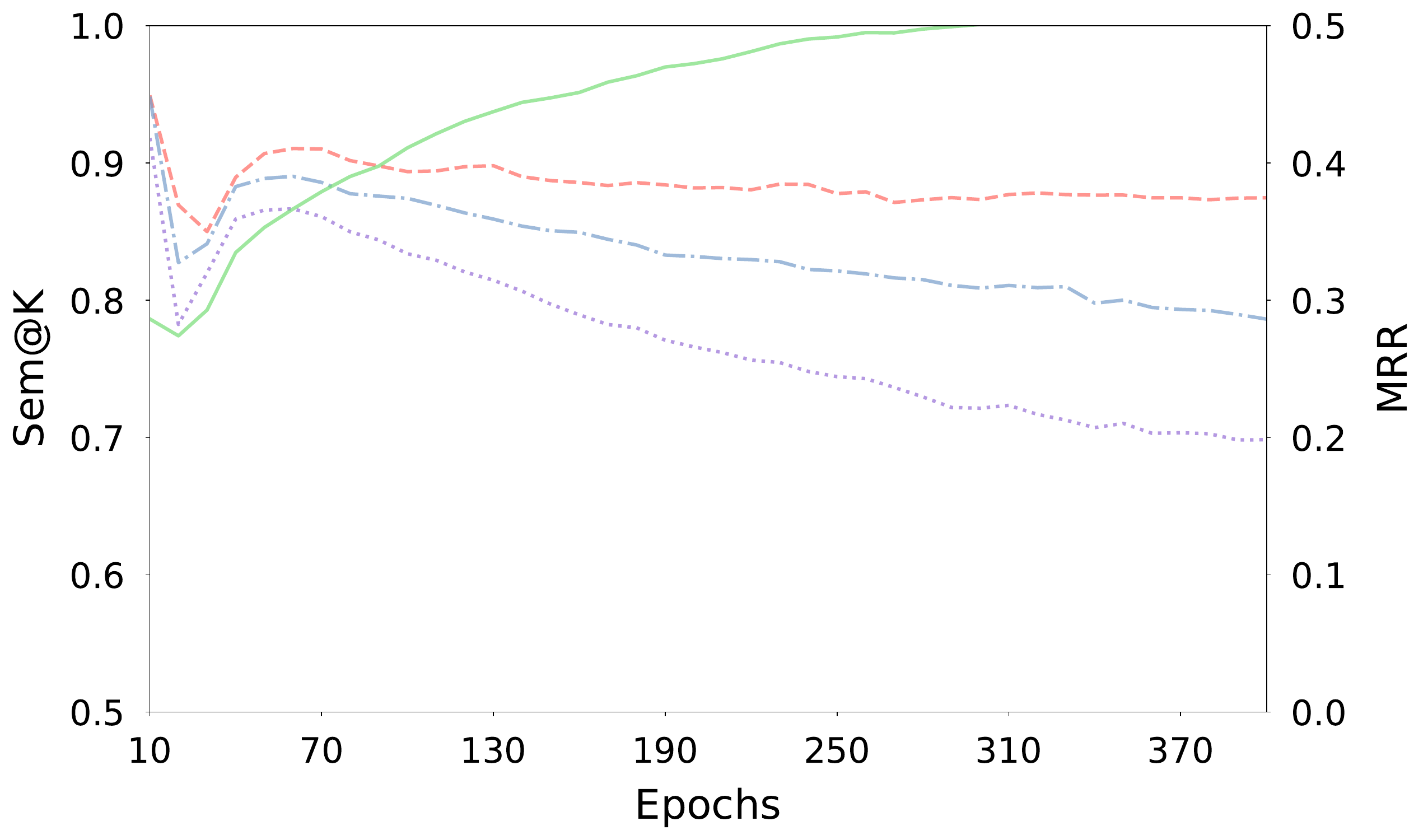}
        \caption{ComplEx -- Codex-S}
        \label{subfig:codexs-complex}
    \end{subfigure}
    
    \begin{subfigure}[c]{0.49\textwidth}
        \centering
        \includegraphics[width=\textwidth]{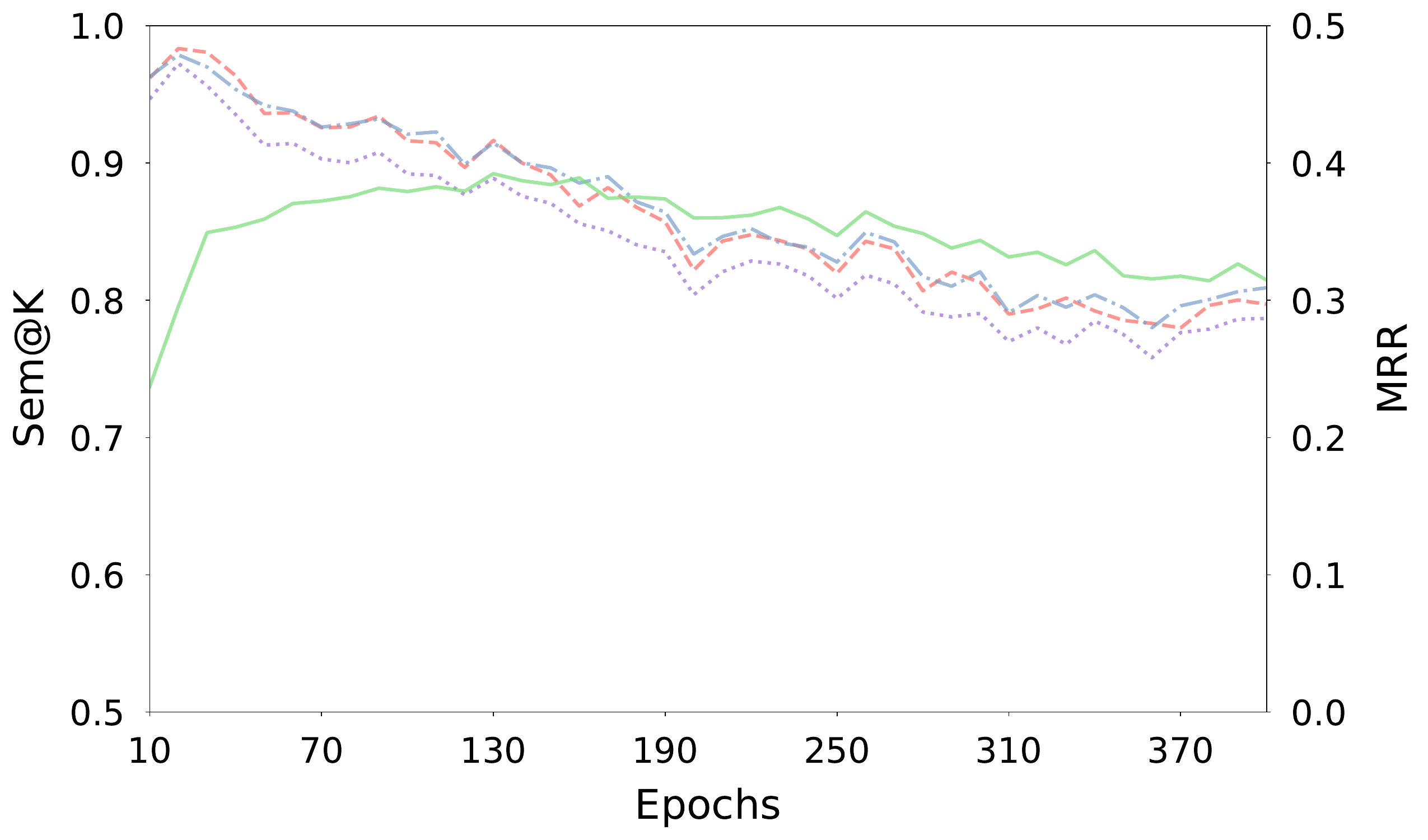}
        \caption{ConvE -- Codex-S}
        \label{subfig:codexs-conve}
    \end{subfigure}
    \hfill
    \begin{subfigure}[c]{0.49\textwidth}
        \centering
        \includegraphics[width=\textwidth]{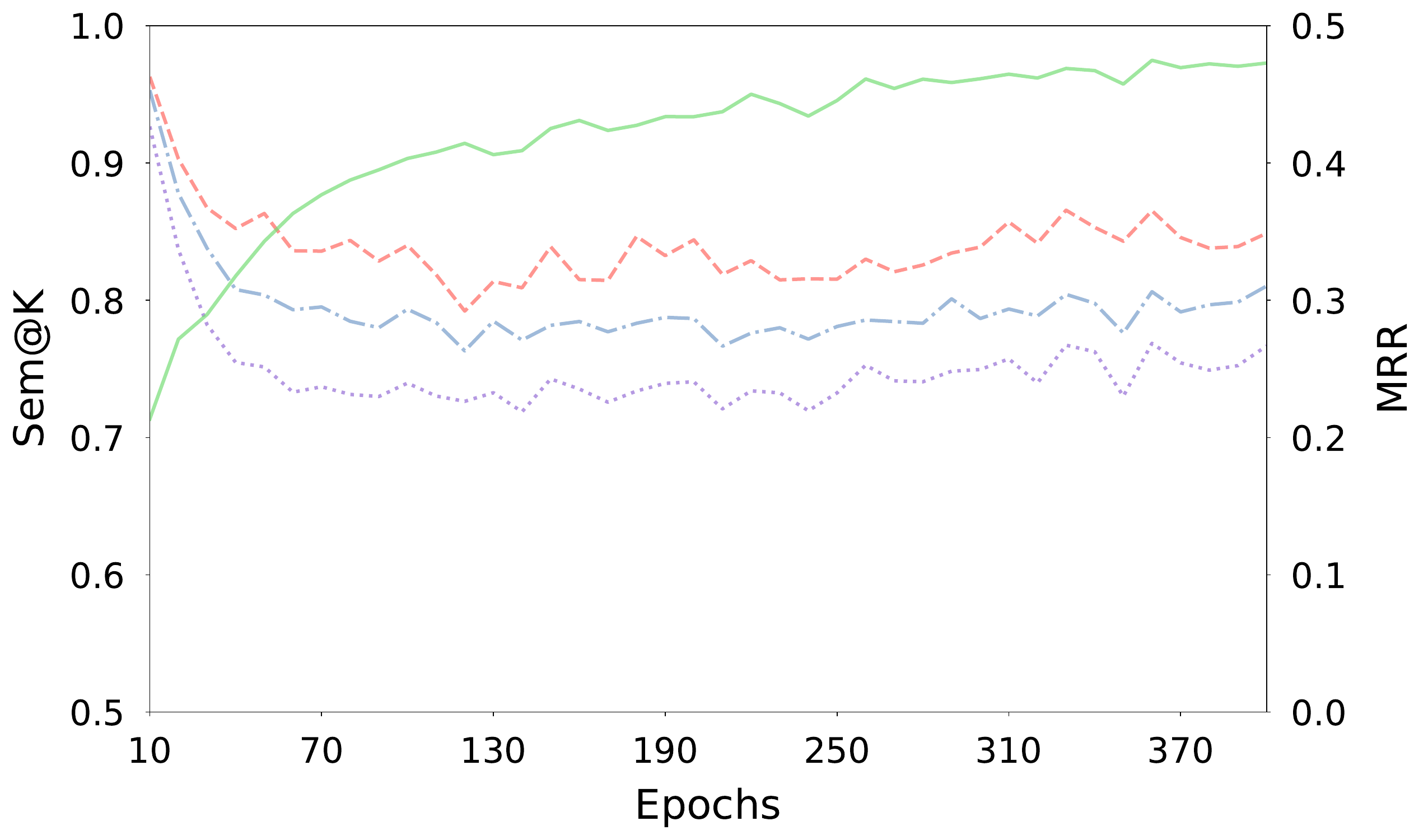}
        \caption{ConvKB -- Codex-S}
        \label{subfig:codexs-convkb}
    \end{subfigure}

    \begin{subfigure}[c]{0.49\textwidth}
        \centering
        \includegraphics[width=\textwidth]{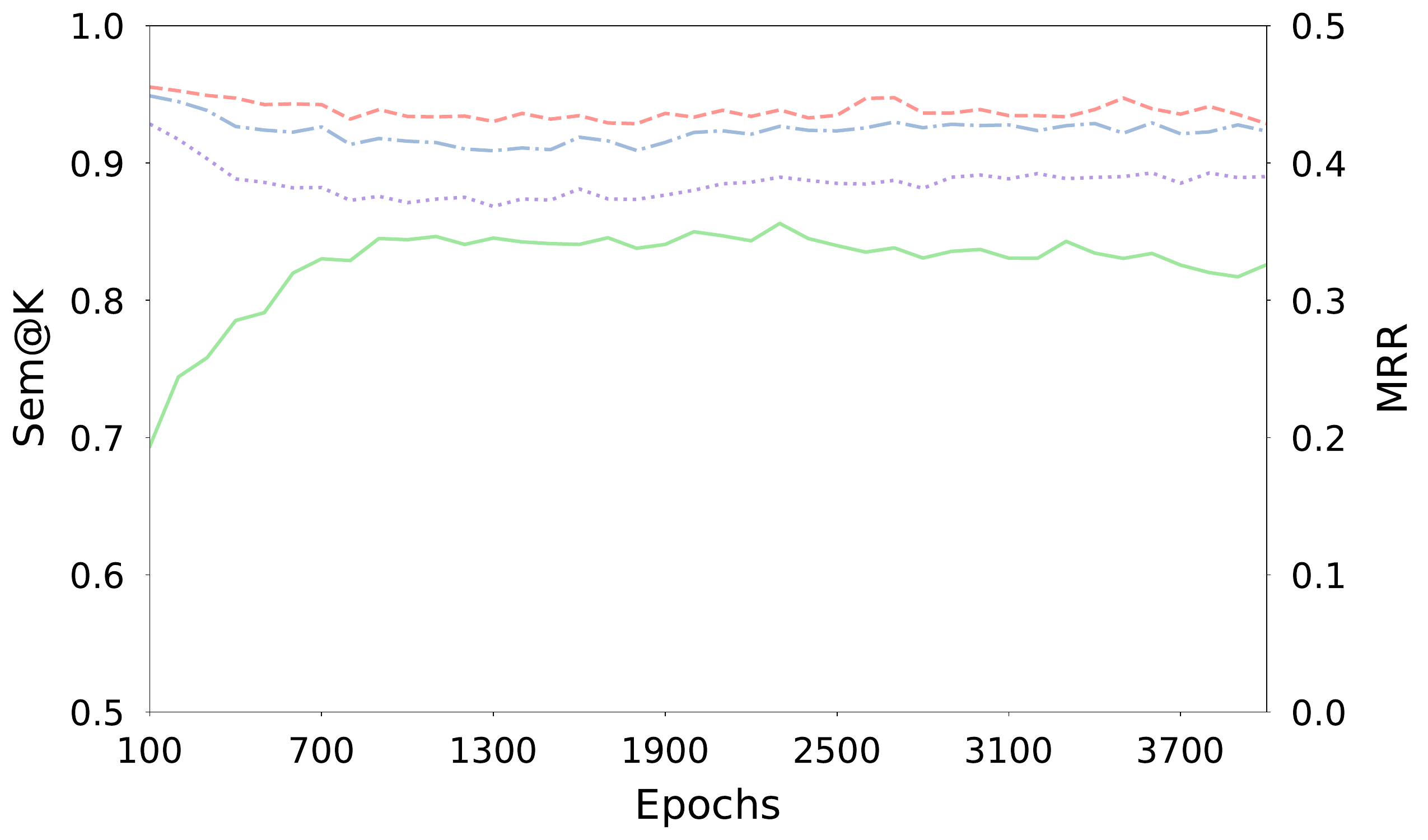}
        \caption{R-GCN -- Codex-S}
        \label{subfig:codexs-rgcn}
    \end{subfigure}
    \hfill
    \begin{subfigure}[c]{0.49\textwidth}
        \centering
        \includegraphics[width=\textwidth]{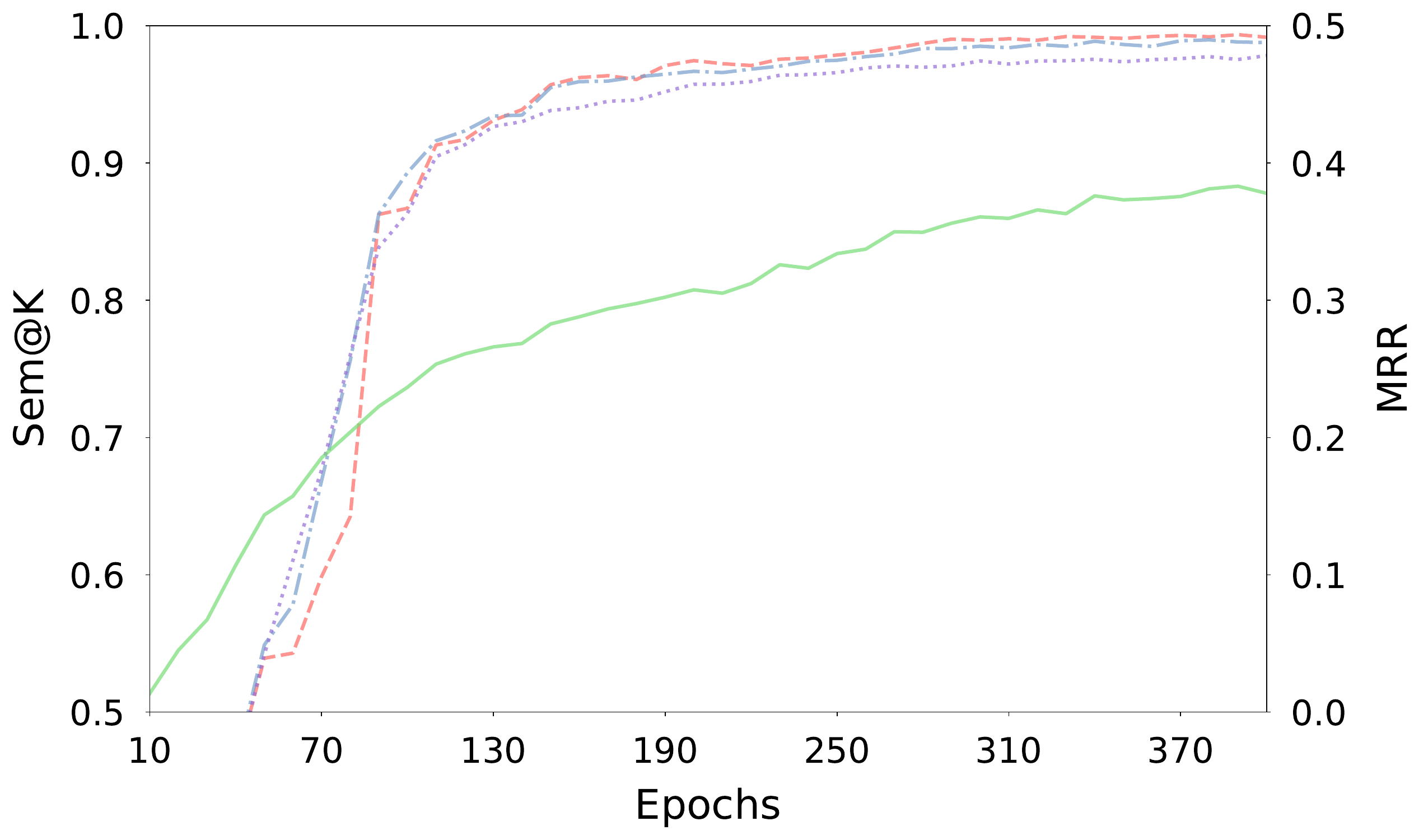}
        \caption{CompGCN -- Codex-S}
        \label{subfig:codexs-compgcn}
    \end{subfigure}

    \caption[]{Evolution of MRR (\raisebox{2pt}{\begin{tikzpicture}[scale=0.5]
        \draw[color=pastelgreen, solid, line width=1pt] (0,0) -- (0.7,0);
    \end{tikzpicture}}), Sem@1 (\raisebox{2pt}{\begin{tikzpicture}[scale=0.5]
        \draw[color=pastelred, dashed, line width=1pt] (0,0) -- (0.65,0);
    \end{tikzpicture}}), Sem@3 (\raisebox{2pt}{\begin{tikzpicture}[scale=0.5]
        \draw[color=darkpastelblue, dash dot, line width=1pt] (0,0) -- (0.7,0);
    \end{tikzpicture}}), and Sem@10 (\raisebox{2pt}{\begin{tikzpicture}[scale=0.5]
        \draw[color=darkpastelpurple, dotted, line width=1pt] (0,0) -- (0.7,0);
    \end{tikzpicture}}) on Codex-S}
    \label{fig:results5}
\end{figure}

\begin{figure}[h]
    \centering
    \begin{subfigure}[c]{0.49\textwidth}
        \centering
        \includegraphics[width=\textwidth]{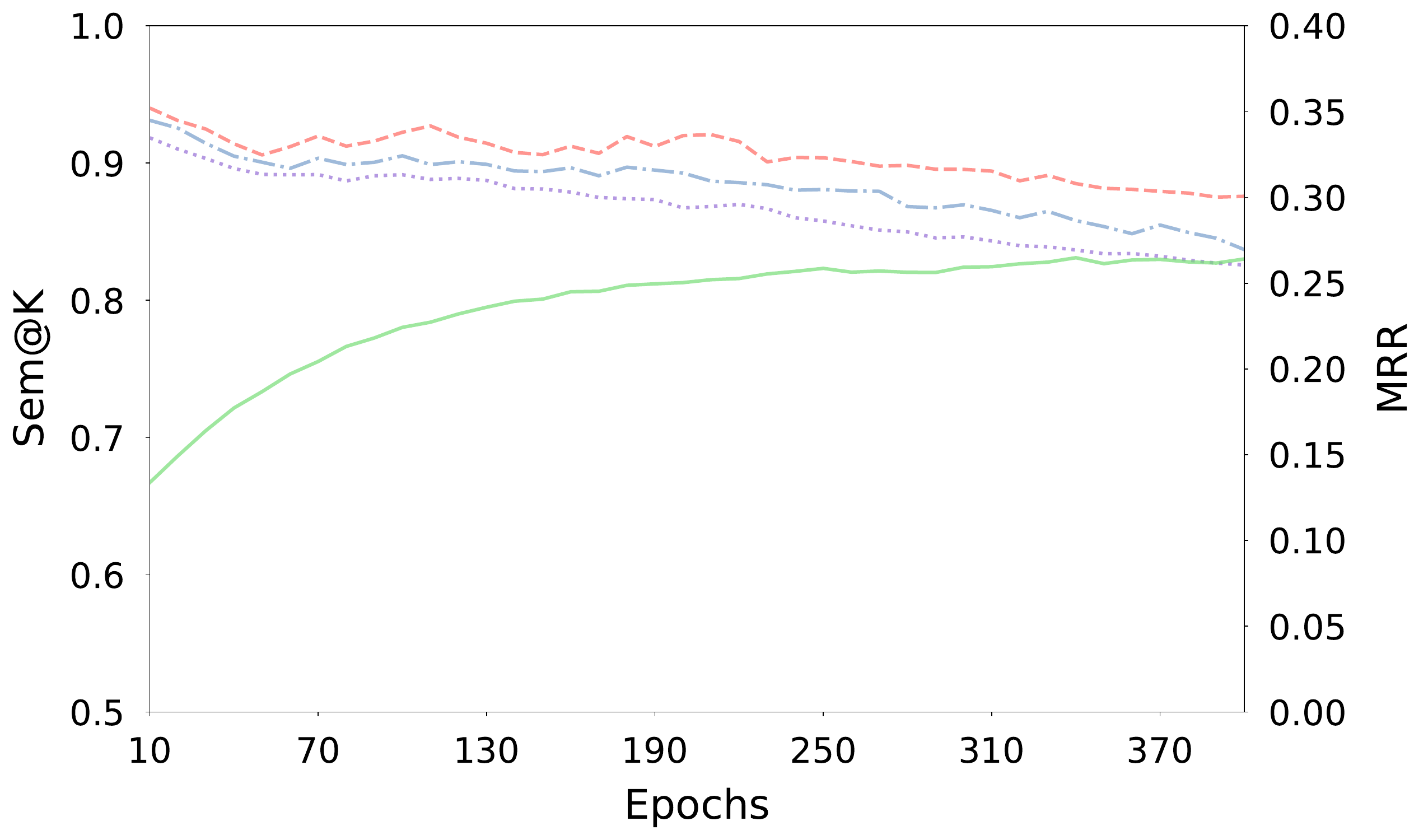}
        \caption{TransE -- Codex-M}
        \label{subfig:codexm-transe}
    \end{subfigure}
    \hfill
    \begin{subfigure}[c]{0.49\textwidth}
        \centering
        \includegraphics[width=\textwidth]{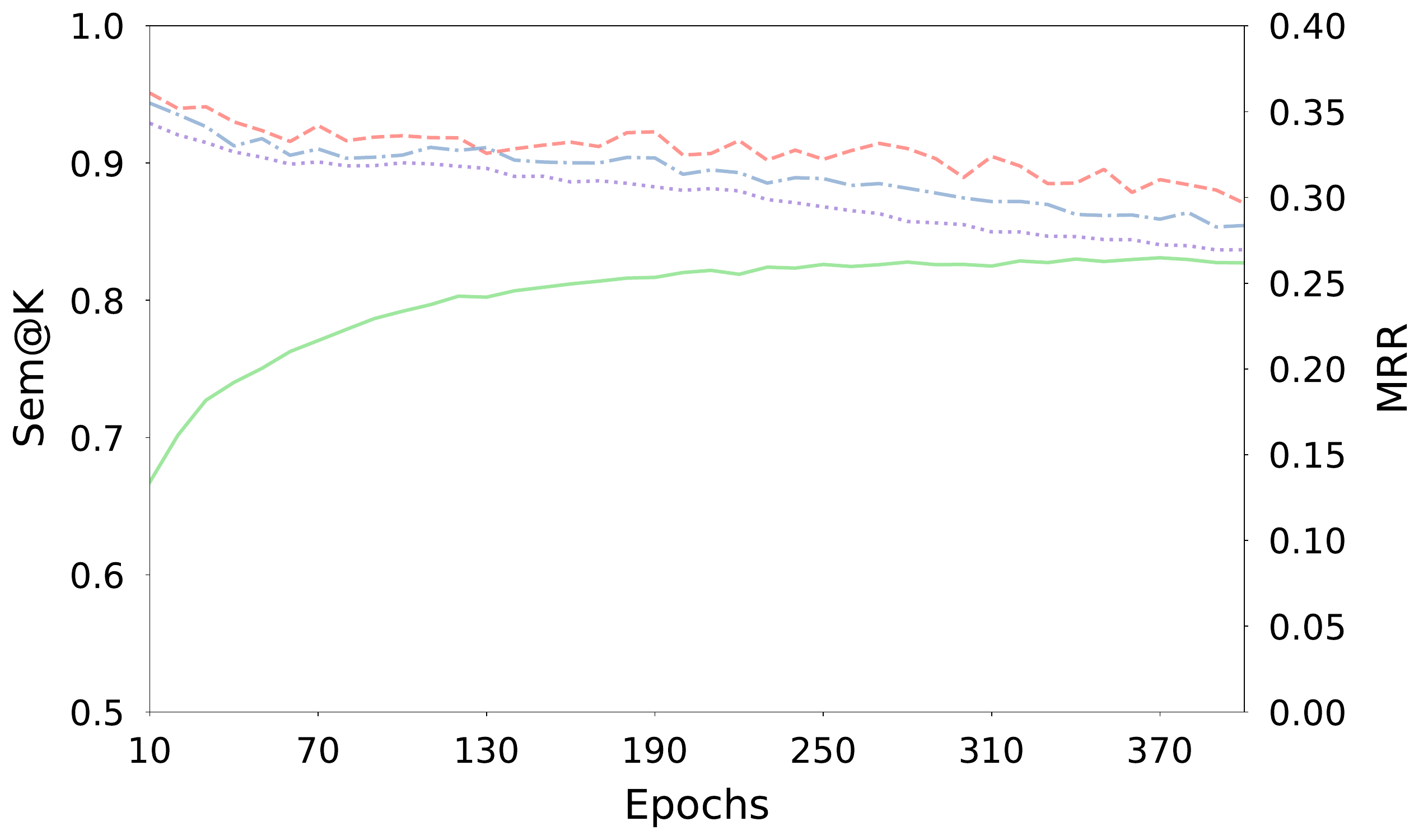}
        \caption{TransH -- Codex-M}
        \label{subfig:codexm-transh}
    \end{subfigure}
    
    \begin{subfigure}[c]{0.49\textwidth}
        \centering
        \includegraphics[width=\textwidth]{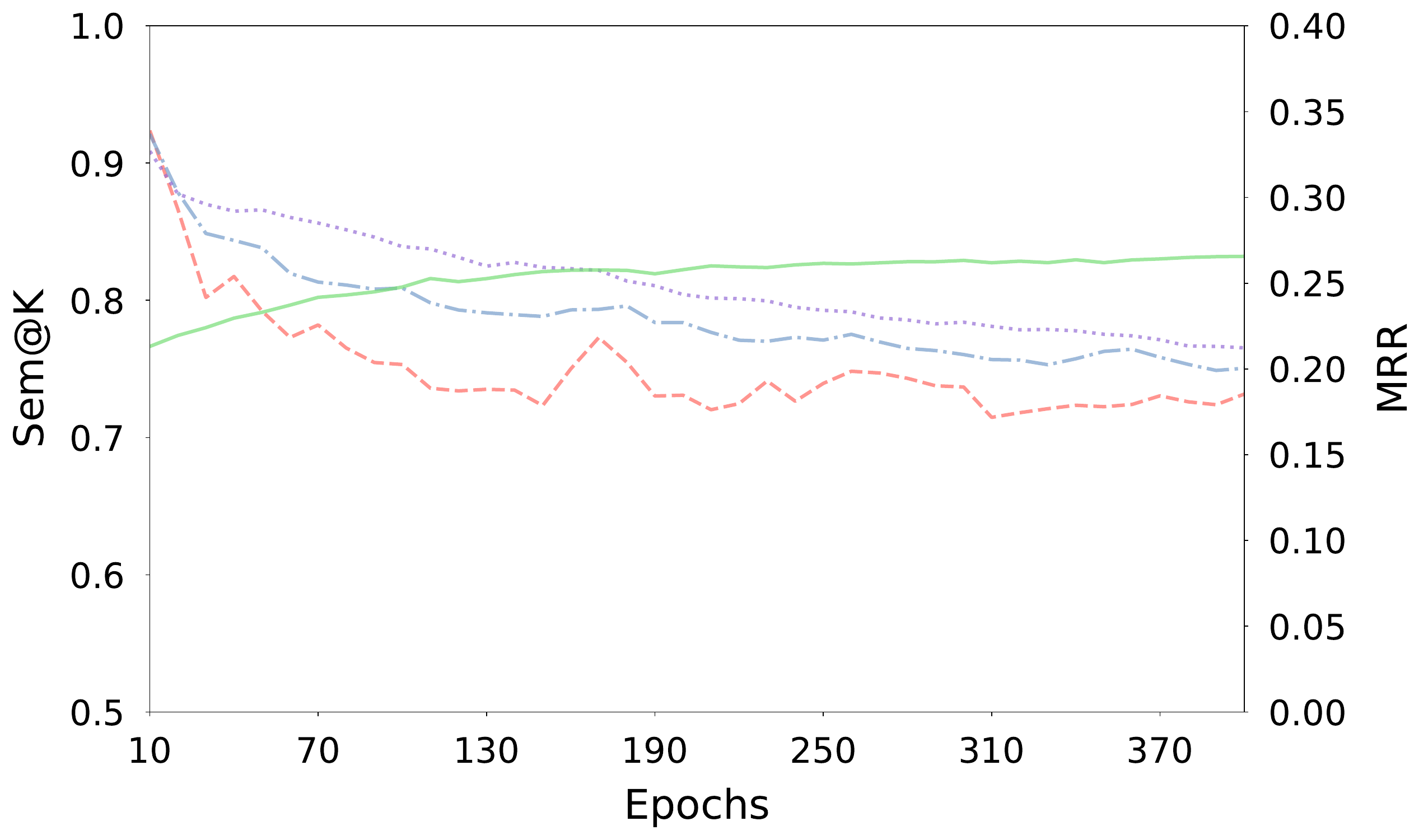}
        \caption{DistMult -- Codex-M}
        \label{subfig:codexm-distmult}
    \end{subfigure}
    \hfill
    \begin{subfigure}[c]{0.49\textwidth}
        \centering
        \includegraphics[width=\textwidth]{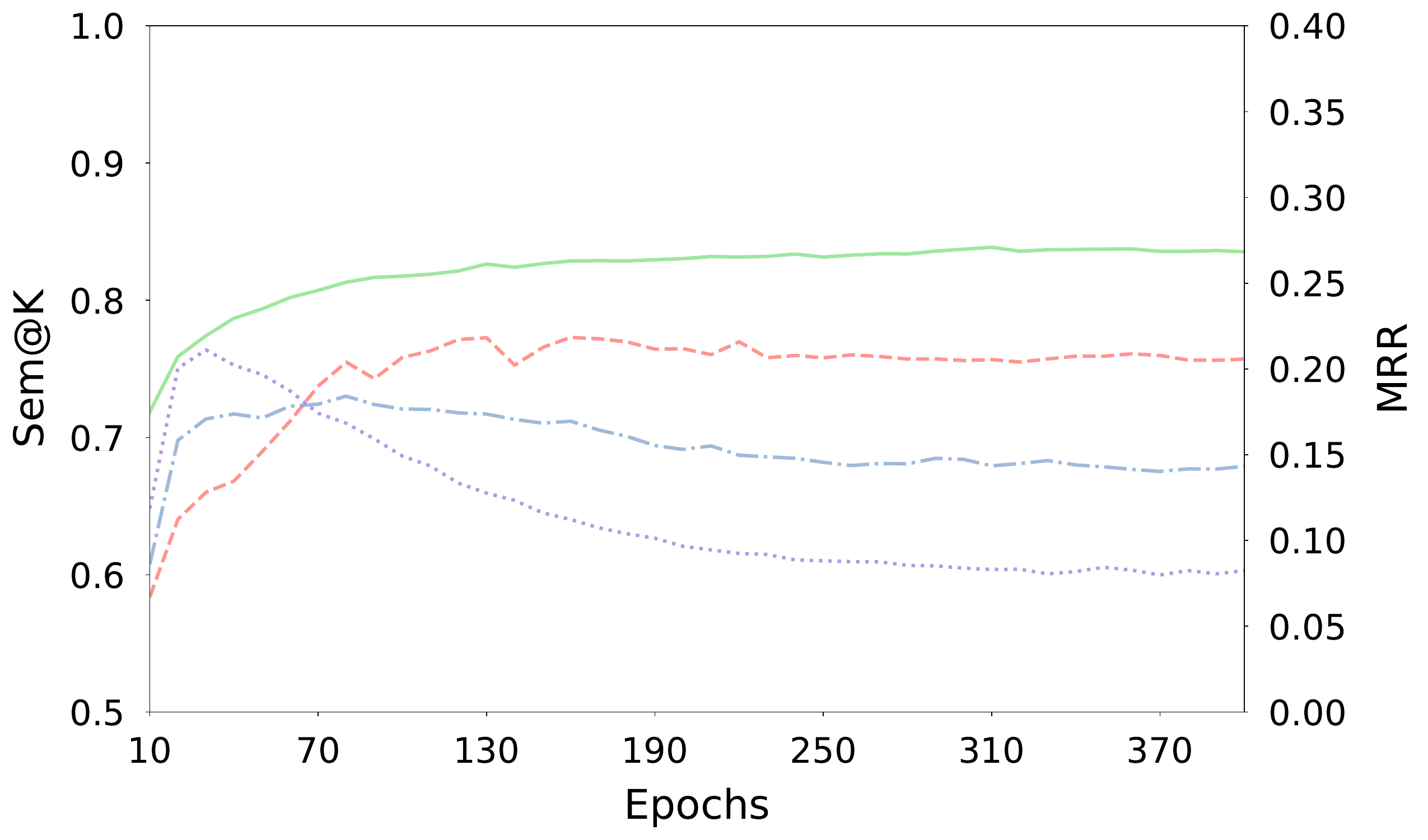}
        \caption{ComplEx -- Codex-M}
        \label{subfig:codexm-complex}
    \end{subfigure}
    
    \begin{subfigure}[c]{0.49\textwidth}
        \centering
        \includegraphics[width=\textwidth]{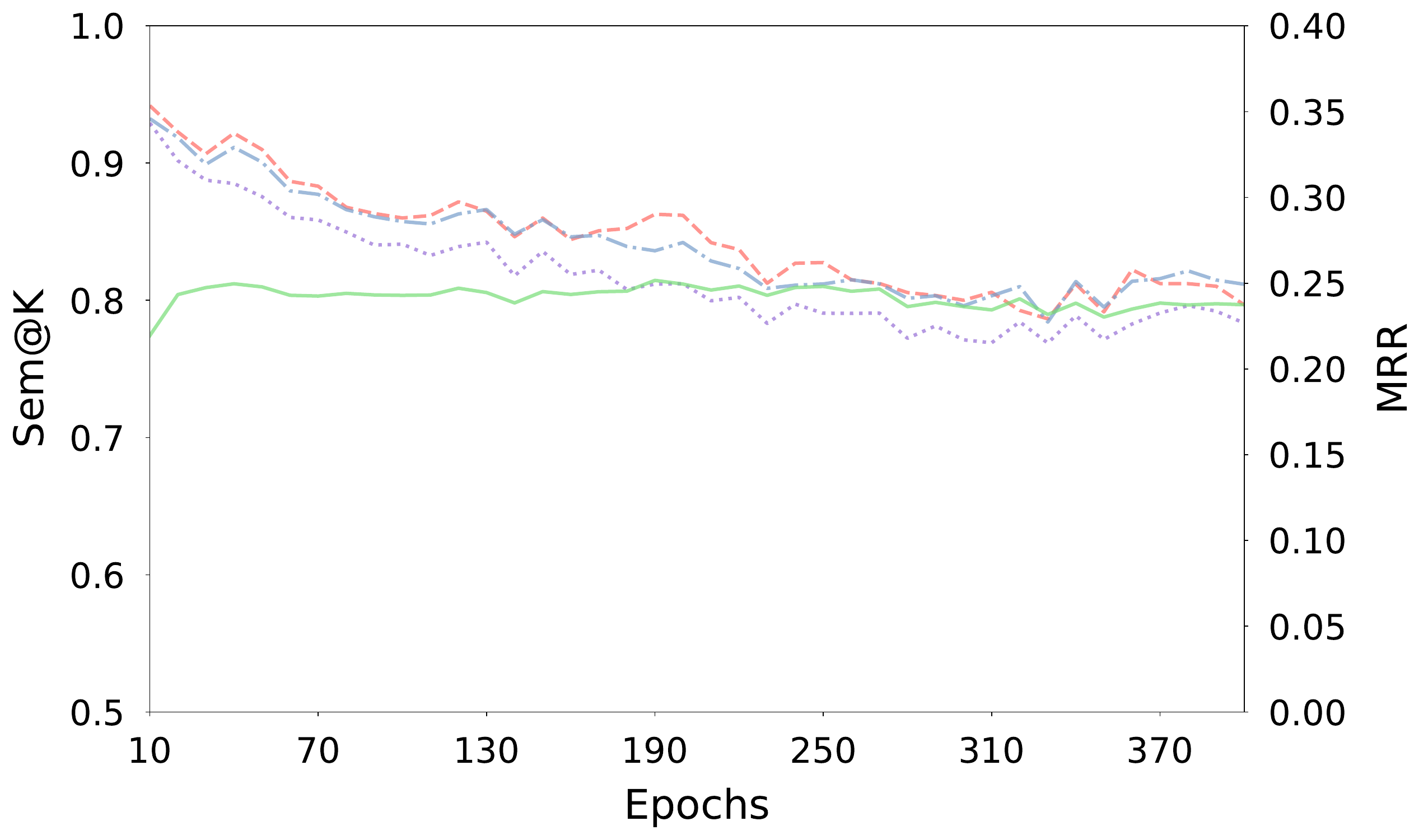}
        \caption{ConvE -- Codex-M}
        \label{subfig:codexm-conve}
    \end{subfigure}
    \hfill
    \begin{subfigure}[c]{0.49\textwidth}
        \centering
        \includegraphics[width=\textwidth]{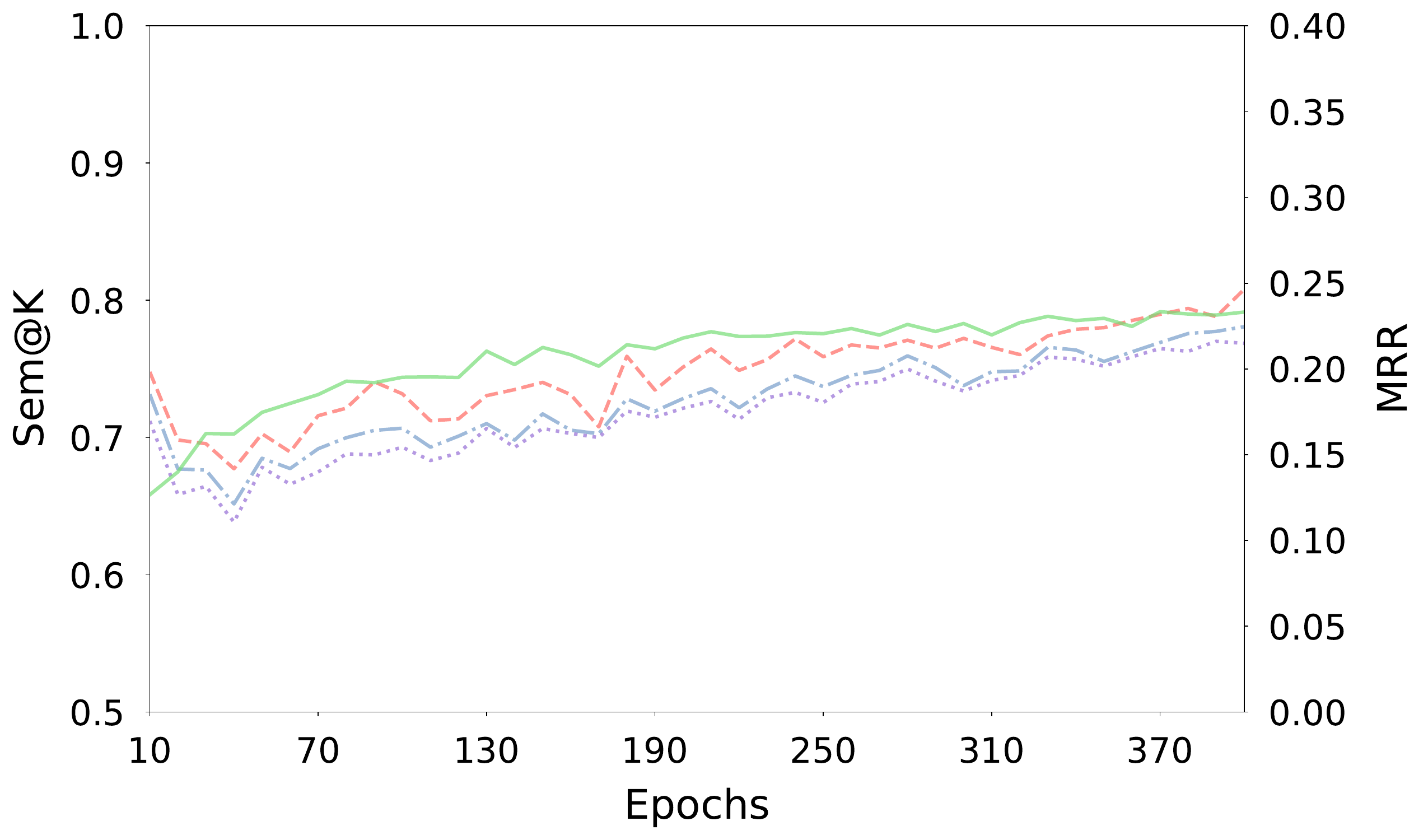}
        \caption{ConvKB -- Codex-M}
        \label{subfig:codexm-convkb}
    \end{subfigure}

    \begin{subfigure}[c]{0.49\textwidth}
        \centering
        \includegraphics[width=\textwidth]{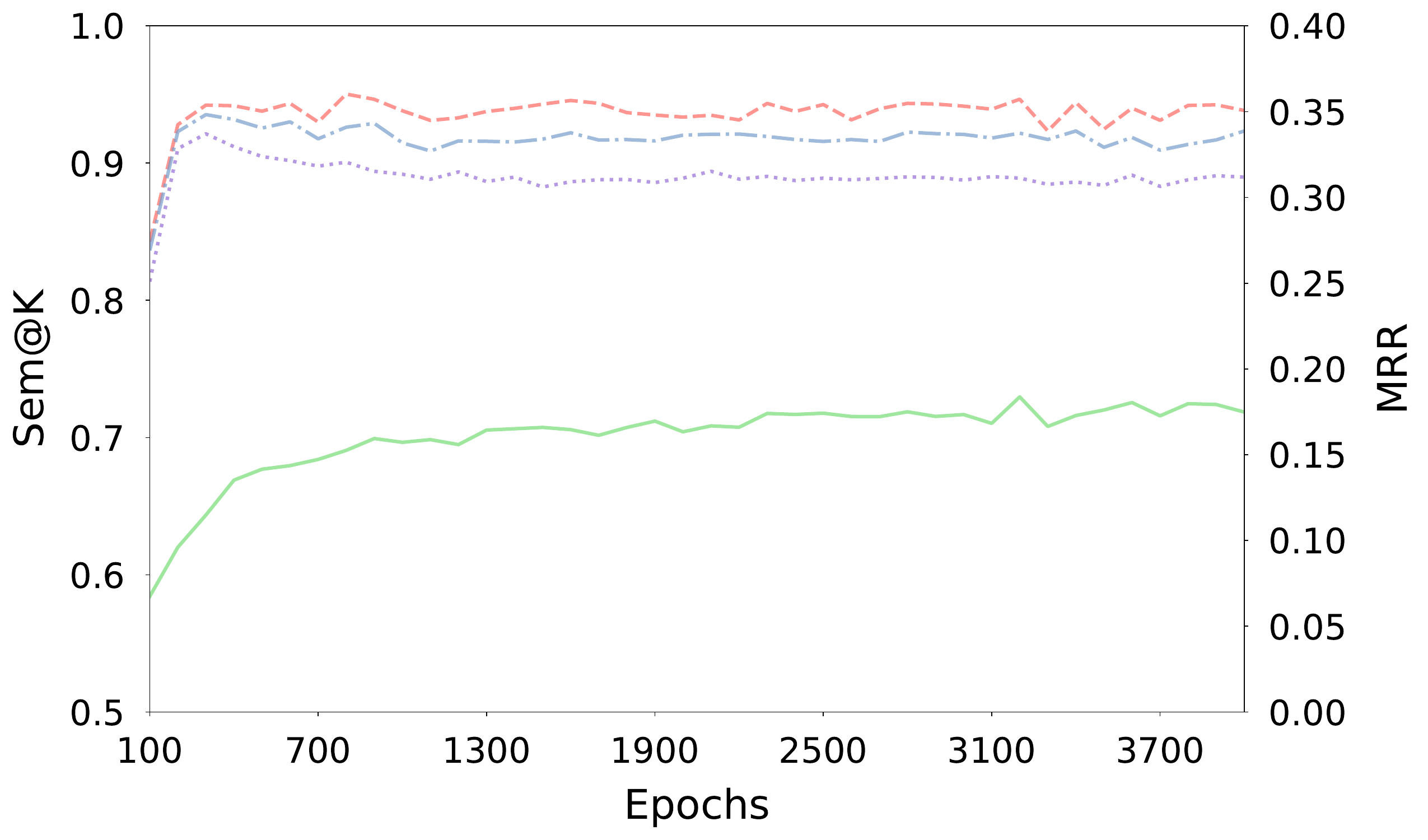}
        \caption{R-GCN -- Codex-M}
        \label{subfig:codexm-rgcn}
    \end{subfigure}
    \hfill
    \begin{subfigure}[c]{0.49\textwidth}
        \centering
        \includegraphics[width=\textwidth]{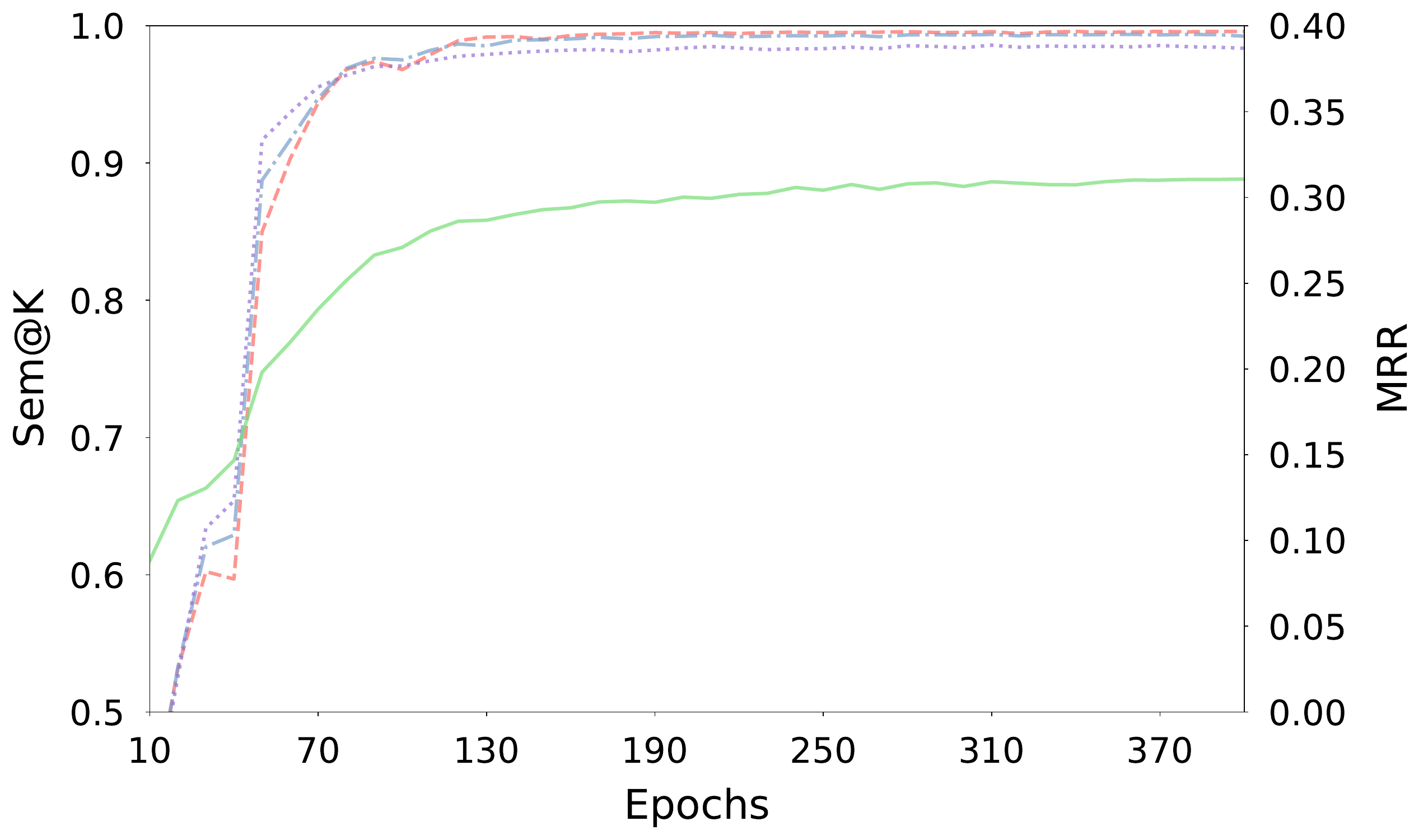}
        \caption{CompGCN -- Codex-M}
        \label{subfig:codexm-compgcn}
    \end{subfigure}

    \caption[]{Evolution of MRR (\raisebox{2pt}{\begin{tikzpicture}[scale=0.5]
        \draw[color=pastelgreen, solid, line width=1pt] (0,0) -- (0.7,0);
    \end{tikzpicture}}), Sem@1 (\raisebox{2pt}{\begin{tikzpicture}[scale=0.5]
        \draw[color=pastelred, dashed, line width=1pt] (0,0) -- (0.65,0);
    \end{tikzpicture}}), Sem@3 (\raisebox{2pt}{\begin{tikzpicture}[scale=0.5]
        \draw[color=darkpastelblue, dash dot, line width=1pt] (0,0) -- (0.7,0);
    \end{tikzpicture}}), and Sem@10 (\raisebox{2pt}{\begin{tikzpicture}[scale=0.5]
        \draw[color=darkpastelpurple, dotted, line width=1pt] (0,0) -- (0.7,0);
    \end{tikzpicture}}) on Codex-M}
    \label{fig:results6}
\end{figure}

\begin{figure}[h]
    \centering
    \begin{subfigure}[c]{0.49\textwidth}
        \centering
        \includegraphics[width=\textwidth]{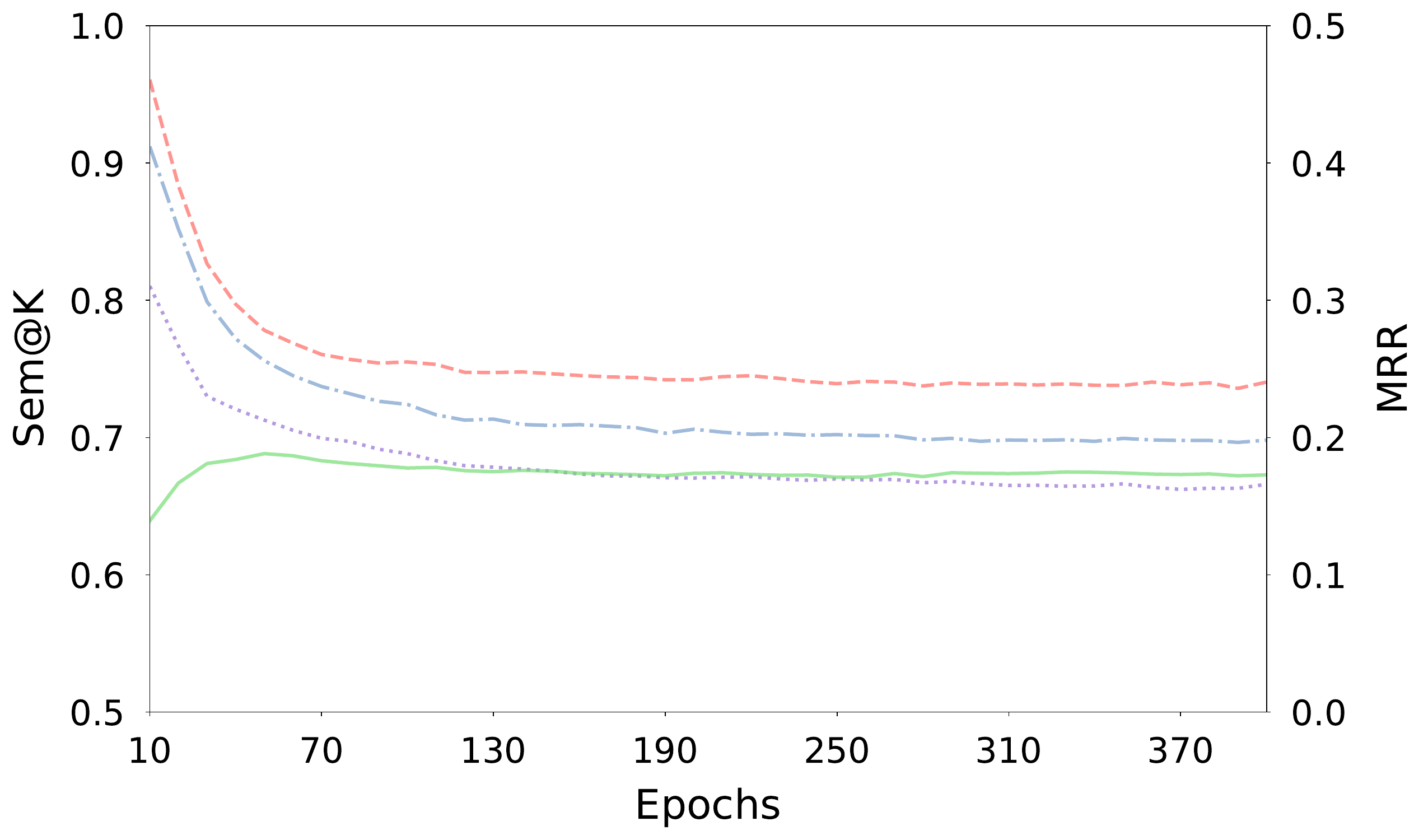}
        \caption{TransE -- WN18RR}
        \label{subfig:wn18rr-transe}
    \end{subfigure}
    \hfill
    \begin{subfigure}[c]{0.49\textwidth}
        \centering
        \includegraphics[width=\textwidth]{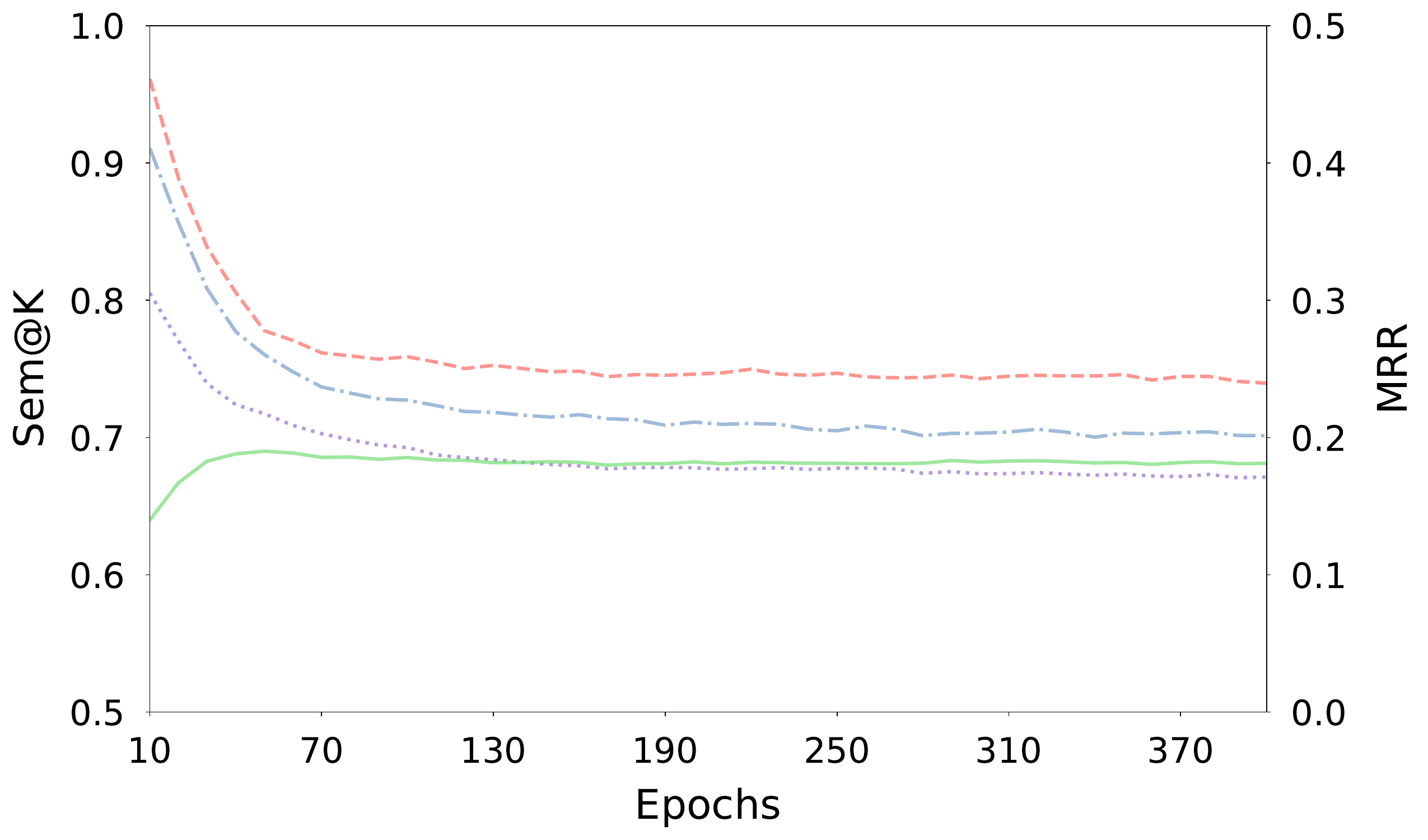}
        \caption{TransH -- WN18RR}
        \label{subfig:wn18rr-transh}
    \end{subfigure}
    
    \begin{subfigure}[c]{0.49\textwidth}
        \centering
        \includegraphics[width=\textwidth]{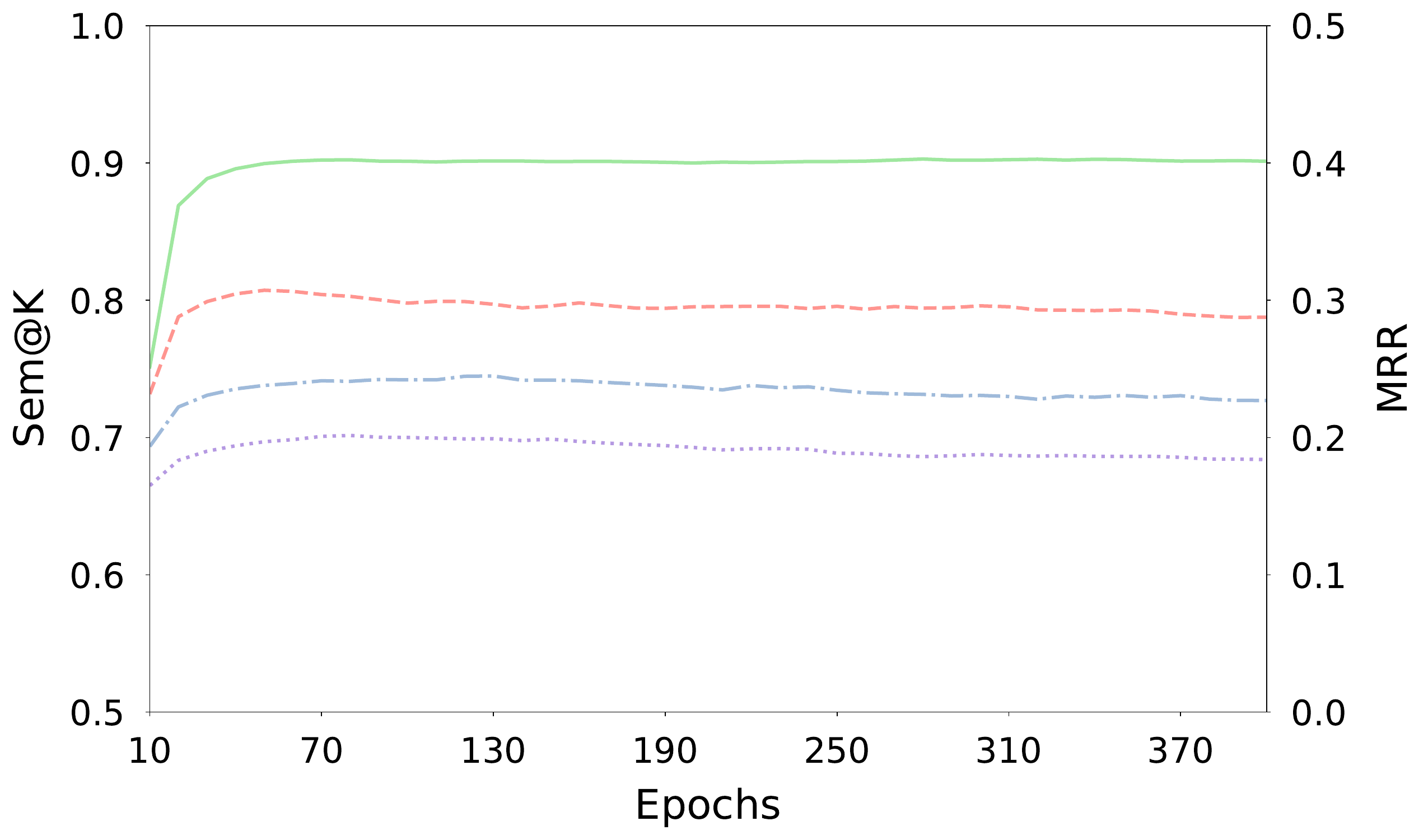}
        \caption{DistMult -- WN18RR}
        \label{subfig:wn18rr-distmult}
    \end{subfigure}
    \hfill
    \begin{subfigure}[c]{0.49\textwidth}
        \centering
        \includegraphics[width=\textwidth]{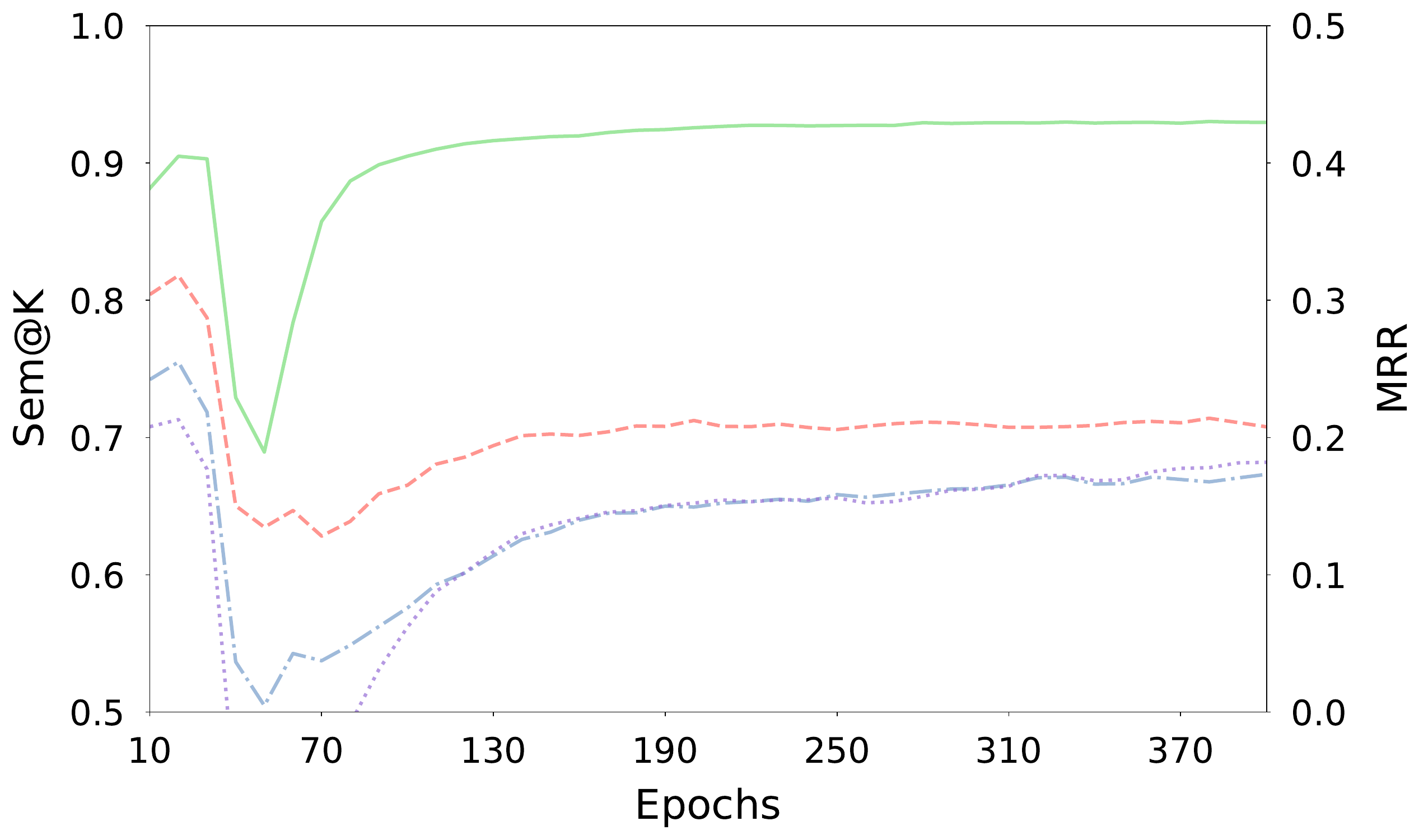}
        \caption{ComplEx -- WN18RR}
        \label{subfig:wn18rr-complex}
    \end{subfigure}
    
    \begin{subfigure}[c]{0.49\textwidth}
        \centering
        \includegraphics[width=\textwidth]{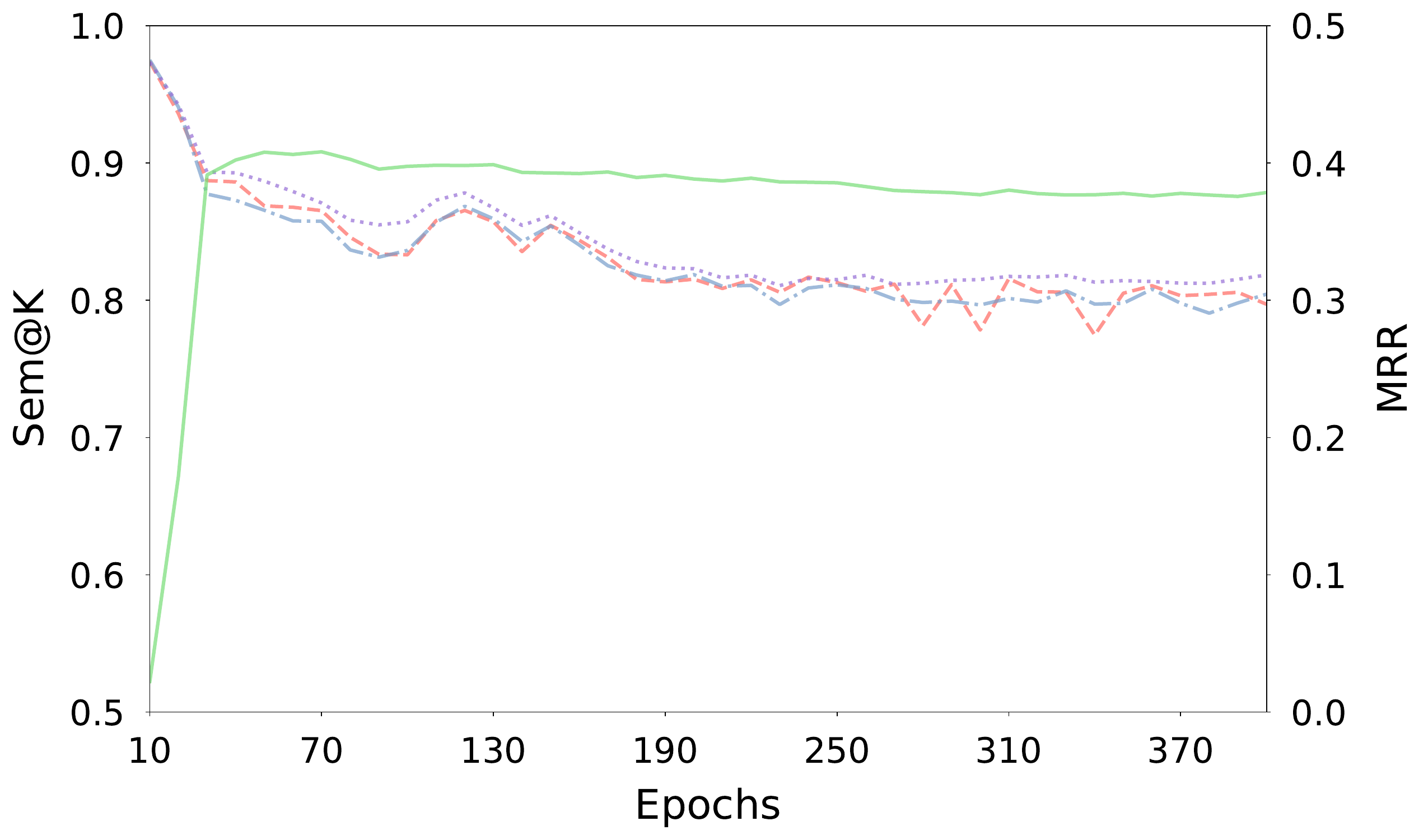}
        \caption{ConvE -- WN18RR}
        \label{subfig:wn18rr-conve}
    \end{subfigure}
    \hfill
    \begin{subfigure}[c]{0.49\textwidth}
        \centering
        \includegraphics[width=\textwidth]{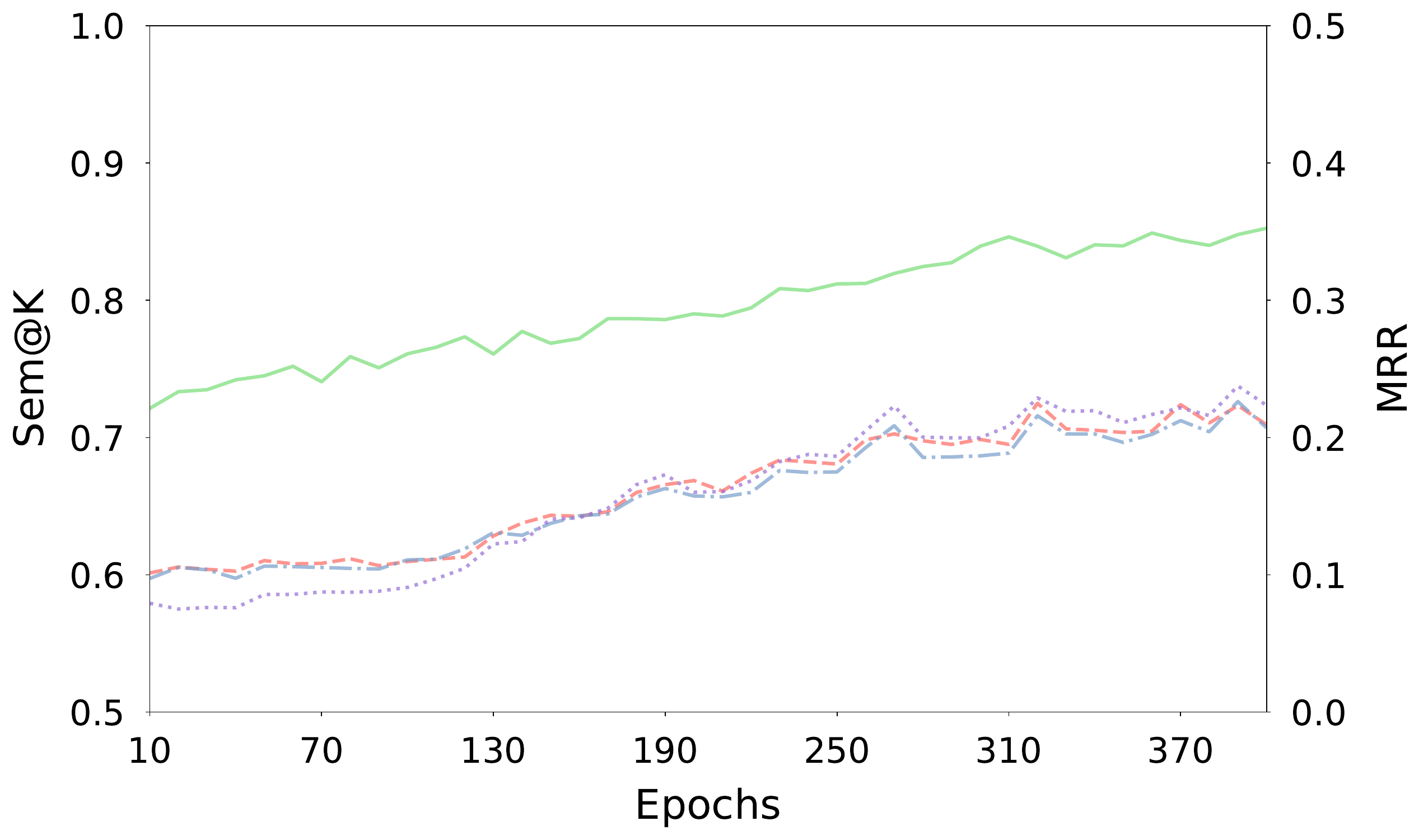}
        \caption{ConvKB -- WN18RR}
        \label{subfig:wn18rr-convkb}
    \end{subfigure}

    \begin{subfigure}[c]{0.49\textwidth}
        \centering
        \includegraphics[width=\textwidth]{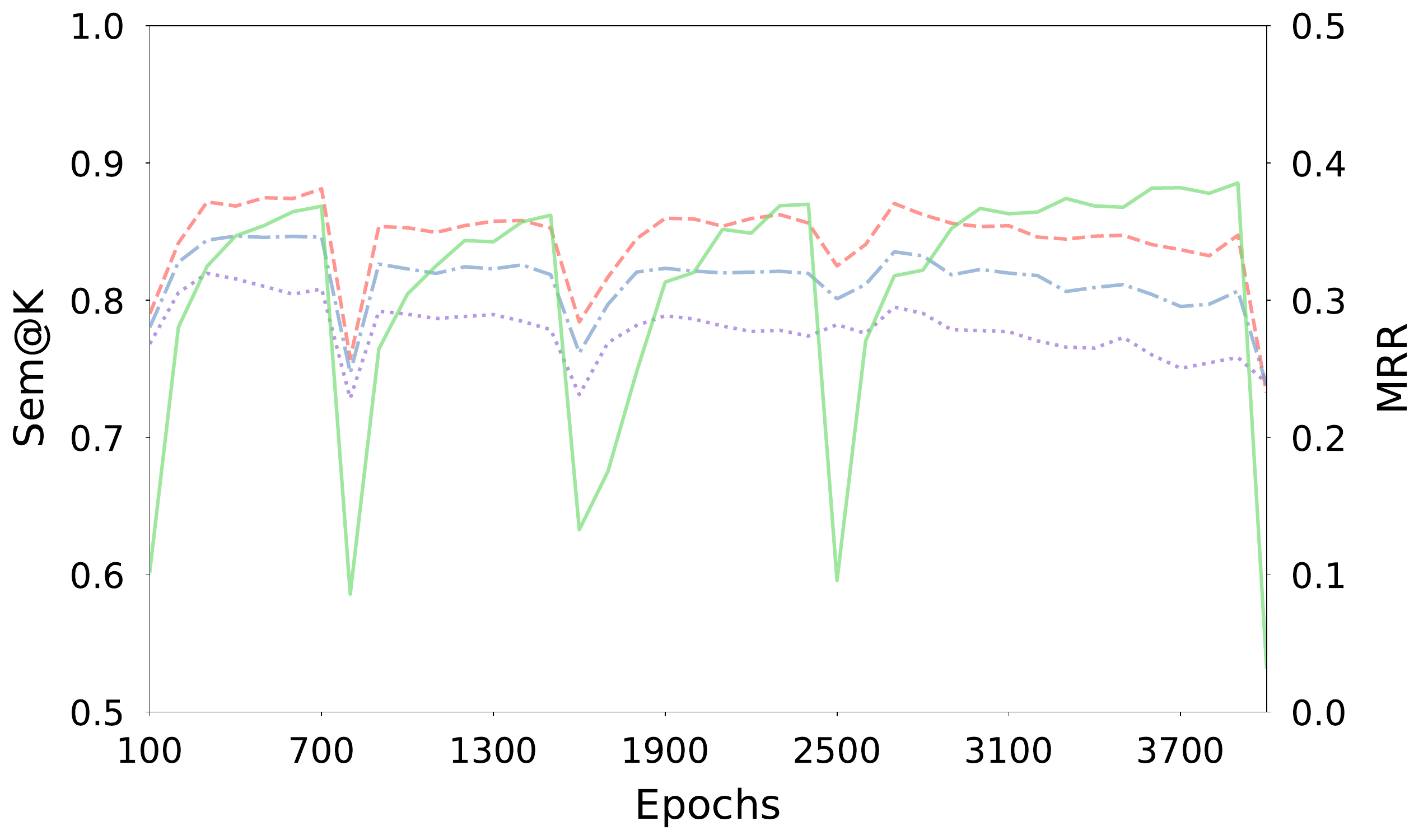}
        \caption{R-GCN -- WN18RR}
        \label{subfig:wn18rr-rgcn}
    \end{subfigure}
    \hfill
    \begin{subfigure}[c]{0.49\textwidth}
        \centering
        \includegraphics[width=\textwidth]{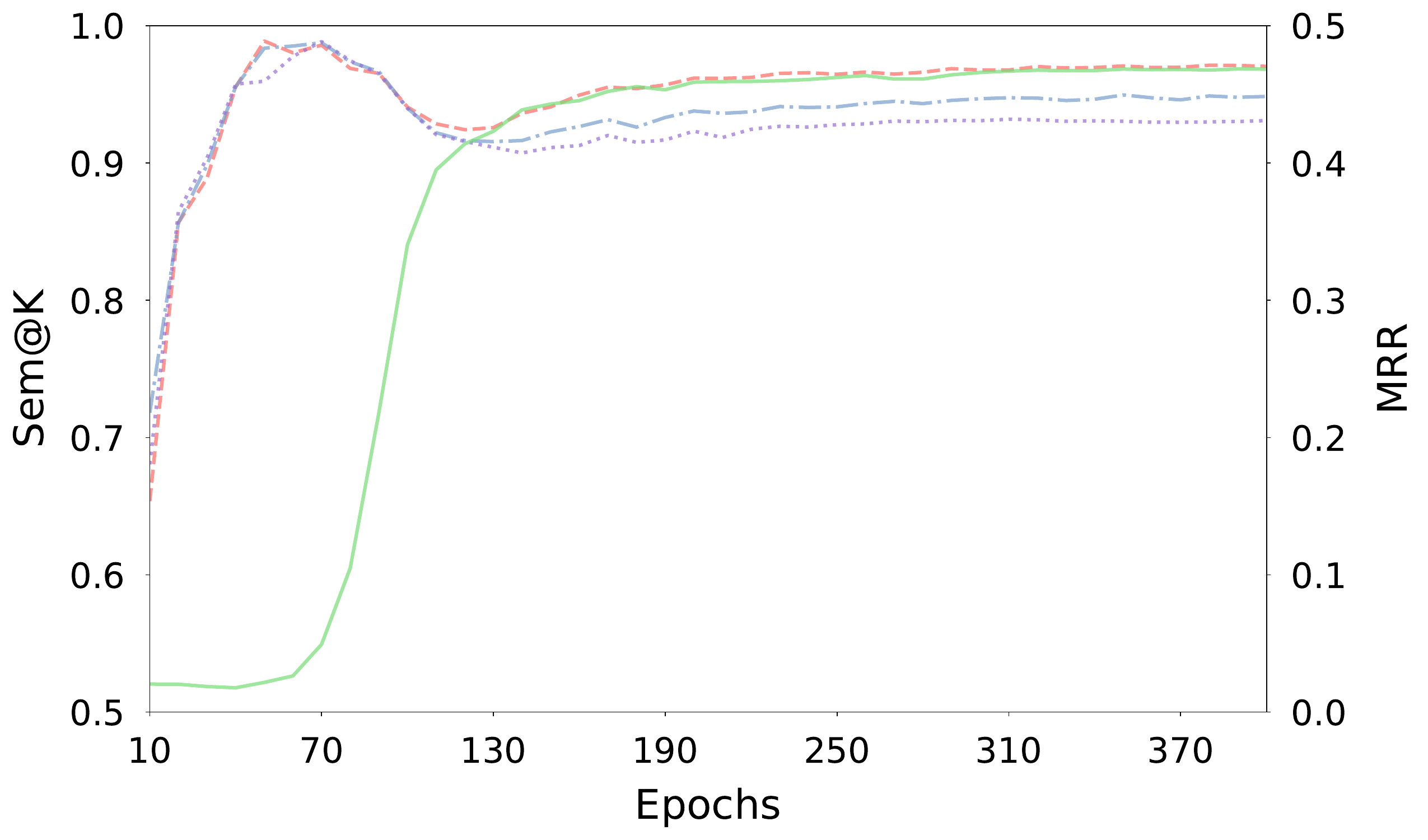}
        \caption{CompGCN -- WN18RR}
        \label{subfig:wn18rr-compgcn}
    \end{subfigure}

    \caption[]{Evolution of MRR (\raisebox{2pt}{\begin{tikzpicture}[scale=0.5]
        \draw[color=pastelgreen, solid, line width=1pt] (0,0) -- (0.7,0);
    \end{tikzpicture}}), Sem@1 (\raisebox{2pt}{\begin{tikzpicture}[scale=0.5]
        \draw[color=pastelred, dashed, line width=1pt] (0,0) -- (0.65,0);
    \end{tikzpicture}}), Sem@3 (\raisebox{2pt}{\begin{tikzpicture}[scale=0.5]
        \draw[color=darkpastelblue, dash dot, line width=1pt] (0,0) -- (0.7,0);
    \end{tikzpicture}}), and Sem@10 (\raisebox{2pt}{\begin{tikzpicture}[scale=0.5]
        \draw[color=darkpastelpurple, dotted, line width=1pt] (0,0) -- (0.7,0);
    \end{tikzpicture}}) on WN18RR}
    \label{fig:results7}
\end{figure}

The evolution of MRR and Sem@$K$ values with respect to the number of epochs is presented in Fig.~\ref{fig:results}, \ref{fig:results2} , \ref{fig:results3}, \ref{fig:results4}, \ref{fig:results5}, \ref{fig:results6}, and~\ref{fig:results7}. For equity and clarity sakes, we choose to present 2 KGEMs for each family of model (translational models, semantic matching models, CNNs, and GNNs). Regarding semantic matching models, DistMult and ComplEx are chosen, as the evolution of their MRR and Sem@$K$ values is less erratic than SimplE. The evolution of MRR and Sem@$K$ values for SimplE are made available on the GitHub repository of the datasets\footnote{\url{https://github.com/nicolas-hbt/benchmark-sematk}}.

\clearpage

\end{appendix}

\bibliographystyle{unsrtnat}
\bibliography{references}
\end{document}